\documentclass[format=acmlarge, review=false, screen=true]{acmart}

\usepackage{booktabs} 
\usepackage{multirow}
\usepackage{subfig}
\usepackage[ruled]{algorithm2e}
\usepackage{longtable}

\acmJournal{IMWUT}
\acmYear{0}\acmVolume{0}\acmNumber{0}\acmArticle{000}\acmMonth{0}
\acmDOI{00.0000/00000.00000}

\begin{document}
\title{Multi-task Self-Supervised Learning for Human Activity Detection}

\author{Aaqib Saeed}
\orcid{0000-0003-1473-0322}
\affiliation{
  \institution{Eindhoven University of Technology}
  \city{Eindhoven}
  \country{The Netherlands}
}
\email{a.saeed@tue.nl}

\author{Tanir Ozcelebi}
\affiliation{
  \institution{Eindhoven University of Technology}
  \city{Eindhoven}
  \country{The Netherlands}
}
\email{t.ozcelebi@tue.nl}

\author{Johan Lukkien}
\affiliation{
 \institution{Eindhoven University of Technology}
 \city{Eindhoven}
 \country{The Netherlands}
}
\email{j.j.lukkien@tue.nl}

\renewcommand\shortauthors{A. Saeed et al.}

\begin{abstract}
Deep learning methods are successfully used in applications pertaining to ubiquitous computing, pervasive intelligence, health, and well-being. Specifically, the area of human activity recognition (HAR) is primarily transformed by the convolutional and recurrent neural networks, thanks to their ability to learn semantic representations directly from raw input. However, in order to extract generalizable features massive amounts of well-curated data are required, which is a notoriously challenging task; hindered by privacy issues and annotation costs. Therefore, unsupervised representation learning (i.e., learning without manually labeling the instances) is of prime importance to leverage the vast amount of unlabeled data produced by smart devices. In this work, we propose a novel self-supervised technique for feature learning from sensory data that does not require access to any form of semantic labels, i.e., activity classes. We learn a multi-task temporal convolutional network to recognize transformations applied on an input signal. By exploiting these transformations, we demonstrate that simple auxiliary tasks of the binary classification result in a strong supervisory signal for extracting useful features for the down-stream task. We extensively evaluate the proposed approach on several publicly available datasets for smartphone-based HAR in unsupervised, semi-supervised and transfer learning settings. Our method achieves performance levels superior to or comparable with fully-supervised networks trained directly with activity labels, and it performs significantly better than unsupervised learning through autoencoders. Notably, for the semi-supervised case, the self-supervised features substantially boost the detection rate by attaining a kappa score between $0.7-0.8$ with only $10$ labeled examples per class. We get similar impressive performance even if the features are transferred from a different data source. Self-supervision drastically reduces the requirement of labeled activity data, effectively narrowing the gap between supervised and unsupervised techniques for learning meaningful representations. While this paper focuses on HAR as the application domain, the proposed approach is general and could be applied to a wide variety of problems in other areas.
\end{abstract}

\ccsdesc[500]{Human-centered computing~Ubiquitous and mobile computing systems and tools}
\ccsdesc[500]{Computing methodologies~Machine learning}

\keywords{Self-supervised learning, multi-task learning, representation learning, semi-supervised learning, transfer learning, temporal convolutional neural networks, human activity recognition, deep learning}

\maketitle

\section{Introduction}

Over the last years, deep neural networks have been widely adopted for time-series and sensory data processing; achieving impressive performance in several application areas pertaining to pervasive sensing, ubiquitous computing, industries, health and well-being~\cite{radu2018multimodal, georgiev2017low, saeed2017personalized, Hannun2019, liu2016deepfood, yao2018sensegan}. In particular, for smartphone-based human activity recognition (HAR), $1$D convolutional and recurrent neural networks trained on raw labeled signals significantly improve the detection rate over traditional methods~\cite{wang2018deep, hammerla2016deep, morales2016deep, yang2015deep, yao2018sensegan}. Despite the recent advances in the field of HAR, learning representations from a massive amount of unlabeled data still presents a significant challenge. Obtaining large, well-curated activity recognition datasets is problematic due to a number of issues. First, smartphone data are privacy sensitive, which makes it hard to collect sufficient amounts of user-activity instances in a real-life setting. Second, the annotation cost and the time it takes to generate a large volume of labeled instances are prohibitive. Finally, the diversity of devices, types of embedded sensors, variations in phone-usage, and different environments are further roadblocks in producing massive human-labeled data. To sum up, such expensive and hard to scale process of gathering labeled data generated by smart devices makes it very difficult to apply supervised learning in this domain directly.   

In light of these challenges, we pose the question \textit{whether it is possible to learn semantic representations in an unsupervised way to circumvent the manual annotation of the sensor data with strong labels, e.g., activity classes. In particular, the goal is to extract features that are on par with those learned with fully-supervised~\footnote{The fully-supervised network is the standard deep model that is trained in an end-to-end fashion directly with activity labels without any pre-training.} methods}. There is an emerging paradigm for feature learning called \textit{self-supervised learning} that defines auxiliary (also known as pretext or surrogate) tasks to solve, where labels are readily extractable from the data without any human intervention, i.e., self-supervised. The availability of strong supervisory signals from the surrogate tasks enables us to leverage objective functions as utilized in a standard supervised learning setting~\cite{doersch2017multi}. For instance, the vision community proposed a considerable number of self-supervised tasks for advancing representation learning\footnote{also known as feature learning} from static images, videos, and audio (see Section~\ref{sec:related_work}). Most prominent among them are: colorization of grayscale images~\cite{larsson2017colorization,zhang2017split}, predicting image rotations~\cite{gidaris2018unsupervised}, solving jigsaw puzzles~\cite{noroozi2016unsupervised}, predicting the direction of video playback~\cite{wei2018learning}, temporal order verification~\cite{misra2016shuffle}, odd sequence detection~\cite{fernando2017self}, audio-visual correspondence~\cite{owens2016ambient, arandjelovic2017objects}, and curiosity-driven agents~\cite{pathak2017curiosity}. The presented methodology for sensor representation learning takes inspiration from these methods and takes leverage of signal transformations to extract highly generalizable features for the down-stream\footnote{or an end-task} task, i.e., HAR.    

\begin{figure}[!htbp]
\centering
\includegraphics[width=3.8in]{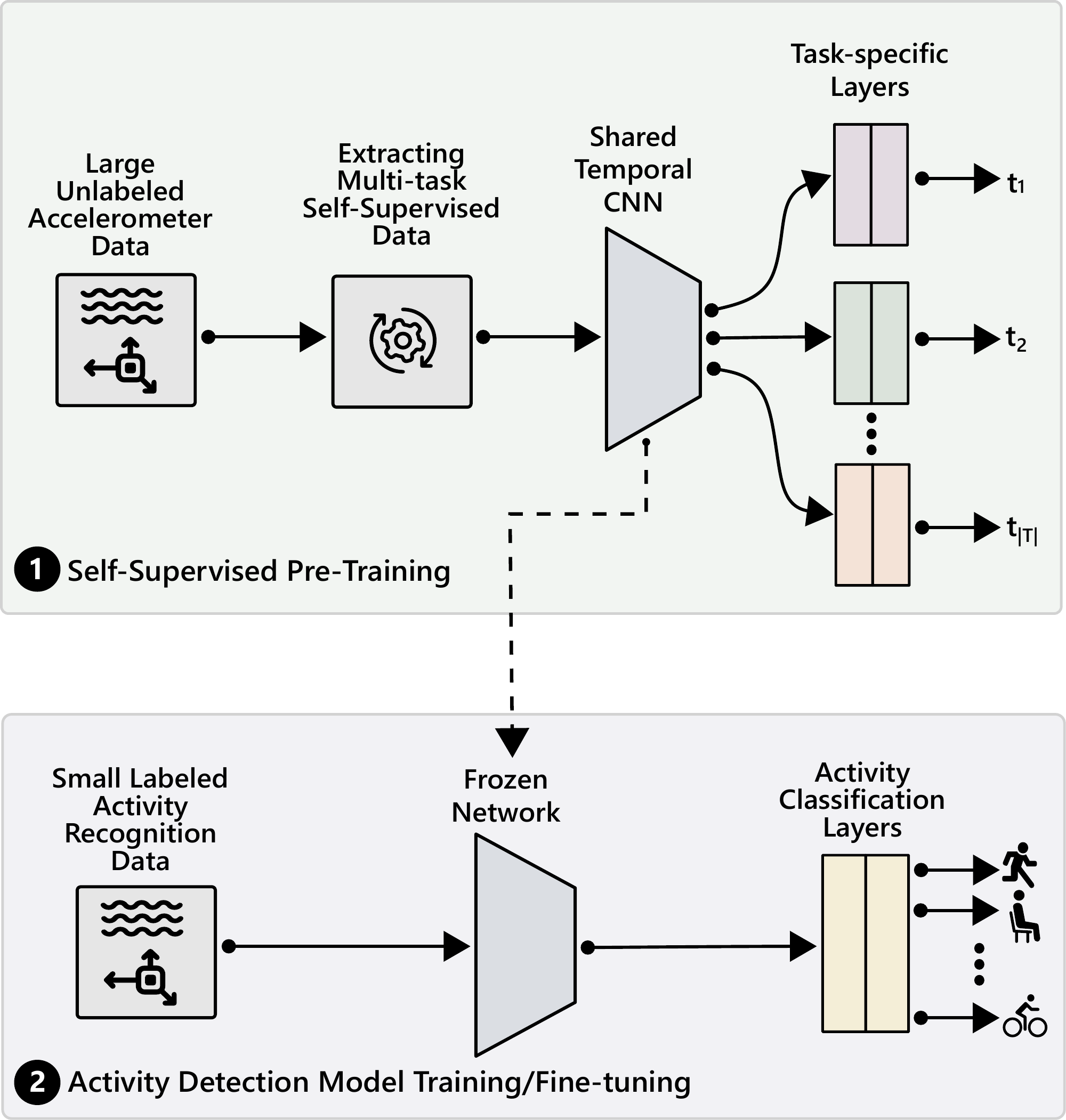}
\caption{Illustration of the proposed multi-task self-supervised approach for feature learning. \small{We train a temporal convolutional network for transformation recognition as a pretext task as shown in Step 1. The learned features are utilized by (or transferred to) the activity recognition model (Step 2) for improved detection rate with a small labeled dataset.}}
\label{fig:ssrl_overview}
\end{figure}

Our work is motivated by the success of jointly learning to solve multiple self-supervised tasks~\cite{doersch2017multi, caruana1997multitask} and we propose to learn accelerometer representations (i.e., features) by training a temporal convolutional neural network (CNN) to recognize the transformations applied to the raw input signal. Particularly, we utilize a set of signal transformations~\cite{um2017data, batista2011complexity} that are applied on each input signal in the datasets, which are then fed into the convolutional network along with the original data for learning to differentiate among them. In this simple formulation, a group of binary classification tasks (i.e., to recognize whether a transformation such as \textit{permutation}, \textit{scaling}, and \textit{channel shuffling} was applied on the original signal or not) act as surrogate tasks to provide a rich supervisory signal to the model. In order to extract highly generalizable features for the end-task of interest, it is essential to utilize transformations that exploit versatile invariances of the temporal data (further details are provided in Section~\ref{sec:approach}). To this end, we utilize eight transformations to train a multi-task network for simultaneously recognizing each of them. The visual illustration of the proposed approach is given in Figure~\ref{fig:ssrl_overview}. In the pre-training phase, the network consisting of a common trunk with a separate head for each task is trained on self-supervised data, and in the second step, the features learned by the shared layers are utilized by the HAR model. Importantly, we want to emphasize that in order for the convolutional network to recognize the transformations, it must learn to understand the core signal characteristics through acquiring knowledge of underlying differences in the accelerometer signals for various activity categories. We support this claim through an extensive evaluation of our method on six publicly available datasets in unsupervised, semi-supervised and transfer learning settings, where it achieves noticeable improvements in all the cases while not requiring manually labeled data for feature learning. 

The main contributions of this paper are:
\begin{itemize}
\item We propose to utilize self-supervision from large unlabeled data for human activity recognition.
\item We design a signal transformation recognition problem as a surrogate task for annotation-free supervision, which provides a strong training signal to the temporal convolutional network for learning generalizable features.
\item We demonstrate through extensive evaluation that the self-supervised features perform significantly better in the semi-supervised and transfer learning settings on several publicly available datasets. Moreover, we show that these features achieve performance that is superior to or comparable with the features learned via the fully-supervised approach (i.e., trained directly with activity labels).
\item We illustrate with SVCCA~\cite{raghu2017svcca}, saliency mapping~\cite{simonyan2013deep}, and t-SNE~\cite{maaten2008visualizing} visualizations that the features extracted via self-supervision are very similar to those learned by the fully-supervised network.
\item Our method substantially reduces the labeled data requirement, effectively narrowing the gap between unsupervised and supervised representation learning. 
\end{itemize} 

The paper is organized as follows. Section 2 provides an overview of related paradigms and methodologies as background information. Section 3 introduces the proposed self-supervised representation learning framework for HAR. Section 4 presents an evaluation of our framework on publicly available datasets. Section 5 gives an overview of the related work. Finally, Section 6 concludes the paper and lists future directions for research.

\section{Preliminaries}
In this section, we provide a brief overview of multiple learning paradigms, including multi-task, transfer, semi-supervised and importantly, representation learning. These either benefit or serve as fundamental building blocks of our self-supervised framework for representation extraction and robust HAR under various settings.

\subsection{Representation Learning}
Representation (feature) learning is concerned with automatically extracting useful information from the data that can be effectively used for an impending machine learning problem such as classification. In the past, most of the efforts were spent on developing (and manually engineering) feature extraction methods based on domain expertise to incorporate prior knowledge in the learning process. However, these methods are relatively limited as they rely on human creativity to come up with novel features and lack the power to capture underlying explanatory factors in the milieu of low-level sensory input. To overcome these limitations and to automate the discovery of disentangled features, neural networks based approaches have been widely utilized, such as autoencoders and their variants~\cite{baldi2012autoencoders}. Deep neural networks are composed of multiple (parameterized) non-linear transformations that are trained through a supervised or unsupervised objective function with the aim of yielding useful representations. These techniques have achieved indisputable empirical success across a broad spectrum of problems~\cite{krizhevsky2012imagenet, taigman2014deepface, sutskever2014sequence, mohamed2012acoustic, Hannun2019, aytar2016soundnet, li2016deep, radu2018multimodal} thanks to the increasing dataset sizes and computing power availability. Nevertheless, representation learning still stands as a fundamental problem in machine intelligence and is an active area of research (see~\cite{bengio2013representation} for a detailed survey).
 
\subsection{Multi-task Learning}
The goal of multi-task learning (MTL) is to enhance the learning efficiency and accuracy through simultaneously optimizing multiple objectives based on shared representations and exploiting relations among the tasks~\cite{caruana1997multitask}. It is widely utilized in several application domains within machine learning such as natural language processing~\cite{hashimoto2017joint}, computer vision~\cite{kendall2018multi}, audio sensing~\cite{georgiev2017low}, and well-being~\cite{saeed2017personalized}. In this learning setting, $T$ supervised tasks, each with a dataset $D^t = \{x_i^t, y_i^t\}_{i=1}^{m}$ and a separate cost function are made available. The multi-objective loss is then generally created through a weighted linear sum of the individual tasks' losses as:
\begin{equation} 
    \mathcal{L}_{Aggregated} = \sum_{t \in T} \psi_t \mathcal{L}_t 
    \label{eq:mtl}
\end{equation}

\noindent where $\psi_t$ is the task weight and $\mathcal{L}_t$ is a task-specific loss function. It is important to note that,  MTL itself does not impose any restriction on the loss type of an individual task. Therefore, unsupervised and supervised tasks or tasks having different cost functions can be conveniently combined for learning representations. 

\subsection{Transfer Learning}
Transfer learning aims to develop methods for preserving and leveraging previously acquired knowledge to accelerate the learning of novel tasks. In recent years, it has shown remarkable improvement in performance on several very challenging problems, especially in areas, where little-labeled data are available such as in natural language understanding, object recognition, and activity recognition~\cite{sharif2014cnn, morales2016deep, howard2018universal}. In this paradigm, the goal is to transfer (or reuse) the learned knowledge from a source domain $\mathcal{D}_{SRC}$ to a target domain $\mathcal{D}_{TRG}$. More precisely, consider domains $\mathcal{D}_{SRC}$ and $\mathcal{D}_{TRG}$ with learning tasks $t_{SRC}$ and $t_{TRG}$, respectively. The goal is to help improve the learning of a predictive function $f(.)$ in $t_{TRG}$ using the knowledge extracted from $\mathcal{D}_{SCR}$ and $t_{SRC}$, where $\mathcal{D}_{SRC} \neq \mathcal{D}_{TRG}$, and/or $t_{SRC} \neq t_{TRG}$, meaning that domains or tasks may be different. This learning formulation enables to develop a high-quality model under different knowledge transfer settings (such as features, instances, weights) from existing labeled data of some related task or domain. For a detailed review of transfer learning, we refer an interested reader to~\cite{pan2010survey}.

\subsection{Semi-supervised Learning}
Semi-supervised learning provides a compelling framework for leveraging unlabeled data in cases when labeled data collection is expensive. It has been repeatedly shown that given enough computational power and supervised data; deep neural networks can achieve human-level performance on a wide variety of problems~\cite{LeCun2015}. However, the curation of large-scale datasets is very costly and time-consuming as it either requires crowdsourcing or domain expertise such as in the case of medical imaging. Likewise, for several practical problems, it is simply not possible to create a large enough labeled dataset (e.g., due to privacy issues) to learn a model of reasonable accuracy. Semi-supervised learning algorithms offer a compelling alternative to fully-supervised methods for jointly learning from few labeled and a large number of unlabeled instances. More specifically, given a labeled training set of input-output pairs $(X, Y) \in D_L$ and unlabeled instance set, $X \in D_U$, the broad aim is to produce a predictive function $f_\theta(X)$ making use of not only $D_L$ but also the underlying structure in $D_U$, where $\theta$ represents the learnable parameters of the model. For a concise review and realistic evaluation of various deep learning based semi-supervised techniques, see~\cite{oliver2018realistic}. 

\subsection{Towards Self-supervision}
\begin{figure}[!htbp]
\centering
\includegraphics[width=\textwidth]{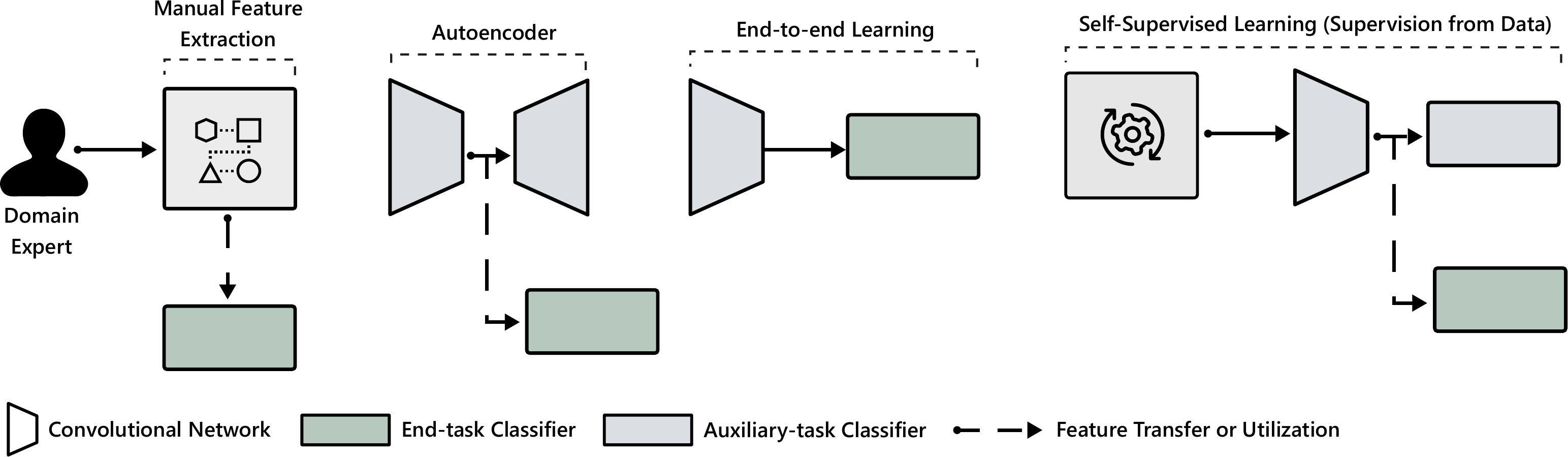}
\caption{Evolution of feature learning approaches from hand-crafted methods towards task discovery for self-supervision.}
\label{fig:rep_techniques}
\end{figure}

Deep learning has been increasingly used for end-to-end HAR with far superior performance that can be achieved through traditional machine learning methods~\cite{yang2015deep, hammerla2016deep, radu2018multimodal, morales2016deep, saeed2018synthesizing}. However, learning from very few labeled data, i.e. few-shot and semi-supervised learning is still an issue as large labeled datasets are required to train a model of sufficient quality. Similarly, the utilization of previously learned knowledge from related data (or task) to rapidly solve a comparable problem is not addressed very well by the existing methods (see Section~\ref{sec:related_work} for more details). In this paper, we explore self-supervised feature learning for HAR that effectively utilizes unlabeled data. The exciting field of self-supervision is concerned with extracting supervisory signals from data without requiring any human intervention. The evolution of feature extraction methods from hand-crafted features towards self-supervised representations is illustrated in Figure~\ref{fig:rep_techniques}. The input to each of the illustrated approaches is raw data, which is not shown for the sake of brevity. 

\section{Approach}
\label{sec:approach}
In this section, we present our self-supervised representation learning framework for HAR. First, we provide an overview of the methodology. Next, we discuss various learning tasks (i.e. transformation classification) and their benefits for generic features extraction from unlabeled data. Finally, we provide a detailed description of the network architecture, its implementation, and the optimization process. 
\subsection{Overview}
The objective of our work is to learn general-purpose sensor representations based on a temporal convolutional network in an unsupervised manner. To achieve this goal, we introduce a self-supervised deep network named \textit{Transformation Prediction Network} (TPN), which simultaneously learns to solve multiple (signal) transformation recognition tasks as shown in Figure~\ref{fig:ssrl_overview}. Specifically, the proposed multi-task TPN $M_\theta(.)$ is trained to produce estimates of the transformations applied to the raw input signal. We define a set of $|T|$ distinct transformations (or tasks) $T = \{J_t(.)\}_{t \in T}$, where $J_t(.)$ is a function that applies a particular signal alteration technique $t$ to the temporal sequence $x \in \mathbb{R}^{(N, C)}$ to yield a transformed version of the signal $J_t(x)$. The network $M_\theta(.)$ that has a common trunk and individual head for each task, it takes an input sequence and produces a probability of the signal being a transformed version of the original, i.e. $P(J_t|x) = M_\theta(x)$. Note, that given a set of unlabeled signals (e.g. of accelerometer), we can automatically construct a self-supervised labeled dataset $D = \{  \{(J_t(x_i), True), (x_i, False)\}_{t \in T}  \}_{i=1}^{m}$. Hence, given this set of $m$ training instances, the multi-task self-supervised training objective that a model must learn to solve is: 
\begin{equation} 
    \min_\theta \sum_{t \in T}^{} \psi_t \Big[- \frac{1}{m_t} \sum_{i=1}^{m_t} (y_i^t\log(M_\theta(x_i^t)) + (1 - y_i^t)\log(1 - M_\theta(x_i^t))) \Big]
    \label{eq:crossentropy_loss}
\end{equation}

\noindent where $M_\theta(x^t)$ is the predicted probability of $x$ being a transformed version $t$ and $\theta$ are the learnable parameters of the network. $m_t$ represents the number of instances for a task (which can vary but are equal in our case) and ${\psi}_t$ is the loss-weight of task $t$.

We emphasize that, although the network has a separate layer to differentiate between original and each of the $T$ transformations it can be extended in a straight-forward manner to recognize multiple transformations applied to the same input signal or for multi-label classification. In the following subsection, we explain the types of signal transformations that are used in this work. 

\subsection{Self-supervised Tasks: \textit{Signal Transformations}}
The aforementioned formulation requires the signal transformations $J$ to define a multi-task classification that enables the convolutional model to learn disentangled semantic representations useful for down-stream tasks, e.g. activity detection. We aimed for conceptually simple, yet diverse tasks to possibly cover several invariances that commonly arise in temporal data~\cite{batista2011complexity}. Intuitively, a diverse set of tasks should lead to a broad spectrum of features, which are more likely to span the feature-space domain needed for a general understanding of the signal's characteristics. In this work, we propose to utilize eight straight-forward signal transformations (i.e. $|T| = 8$)~\cite{batista2011complexity, um2017data} for the self-supervision of a network. More specifically, when transformations are applied on an input signal $x$, they result in eight variants of $x$. As mentioned earlier, the temporal convolutional model is then trained jointly on all the tasks' data to solve a problem of transformation recognition, which allows the model to extract high-level abstractions from the raw input sequence. The transformations utilized in this work are summarized below:

 \begin{itemize}
     \item \textbf{Noised:}  Given sensor readings of a fixed length, a possible transformation is the addition of random noise (or jitter) in the original signal. Heterogeneity of device sensors, software, and other hardware can cause variations (noisy samples) in the produced data. A model that is robust against noise will generalize better as it learns features that are invariant to minor corruption in the signal.
    
     \item \textbf{Scaled:} A transformation that changes the magnitude of the samples within a window through multiplying with a randomly selected scalar. A model capable of handling scaled signals produces better representations as it becomes invariant to amplitude and offset invariances. 
    
     \item \textbf{Rotated:} Robustness against arbitrary rotations applied on the input signal can achieve sensor-placement (orientation) invariance. This transformation inverts the sample signs (without changing the associated class-label) as frequently happens if the sensor (or device) is, for example, held upside down. 

     \item \textbf{Negated:} This simple transformation is an instance of both \textit{scaled} (scaling by $-1$) and \textit{rotated} transformations. It negates samples within a time window, resulting in a vertical flip or a mirror image of the input signal. 

     \item \textbf{Horizontally Flipped:} This transformation reverses the samples along the time-dimension, resulting in a complete mirror image of an original signal as if it were evolved in the opposite time direction. 
    
     \item \textbf{Permuted:} This transformation randomly perturbs the events within a temporal window through slicing and swapping different segments of the time-series to generate a new one, hence, facilitating the model to develop permutation invariance properties. 
    
     \item \textbf{Time-Warped:} This transformation locally stretches or warps a time-series through a smooth distortion of time intervals between the values (also known as local scaling).
        
     \item \textbf{Channel-Shuffled:} For a multi-component signal such as a triaxial accelerometer, this transformation randomly shuffles the axial dimensions.  
\end{itemize}

There are several benefits of utilizing transformations recognition as auxiliary tasks for feature extraction from unlabeled data.

\textbf{Enabling the learning of generic representations:} The primary motivation is that the above-defined pretext tasks enable the network to capture the core signal characteristics. More specifically, for the TPN to successfully recognize if the signal is transformed or not, it must learn to detect high-level semantics, sensor behavior under different device placements, time-shift of the events, varying amplitudes, and robustness against sensor noise, thus, contributing to solving the ultimate task of HAR.

\textbf{Task diversification and elimination of low-level input artifacts:} A clear advantage of using multiple self-supervised tasks as opposed to a single one is that it will lead to a more diverse set of features that are invariant to low-level artifacts of the signals. Had we chosen to utilize signal reconstruction, e.g. with autoencoders, this would learn to compress the input, but due to a weak supervisory signal (as compared to self-supervision), it may discover trivial features with no practical value for the activity recognition or any other task of interest. We compare our approach against other methods in section~\ref{sec:results}.

\textbf{Transferring knowledge:} Furthermore, with our approach, the unlabeled sensor data that are produced in huge quantity can be effectively utilized with no human intervention to pre-train a network that is suitable for semi-supervised and transfer learning settings. It is particularly of high value for training networks in a real-world setting, where very little or no supervision is available to learn a model of sufficient quality from scratch.

\textbf{Other benefits:} Our self-supervised method has numerous other benefits. It has an equivalent computational cost to supervised learning but with better convergence accuracy, making it a suitable candidate for continuous unsupervised representation learning in-the-wild. Moreover, our technique neither requires a sophisticated pre-processing (apart from z-normalization) nor needs a specialized architecture (which also requires labeled data) to exploit invariances. We will show in Section~\ref{sec:results} through extensive evaluation that the self-supervised models learn useful representations and dramatically improve performance over other learning strategies. Despite the simplicity of the proposed scheme, it allows utilizing data collected through a wide variety of devices from a diverse set of users. 

\subsection{Network Architecture and Implementation}
\label{sec:nai}
\begin{figure}[!htbp]
\centering
\includegraphics[width=3.8in,angle=90]{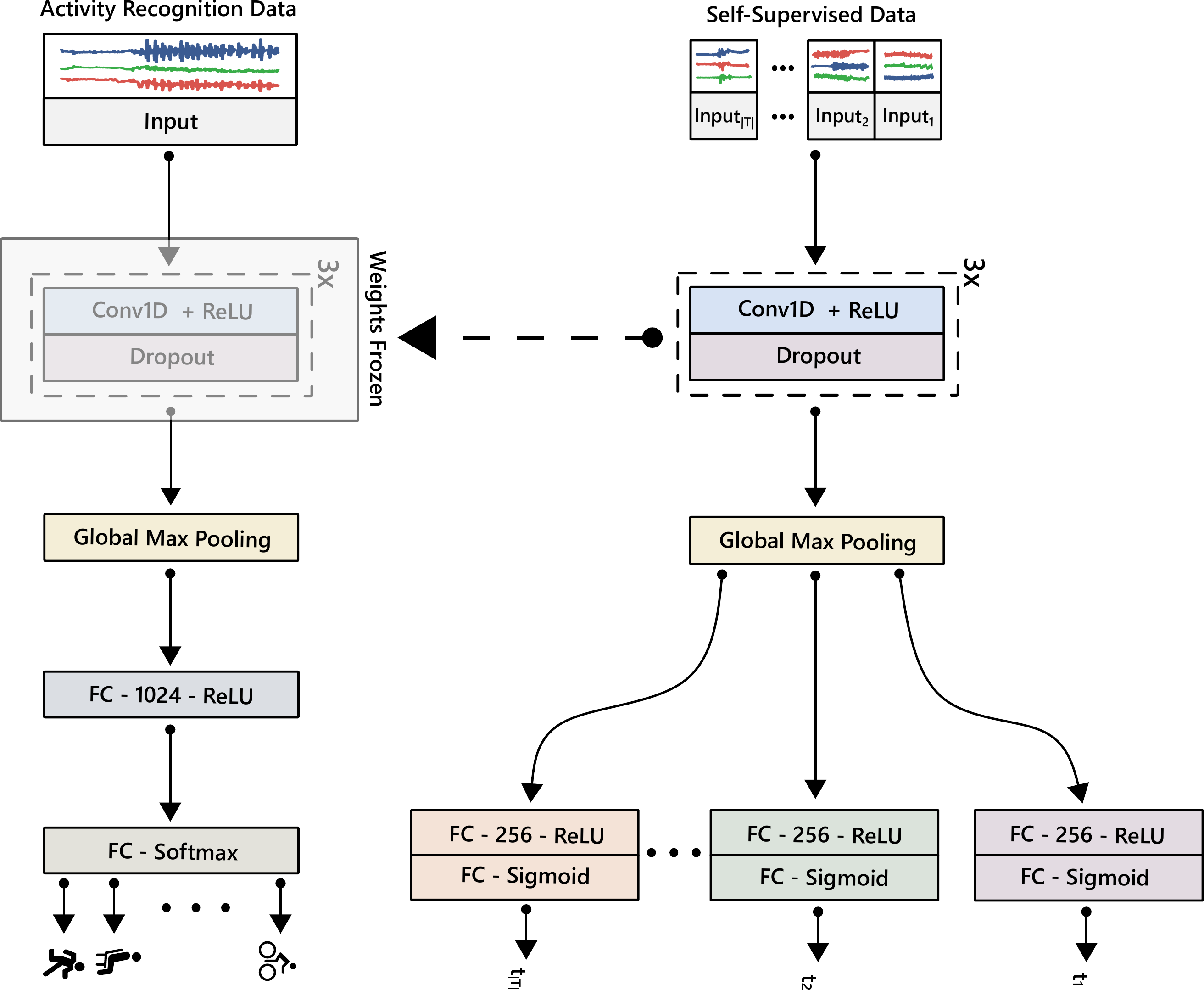}
\caption{Detailed architectural specification of transformation prediction and activity recognition networks. \small{We propose a framework for self-supervised representation learning from unlabeled sensor data (such as an accelerometer). Various signal transformations are utilized to establish supervisory tasks, and the network is trained to differentiate between an original and transformed version of the input. The three blocks of \textit{Conv + ReLU} and \textit{Dropout} layers, which is followed by a \textit{Global Max Pooling} are similar across both networks. However, the multi-task model has a separate head for each task. Likewise, the activity recognizer has an additional densely connected layer. The TPN is pre-trained on self-supervised data, and the learned weights are transferred (depicted by a dashed arrow) and kept frozen to the lower model, which is then trained to detect various activities.}}
\label{fig:architecture}
\end{figure}

We implement the TPN $M_\theta(.)$ as a multi-branch temporal convolutional neural network with a common trunk (shared layers) and a distinct head (private layers) for each task with a separate loss function. Hard parameter sharing is employed between all the task-specific layers to encourage strong weight utilization from the trunk. Figure~\ref{fig:architecture} illustrates the TPN containing three $1$D convolutional layers consisting of $32$, $64$, and $96$ feature maps with kernel sizes of $24$, $16$ and $8$ respectively, and having a stride of $1$. Dropout is used after each of the layers with a rate of $0.1$, and L$2$ regularization is applied with a rate of $0.0001$. Global max pooling is used after the last convolution layer to aggregate high-level discriminative features. Moreover, each task-specific layer is comprised of a fully-connected layer of $256$ hidden units followed by a sigmoidal output layer for binary classification. We use $ReLU$ as non-linearity in all the layers (except the output) and train a network with Adam optimizer~\cite{kingma2014adam} for a maximum of 30 epochs with a learning rate of $0.0003$, unless stated otherwise. Furthermore, the activity recognition model has a similar architecture to the TPN except for a fully-connected layer that consists of $1024$ hidden units followed by a softmax output layer with units depending on the activity detection task under consideration. Additionally, during training of this model, we apply early-stopping, if the network fully converges on the training set to avoid overfitting. 

The motivation for keeping the TPN architecture simple arises from the fact that we want to show the performance gain does not come from the number of parameters (or layers) or due to the utilization of other sophisticated techniques such as batch normalization but the improvement is due to self-supervised pre-training. Likewise, the choice of multi-task learning setting, where each task has an additional private layer manifests in letting the model push pretext task-specific features to the last layers and let the initial layers extract generic representations that are important for a wide variety of end-tasks. Moreover, our architectural specification allows for a straightforward extension to add other related tasks, if needed, such as input reconstruction. Although, we do not explore applying multiple transformations to the same sequence or train models for their recognition the network design is intrinsically capable of performing this multi-label classification task.

\begin{algorithm}[htbp]
\KwIn{Unlabeled instance set $D_U$, labeled dataset $D_L$, task-specific weights $\psi$, numbers of epochs $E_M$ and $E_C$}
\KwOut{Self-supervised network $M$, activity classification model $C$ with $n$ classes}
initialize $(X, Y)_1, \ldots, (X, Y)_T$ to hold instance-label pairs for multiple tasks in $T$\\
initialize $M$ with parameters $\theta_M$ and $C$ with parameters $\theta_C$\\

\tcp{Labeled data generation for self-supervision}
\For{each instance $x$ in $D_U$}
{
  \For{each transformation $t \in T$}
  {
    Insert $(x$, $False)$ to $(X, Y)_t$ and $(J_t(x)$, $True)$ to $(X, Y)_t$\\
  }
}
\For{each epoch $e_m$ from $1$ to $E_M$}
{
    Randomly sample a mini-batch of $m$ samples for all tasks $\{(X, Y)_1, \ldots, (X, Y)_T\}$ \\
    Update $\theta_M$ by descending along its gradient $\nabla_{\theta_{M}} \Bigg[  \sum_{t \in T} \psi_t \Big[ - \frac{1}{m_t} \sum_{i=1}^{m_t} (y_i^t\log(M_\theta(x_i^t)) + (1 - y_i^t)\log(1 - M_\theta(x_i^t))) + \beta \left\lVert \theta\right\rVert^2 \Big]  \Bigg]$
}
Assign learned parameters from $\theta_M^{1 \ldots L}$ to $\theta_C^{1 \ldots L}$ \\
Keep the transferred weights $\theta_C^{1 \ldots L}$ of network $C$ frozen \\
\For{each epoch $e_c$ from $1$ to $E_C$}
{
    Randomly sample a mini-batch of $m$ labeled activity recognition samples $D_L$\\
    Update $\theta_C$ by descending along its gradient  $\nabla_{\theta_{C}} \Bigg[ - \big( \frac{1}{m} \sum^m_{i=1} \sum^n_{k=1} \ y_{i,k} \ \log(C_\theta(x_{i})) \big)  + \beta (\left\lVert \theta\right\rVert^2)      \Bigg]$
}

Gradient-based updates can use any standard gradient-based learning technique. We used Adam~\cite{kingma2014adam} in all our experiments.

\caption{Multi-task Self-Supervised Learning}
\label{alg:one}
\end{algorithm}

Our training process is summarized in Algorithm~\ref{alg:one}. For every instance, we first generate transformed versions of a signal for the self-supervised pre-training of the network. At each training iteration of the TPN model, we feed the data from all tasks simultaneously, and the overall loss is calculated as a weighted sum of the losses of different tasks. Once pre-training converges, we transfer the weights of convolutional layers from model $M_\theta$ to an activity recognition network $C_\theta$ for learning the final supervised task. Here, either all the transferred layers are kept frozen, or the last convolutional layer is fine-tuned depending on the learning paradigm. Figure~\ref{fig:architecture} depicts this process graphically, where shaded convolutional layers represent frozen weights, while others are either trained from scratch or optimized further on the end-task. To avoid ambiguity, in the experiment section, we explicitly mention when the results are from a fully-supervised or self-supervised (including fine-tuned) network.   

\section{Evaluation}
In this section, we conduct an extensive evaluation of our approach on several publicly available datasets for human activity recognition (HAR) in order to determine the quality of learned representations, transferability of the features, and benefits of this in the low-data regime. The self-supervised tasks (i.e., transformation predictions) are utilized for learning rich sensor representations that are suitable for an end-task. We emphasize that achieving high performance on these surrogate tasks is \textbf{not} our focus. 

\begin{table}[htbp]
\caption{Summary of datasets used in our evaluation. \small{These datasets are selected based on the diversity of participants, device types and activity classes. Further details on the pre-processing of each data source and the number of users utilized are discussed in Section~\ref{sec:dataset}.}}
\begin{tabular}{lcc}
\hline
\multicolumn{1}{l}{\textbf{Dataset}} & \textbf{No. of users} & \textbf{No. of activity classes} \\ \hline
HHAR                                 & 9                     & 6                       \\
UniMiB                               & 30                    & 9                       \\
UCI HAR                              & 30                    & 6                       \\
MobiAct                              & 67                    & 11                      \\
WISDM                                & 36                    & 6                       \\
MotionSense                          & 24                    & 6                       \\ \hline
\end{tabular}
\label{tab:dataset}
\end{table}

\subsection{Datasets}
\label{sec:dataset}
We consider six publicly available datasets to cover a wide variety of device types, data collection protocols, and activity recognition tasks performed with smartphones in different environments. Some important aspects of the data are summarized in Table~\ref{tab:dataset}. Below, we give brief descriptions of every dataset summarizing its key points. 

\subsubsection{HHAR}
The Heterogeneity Human Activity Recognition (HHAR) dataset~\cite{stisen2015smart} contains signals from two sensors (accelerometer and gyroscope) of smartphones and smartwatches for $6$ different activities, i.e. biking, sitting, standing, walking, stairs-up and stairs-down. The $9$ participants executed a scripted set of activities for $5$ minutes to get equal class distribution. The subjects had $8$ smartphones in a tight pouch carried around their waist and $4$ smartwatches, $2$ worn on each arm. In total, they used $36$ different smart devices of $13$ models from $4$ manufacturers to cover a broad range of devices for sampling rate heterogeneity analysis. The sampling rate of signals varied significantly across phones with values between $50$-$200$Hz.  

\subsubsection{UniMiB}
This dataset~\cite{micucci2017unimib} contains triaxial accelerometer signals collected from a Samsung Galaxy Nexus smartphone at $50$Hz. Thirty subjects participated in the data collection process forming a diverse sample of the population with different height, weight, age, and gender. The subject placed the device in her trouser's front left pocket for a partial duration and in the right pocket for the remainder of the experiment. We utilized the data of $9$ activities of daily living (i.e., standing up from sitting, standing up from lying, walking, running, upstairs, jumping, downstairs, lying down from sitting, sitting) in this paper. 

\subsubsection{UCI HAR}
The UCI HAR dataset~\cite{anguita2013public} is obtained from a group of $30$ volunteers with a waist-mounted Samsung Galaxy S$2$ smartphone. The accelerometer and gyroscope signals are collected at $50$Hz when subjects performed the following six activities: standing, sitting, laying down, walking, downstairs and upstairs. 

\subsubsection{MobiAct}
The MobiAct\footnote{second release} dataset~\cite{chatzaki2016human} contains signals from a smartphone's inertial sensors (accelerometer, gyroscope, and orientation) for $11$ different activities of daily living and $4$ types of falls. It is collected with a Samsung Galaxy S$3$ smartphone from $66$ participants of different gender, age group, and weight through more than $3200$ trials. The device is placed in a trouser's pocket freely selected by the subject in any random orientation to capture everyday usage of the phone. We used the data from $61$ participants who have data samples for any of the following $11$ activities: sitting, walking, jogging, jumping, stairs up, stairs down, stand to sit, sitting on a chair, sit to stand, car step-in, and car step-out. 

\subsubsection{WISDM}
The dataset from the Wireless Sensor and Data Mining (WISDM) project~\cite{kwapisz2011activity} was collected in a controlled study from $29$ volunteers, who carried the cell phone in their pockets.  The data were recorded for $6$ different activities (i.e., sit, stand, walk, jog, ascend stairs, descend stairs) via an app developed for an Android phone. The accelerometer signal was acquired every $50$ms (sampling rate of $20$Hz). We use the data of all the users available in the raw data file with user ids ranging from $1$ to $36$. 

\subsubsection{MotionSense}
The MotionSense dataset~\cite{malekzadeh2018protecting} comprises an accelerometer, gyroscope, and altitude data from $24$ participants of varying age, gender, weight, and height. It was collected using an iPhone$6$s, which is kept in the user's front pocket. The subjects performed $6$ different activities (i.e., walking, jogging, downstairs, upstairs, sitting, and standing.) in $15$ trials under similar environments and conditions. The study aimed to infer physical and demographics attributes from time-series data in addition to the detection of activities.       

\subsection{Data Preparation and Assessment Strategy}
We applied minimal pre-processing on the accelerometer signals as deep neural networks are very good at learning abstract representations directly from raw data~\cite{LeCun2015}. We segmented the signals into fixed size windows that have $400$ samples with $50\%$ overlap, for all the datasets under consideration. The appropriate window size is a task-specific parameter and could be tuned or chosen based on prior knowledge for improved performance. Here, we utilize the same window size based on earlier exploration across datasets and to keep experimental evaluation impartial towards the effect of this hyper-parameter. Next, we divide each dataset into training and test sets through randomly selecting $20-30\%$ of the users for testing and the rest for training and validation; depending on the dataset size. We used the ceiling function to select number of users, e.g. from HHAR dataset $3$ users are used for evaluation out of $9$. The training set users' data are further divided into $80\%$ for training the network and $20\%$ for validation and hyper-parameter tuning. Importantly, we also evaluate our models through user-split based $5$-folds cross-validation, wherever it is appropriate. Finally, we normalize the data by applying z-normalization with summary statistics calculated from the training set. We generate self-supervised data from an unlabeled training set that is produced as a result of the processing as mentioned earlier. We utilize the data generation procedure as explained earlier in Section~\ref{sec:nai}.

Furthermore, due to the large size of the \textit{HHAR} dataset and in order to reduce computational load, we randomly sample $4000$ instances from each users' data to produce transformed signals. Likewise, in the case of \textit{UniMiB} because of its relatively small size, we generate $5$ times more transformed instances. We evaluate the performance with Cohen's kappa, a weighted version of precision, recall and f-score metrics to be robust against inherent imbalanced nature of the datasets. It is important to highlight that, \textit{we use a network architecture with the same configuration across the datasets to evaluate models' performance in order to highlight improvement is indeed due to self-supervision and not due to architectural modifications}.  

\subsection{Results}
\label{sec:results}
\subsubsection{Quantifying the Quality of Learned Feature Hierarchies}
\begin{figure}[!htbp]
\centering
\includegraphics[width=5in]{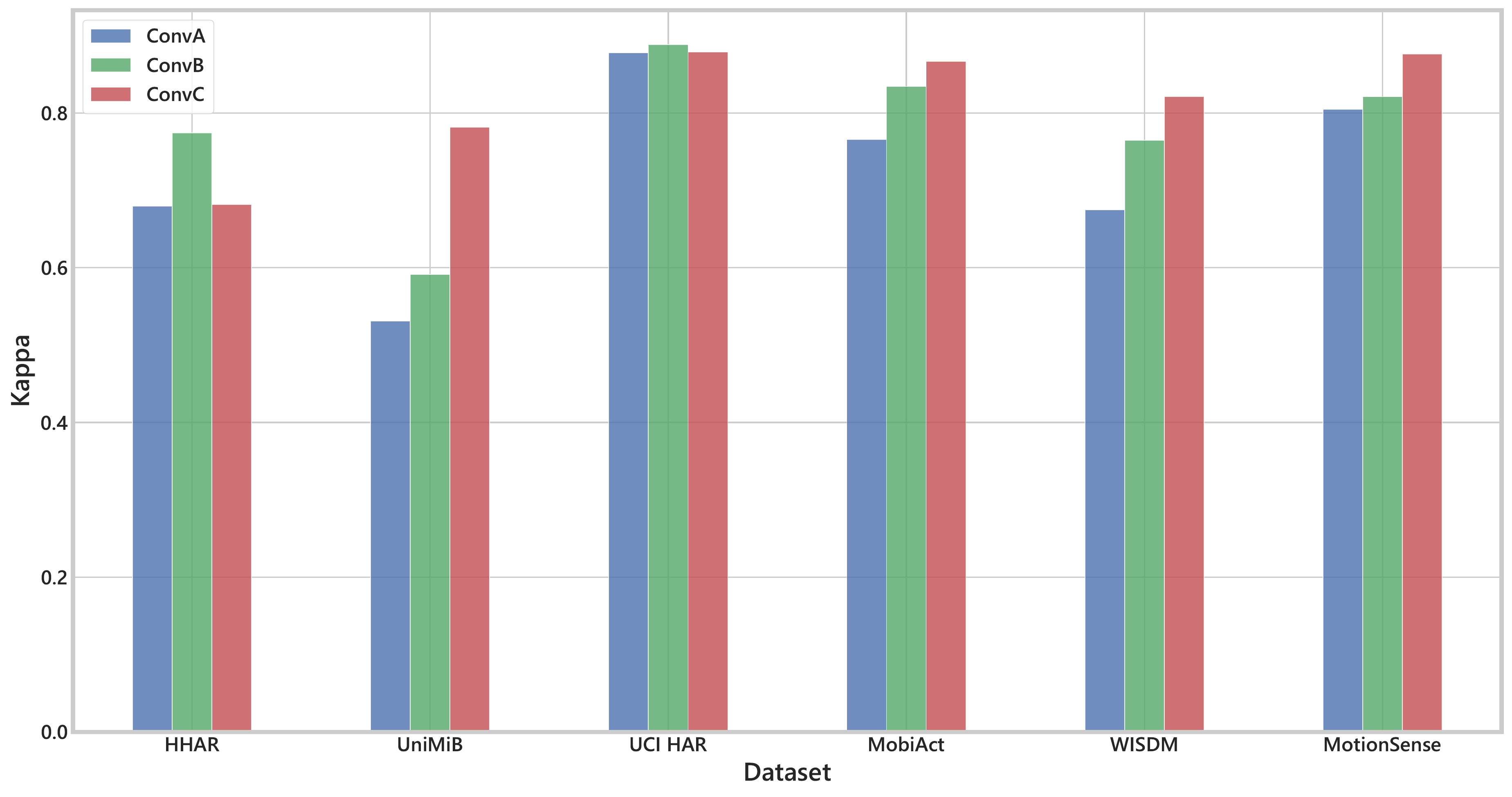}
\caption{Evaluation of activity classification performance using the features learned based on self-supervision (per layer). \small{We train an activity classifier on-top of each of the temporal convolution blocks ($ConvA$, $ConvB$, and $ConvC$) that are pre-trained with self-supervision. The reported results are averaged over $10$ independent runs (i.e., training an activity classifier from scratch). $ConvA$, $ConvB$, and $ConvC$ have $32$, $64$, and $96$ feature maps, respectively.}}
\label{fig:layerwise_evaluation}
\end{figure}

We first evaluate our approach to determine the quality of learned representations versus the model depth (i.e., the layer number from which the features come). This analysis helps in understanding \textit{whether the features coming from different layers vary in quality concerning their performance on an end-task and if so, which layer should be utilized for this purpose}. To this end, we first pre-train our TPN in a self-supervised manner and learn classifiers on top of $ConvA$, $ConvB$, and $ConvC$ layers independently, for several activity recognition datasets. These classifiers (see Figure~\ref{fig:architecture}) are trained in a supervised way while keeping the learned features fixed during the optimization process. Figure~\ref{fig:layerwise_evaluation} provides kappa values on test sets averaged across 10-independent runs to be robust against differences in weight initializations of the classifiers. We observe that for a majority of the datasets the model performance improves with increasing depth apart from \textit{HHAR}, where features from $ConvB$ layer results in improved detection rate with a kappa of $0.774$ compared to $0.679$ of $ConvC$. It may be because the representation of the last layer starts to become too specific on the transformation prediction task or it may also be because we did not utilize the entire dataset for the self-supervision. To be consistent, in the subsequent experiments we used features from the last convolutional layer for all the considered datasets. For a new task or recognition problem, we recommend performing a similar analysis to identify layer/block of the network that gives optimal results on the particular dataset. 

\subsubsection{Comparison against Fully-Supervised and Unsupervised Approaches}
In this subsection, we assess our self-supervised representations learned with TPN against other unsupervised and fully-supervised techniques for feature learning. Table~\ref{tab:fs_ss_eval} summarizes the results with respect to four evaluation metrics (namely, precision, recall, f-score, and kappa) for 10-independent runs on the six datasets described earlier. For the \textit{Random Init.} entries, we keep the convolutional network layers frozen during optimization and train only a classifier in a supervised manner. Likewise, for an \textit{Autoencoder}, we keep the network architecture the same and pre-train it in an unsupervised way. Afterward, the weights of the encoder are kept frozen, and a classifier is trained on top as usual. The \textit{Self-Supervised} entries show the result of the convolutional network pre-trained with our proposed method, where a classifier is trained on top of the frozen network in a supervised fashion. Furthermore, \textit{Self-Supervised (FT)} entries highlight the performance of the network trained with self-supervision but the last convolution layer, i.e. $ConvC$ is fine-tuned along with a classifier during training on the activity recognition task. Training an activity classification model on top of randomly initialized convolutional layers poorly performs as expected, which is evidence that the performance improvement is not only because of the activity classifier. These results are followed by a widely used unsupervised learning method, i.e. an autoencoder. The self-supervised technique outperforms existing methods and achieves results that are on par with the fully-supervised model. It is important to note that, for our proposed technique, only the classifier layers are randomly initialized and trained with activity specific labels (the rest is transferred from the self-supervised network). We also observe that fine-tuning the last convolutional layer further improves the classification performance of the down-stream tasks on several datasets such as \textit{UniMiB}, \textit{HHAR}, \textit{MobiAct}, and \textit{UCI HAR}. The results show that TPN can learn highly generalizable representations, thus reducing the performance gap of feature learning with the (end-to-end) supervised case. For a more rigorous evaluation, we also performed $5$-folds (user split based) cross-validation for every method on all the datasets. The results are provided in Table~\ref{tab_appendix:fs_ss_eval} of the appendix, which also shows that the self-supervised method reduces the performance gap with the supervised setting.   

\begin{table}[!ht]
   \caption{Task Generalization: Evaluating self-supervised representations for activity recognition. \small{We compare the proposed self-supervised method for representation learning with fully-supervised and unsupervised approaches. We use the same architecture across all the experiments. The self-supervised TPN is trained to recognize transformations applied on the input signal while the activity classifier is trained on top of these learned features where \textit{Self-Supervised (FT)} entry provides results when the last convolution layer is fine-tuned. The \textit{Random Init.} entries present results when the convolution layers are randomly initialized and kept frozen during the training of the classifier. The results reported are averaged over $10$ independent runs to be robust against variations in the weight initialization and the optimization process.}}
   \centering
   \subfloat[HHAR]{
     \scriptsize
     \centering
     \begin{tabular}{l|cccc}
& \textbf{P}    & \textbf{R}    & \textbf{F}    & \textbf{K}    \\ \hline
Random Init.              & 0.3882$\pm$0.0557 & 0.3101$\pm$0.0409 & 0.2141$\pm$0.0404 & 0.1742$\pm$0.0488 \\
Supervised           & 0.7624$\pm$0.0312 & 0.7353$\pm$0.0308 & 0.7276$\pm$0.0297 & 0.6816$\pm$0.0371 \\
Autoencoder          & 0.7317$\pm$0.0451 & 0.6657$\pm$0.0663 & 0.6585$\pm$0.0724 & 0.5994$\pm$0.0784 \\
Self-Supervised      & 0.7985$\pm$0.0155 & 0.777$\pm$0.0199  & 0.7666$\pm$0.0234 & 0.731$\pm$0.0243  \\
Self-Supervised (FT) & 0.8218$\pm$0.0256 & 0.797$\pm$0.0211  & 0.7862$\pm$0.0187 & \textbf{0.7555}$\pm$\textbf{0.025}
\end{tabular}
   } \\
   \subfloat[UniMiB]{
     \scriptsize
     \centering
    \begin{tabular}{l|cccc}
 & \textbf{P}    & \textbf{R}    & \textbf{F}    & \textbf{K}    \\ \hline
Random Init.              & 0.4256$\pm$0.0468 & 0.3546$\pm$0.037  & 0.2775$\pm$0.0491 & 0.2243$\pm$0.0474 \\
Supervised           & 0.8276$\pm$0.0148 & 0.8096$\pm$0.0266 & 0.8097$\pm$0.0248 & 0.7815$\pm$0.0299 \\
Autoencoder          & 0.5922$\pm$0.0191 & 0.5557$\pm$0.0232 & 0.5376$\pm$0.0339 & 0.4824$\pm$0.0275 \\
Self-Supervised      & 0.8133$\pm$0.0077 & 0.7954$\pm$0.014  & 0.7929$\pm$0.016  & 0.7642$\pm$0.0162 \\
Self-Supervised (FT) & 0.8506$\pm$0.007  & 0.8432$\pm$0.0049 & 0.8425$\pm$0.0054 & \textbf{0.8197}$\pm$\textbf{0.005}
\end{tabular}
 } \\
   \subfloat[UCI HAR]{
     \scriptsize
     \centering
     \begin{tabular}{l|cccc}
& \textbf{P}    & \textbf{R}    & \textbf{F}    & \textbf{K}    \\ \hline
Random Init.               & 0.6189$\pm$0.0648 & 0.4392$\pm$0.0692 & 0.3713$\pm$0.0952 & 0.3133$\pm$0.0866 \\
Supervised           & 0.9059$\pm$0.0133 & 0.8998$\pm$0.0139 & 0.8981$\pm$0.0148 & 0.8789$\pm$0.0168 \\
Autoencoder          & 0.8314$\pm$0.0590 & 0.7877$\pm$0.1112 & 0.7772$\pm$0.1306 & 0.7425$\pm$0.1359 \\
Self-Supervised      & 0.9100$\pm$0.0081 & 0.9011$\pm$0.0139 & 0.8987$\pm$0.0155 & \textbf{0.8803}$\pm$\textbf{0.0169} \\
Self-Supervised (FT) & 0.9057$\pm$0.0121 & 0.897$\pm$0.0185  & 0.8946$\pm$0.019  & 0.8754$\pm$0.0222
\end{tabular}
   } \\
   \subfloat[MobiAct]{
     \scriptsize
     \centering
\begin{tabular}{l|cccc}
 & \textbf{P}    & \textbf{R}    & \textbf{F}    & \textbf{K}    \\ \hline
Random Init.              & 0.4749$\pm$0.1528&0.3452$\pm$0.1128&0.2813$\pm$0.0982&0.1915$\pm$0.1017\\
Supervised           & 0.908$\pm$0.0066&0.895$\pm$0.0167&0.8975$\pm$0.0133&0.8665$\pm$0.0202\\
Autoencoder          & 0.7493$\pm$0.0328&0.7581$\pm$0.0354&0.7293$\pm$0.0452&0.6772$\pm$0.0517\\
Self-Supervised      & 0.9095$\pm$0.0035&0.9059$\pm$0.0059&0.906$\pm$0.0053&0.8795$\pm$0.0073\\
Self-Supervised (FT) & 0.9194$\pm$0.0057&0.9102$\pm$0.0114&0.9117$\pm$0.0093&\textbf{0.8855}$\pm$\textbf{0.014}
\end{tabular}
   } \\
  \subfloat[WISDM]{
     \scriptsize
     \centering
\begin{tabular}{l|cccc}
& \textbf{P}    & \textbf{R}    & \textbf{F}    & \textbf{K}    \\ \hline
Random Init.               & 0.5942$\pm$0.0599 & 0.3543$\pm$0.077  & 0.358$\pm$0.0837  & 0.2224$\pm$0.0656 \\
Supervised           & 0.9024$\pm$0.0076 & 0.8657$\pm$0.0206 & 0.8764$\pm$0.0168 & 0.8211$\pm$0.0258 \\
Autoencoder          & 0.6561$\pm$0.2775 & 0.6631$\pm$0.1623 & 0.6358$\pm$0.2355 & 0.5106$\pm$0.288  \\
Self-Supervised      & 0.8894$\pm$0.0096&0.8484$\pm$0.0269&0.8593$\pm$0.0225&0.7986$\pm$0.0334  \\
Self-Supervised (FT) & 0.8999$\pm$0.0111&0.8568$\pm$0.0375&0.8686$\pm$0.0314&0.8106$\pm$0.0466
\end{tabular}
  } \\
  \subfloat[MotionSense]{
     \scriptsize
     \centering
\begin{tabular}{l|cccc}
& \textbf{P}    & \textbf{R}    & \textbf{F}    & \textbf{K}    \\ \hline
Random Init.               & 0.5999$\pm$0.0956 & 0.5029$\pm$0.0931 & 0.4681$\pm$0.1105 & 0.376$\pm$0.1176  \\
Supervised           & 0.9164$\pm$0.0053 & 0.8993$\pm$0.0091 & 0.9027$\pm$0.0085 & 0.8763$\pm$0.011  \\
Autoencoder          & 0.8255$\pm$0.0132 & 0.8116$\pm$0.0195 & 0.8109$\pm$0.0169 & 0.7659$\pm$0.0226 \\
Self-Supervised      & 0.8979$\pm$0.0073 & 0.8856$\pm$0.0087 & 0.8864$\pm$0.0083 & 0.8589$\pm$0.0106 \\
Self-Supervised (FT) & 0.9153$\pm$0.0088 & 0.8979$\pm$0.0092 & 0.9005$\pm$0.0094 & 0.8744$\pm$0.0112
\end{tabular}
  }
\label{tab:fs_ss_eval}
\end{table}

\subsubsection{Assessment of Individual Self-Supervised Tasks in Contrast with Multiple Tasks}
\begin{figure}[htbp]
\centering
\subfloat{\includegraphics[width=.32\textwidth]{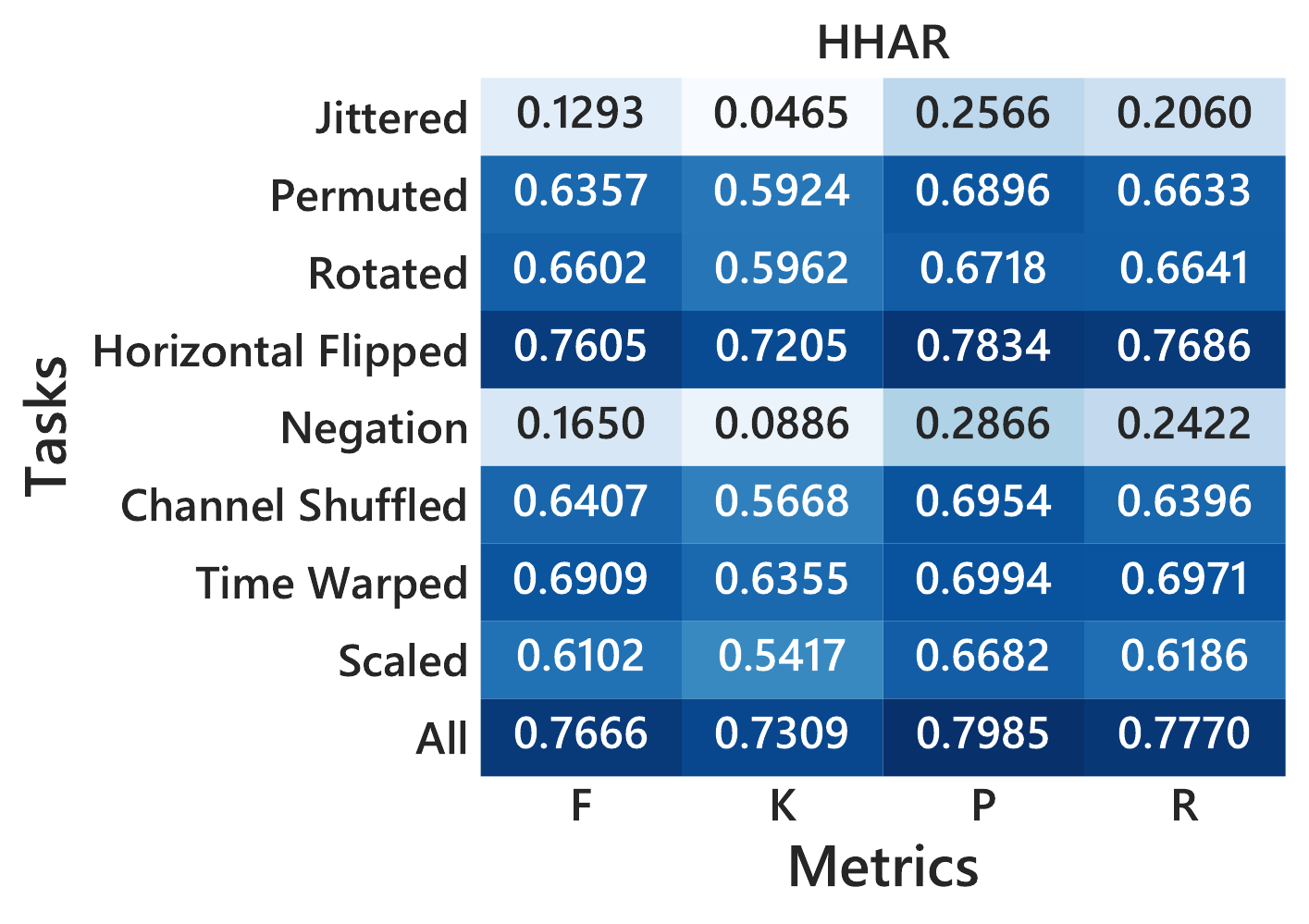}}\hspace{0.01cm}
\subfloat{\includegraphics[width=.32\textwidth]{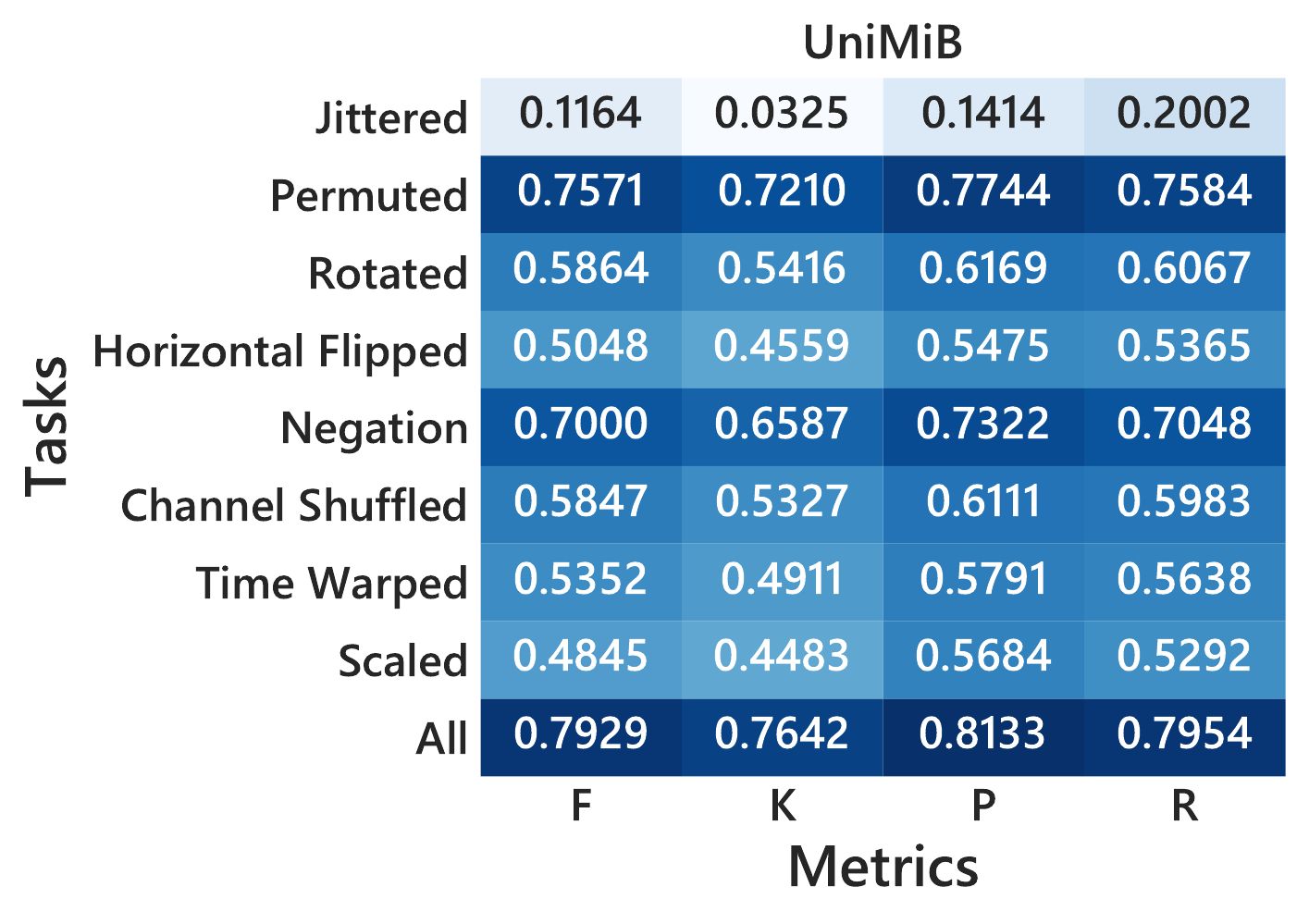}} \hspace{0.01cm}
\subfloat{\includegraphics[width=.32\textwidth]{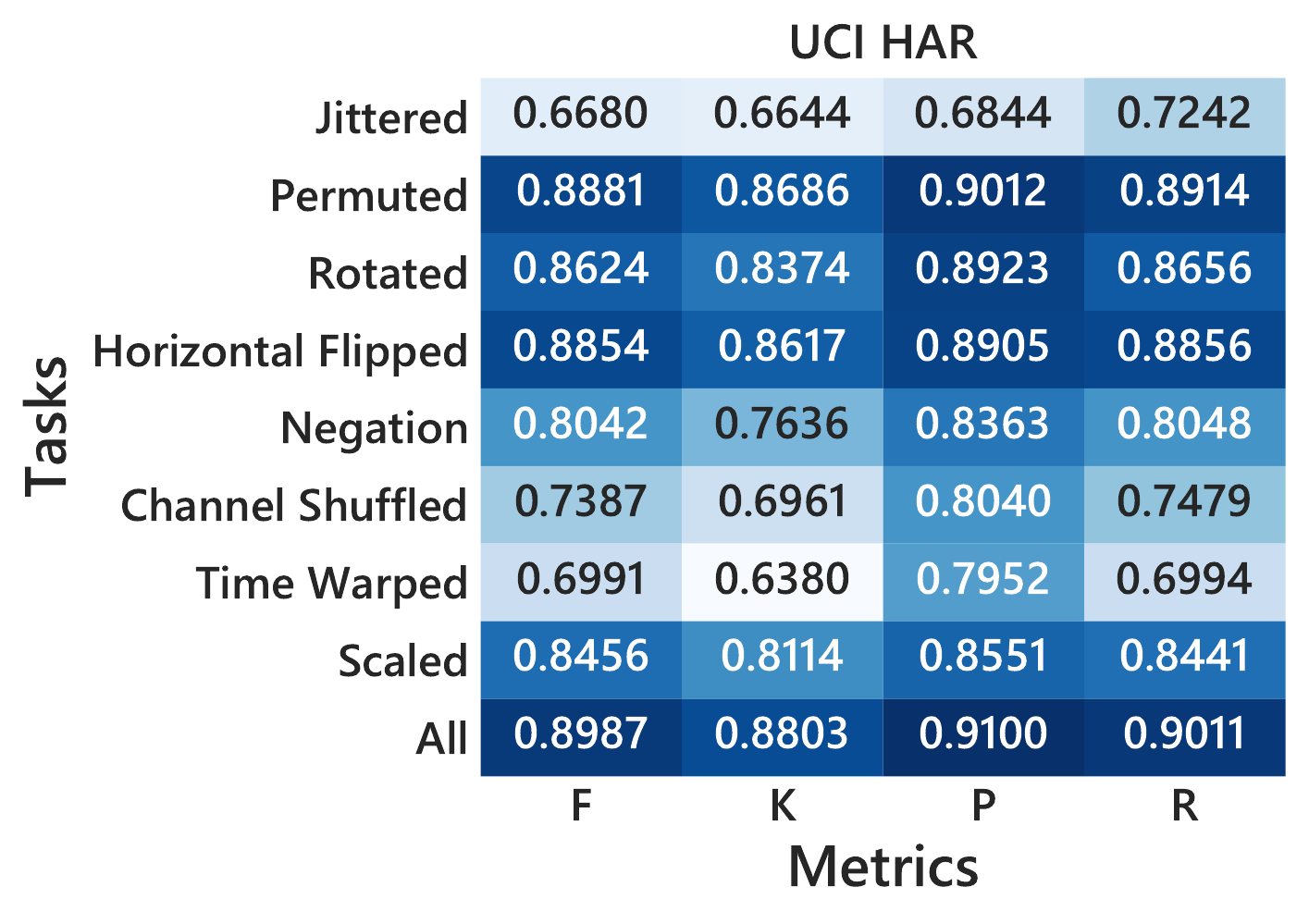}} \hspace{0.01cm}\\
\subfloat{\includegraphics[width=.32\textwidth]{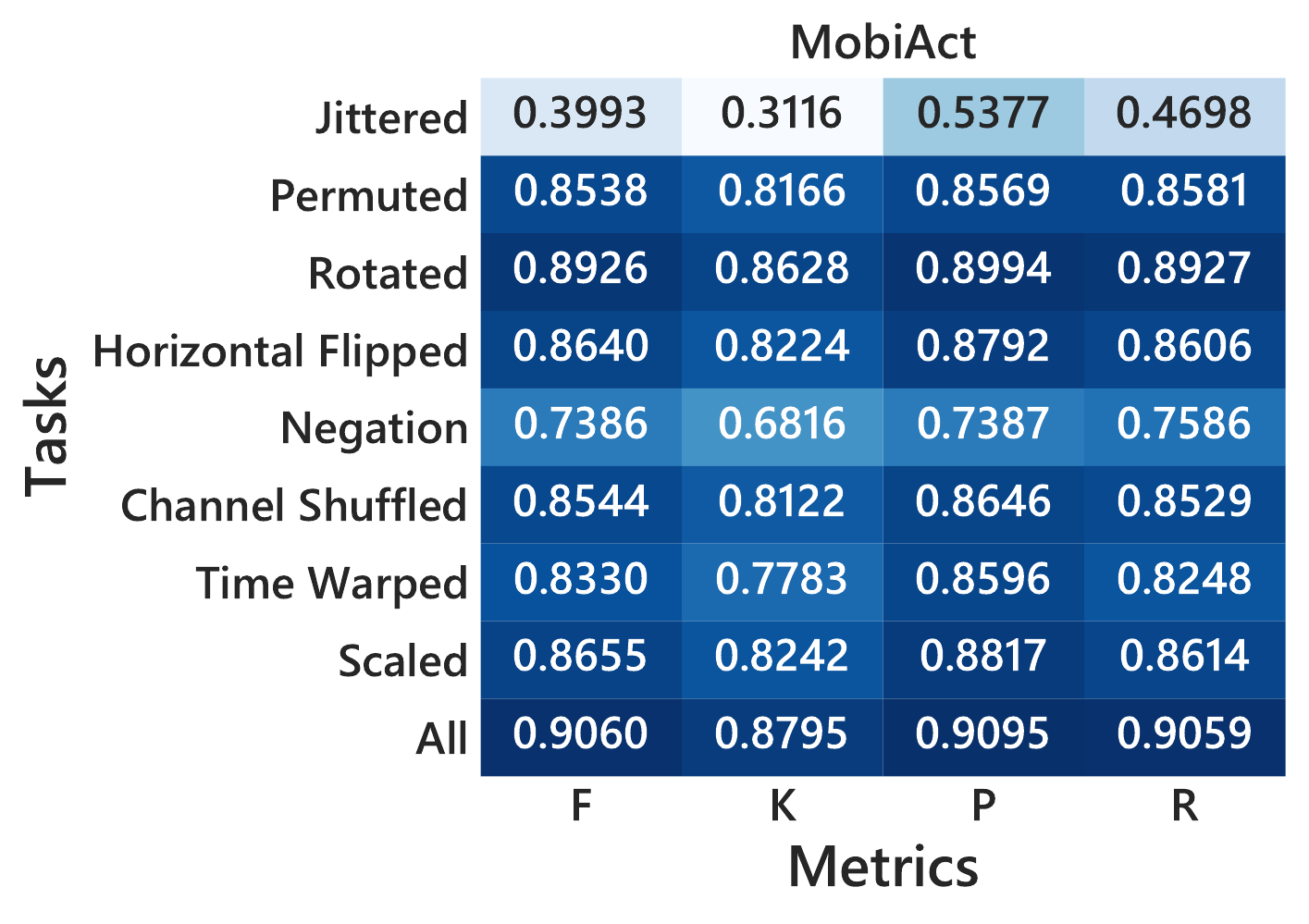}} \hspace{0.01cm}
\subfloat{\includegraphics[width=.32\textwidth]{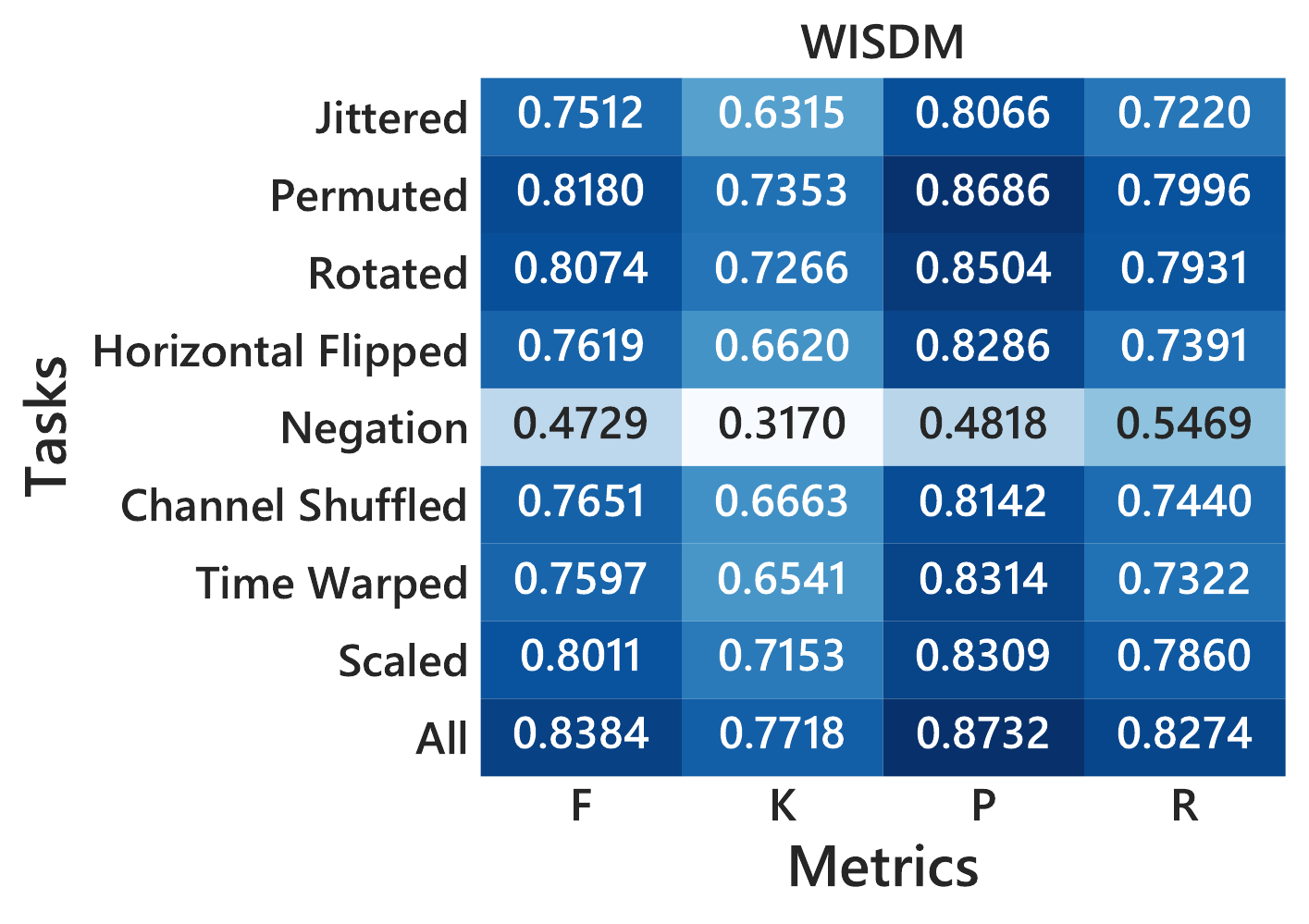}}\hspace{0.01cm}
\subfloat{\includegraphics[width=.32\textwidth]{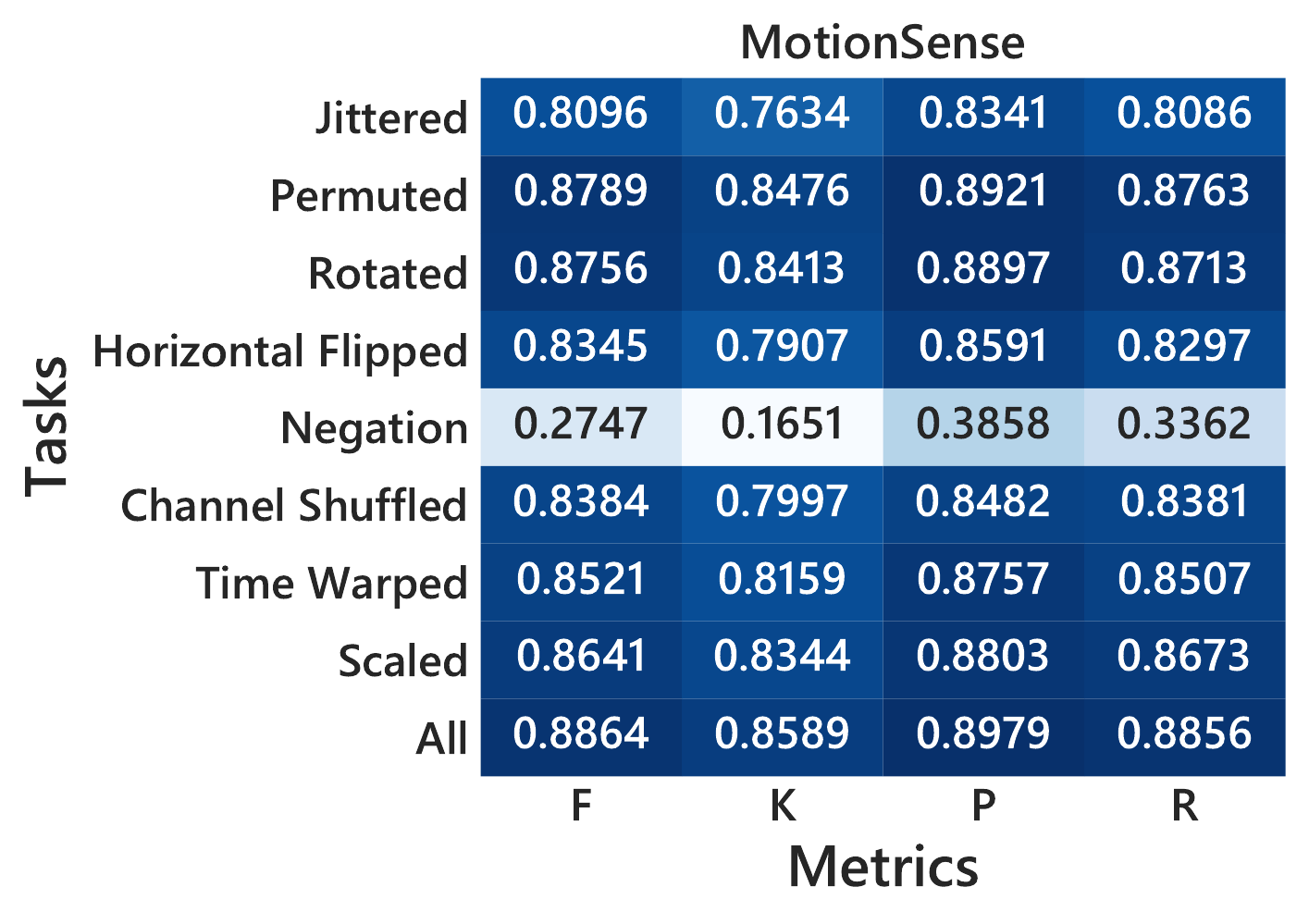}}
\caption{Comparison of individual self-supervised tasks with the multi-task setting. \small{The TPN is pre-trained for solving a particular task and the activity classifier is trained on-top of the learned features. We report the averaged results of evaluation metrics for $10$ independent runs, where F, K, P, and R refer to F-score, Kappa, Precision and Recall, respectively. We observe that multi-task learning improves performance in all the cases with tasks such as \textit{Channel Shuffled}, \textit{Permuted}, and \textit{Rotated} consistently performed better compared to other tasks across datasets.}}
\label{fig:tasks_eval}
\end{figure}

In Figure~\ref{fig:tasks_eval}, we show comparative performance analysis of single self-supervised tasks with each other and importantly with a multi-task setting. This assessment helps us in understanding \textit{whether self-supervised features extracted via jointly learning to solve multiple tasks are any better (for activity classification) than independently solving individual tasks} and \textit{whether multi-task learning helps in learning more useful sensor semantics}. To achieve this, we pre-train a TPN on each of the self-supervised tasks and transfer the weights for learning an activity recognition classifier. We observe in all the cases that learning representations via solving multiple tasks lead to far better performance on the end-task. This further highlights that the features learned through various self-supervised tasks have different strengths and weaknesses. Therefore, merging multiple tasks results in an improvement in learning a diverse set of features. However, we notice that some tasks (such as \textit{Channel Shuffled}, \textit{Permuted}, and \textit{Rotated}) consistently performed better compared to others across datasets; achieving a kappa score above $0.60$ as evaluated on different activity recognition problems. It highlights an important point that there may exist a group of tasks, which are reasonably sufficient to achieve a model of good quality. Furthermore, in Figure~\ref{fig_appendix:trf} of the appendix, we plot the kappa score achieved by a multi-task TPN on transformation recognition tasks as a function of the number of training epochs. This analysis highlights that task complexity varies greatly from one dataset to another and may help with the identification of trivial auxiliary tasks that may lead to non-generalizable features. 

In addition to activity classification, for any learning task involving time-series sensor data (e.g., as encountered in a various Internet of Things applications), we recommend extracting features through first solving individual tasks and later focusing on the multi-task scenario; discarding low performing tasks or assigning low-weights to the loss functions of the respective tasks. Another approach could be to auto-tune the task-loss weight by taking homoscedastic uncertainty of each task into account~\cite{kendall2018multi}.  

\subsubsection{Effectiveness under Semi-Supervised Setting}
\label{sec:esss}
Our proposed self-supervised feature learning method attains very high performance on different activity recognition datasets. This brings up the question, \textit{whether the self-supervised representations can boost performance in the semi-supervised learning setting as well or not.} In particular, can we use this to perform activity detection with very little labeled data? Intrigued by this, we also evaluate the effectiveness of our approach to semi-supervised learning. Specifically, we initially train a TPN on an entire training set for transformation prediction. Subsequently, we learn a classifier on top of the last layer's feature maps with only a subset of the available accelerometer samples and their corresponding activity labels. For training an activity classifier, we use for each category (class) $2$, $5$, $10$, $20$, $50$, and $100$ examples. Note that, $2$-$10$ samples per class represent a real-world scenario of acquiring a (small) labeled dataset from human users with minimal interruption to their daily routines, hence, making self-supervision from unlabeled data of great value. Likewise, we believe, our analysis of learning with very few labeled instances across datasets is the first attempt in quantifying the amount of labeled data required to learn an activity recognizer of decent quality. For self-supervised models, as earlier, we either kept the weights frozen or only fine-tune the last $ConvC$ layer.

\begin{figure}[htbp]
\centering
\subfloat{\includegraphics[width=.3\textwidth]{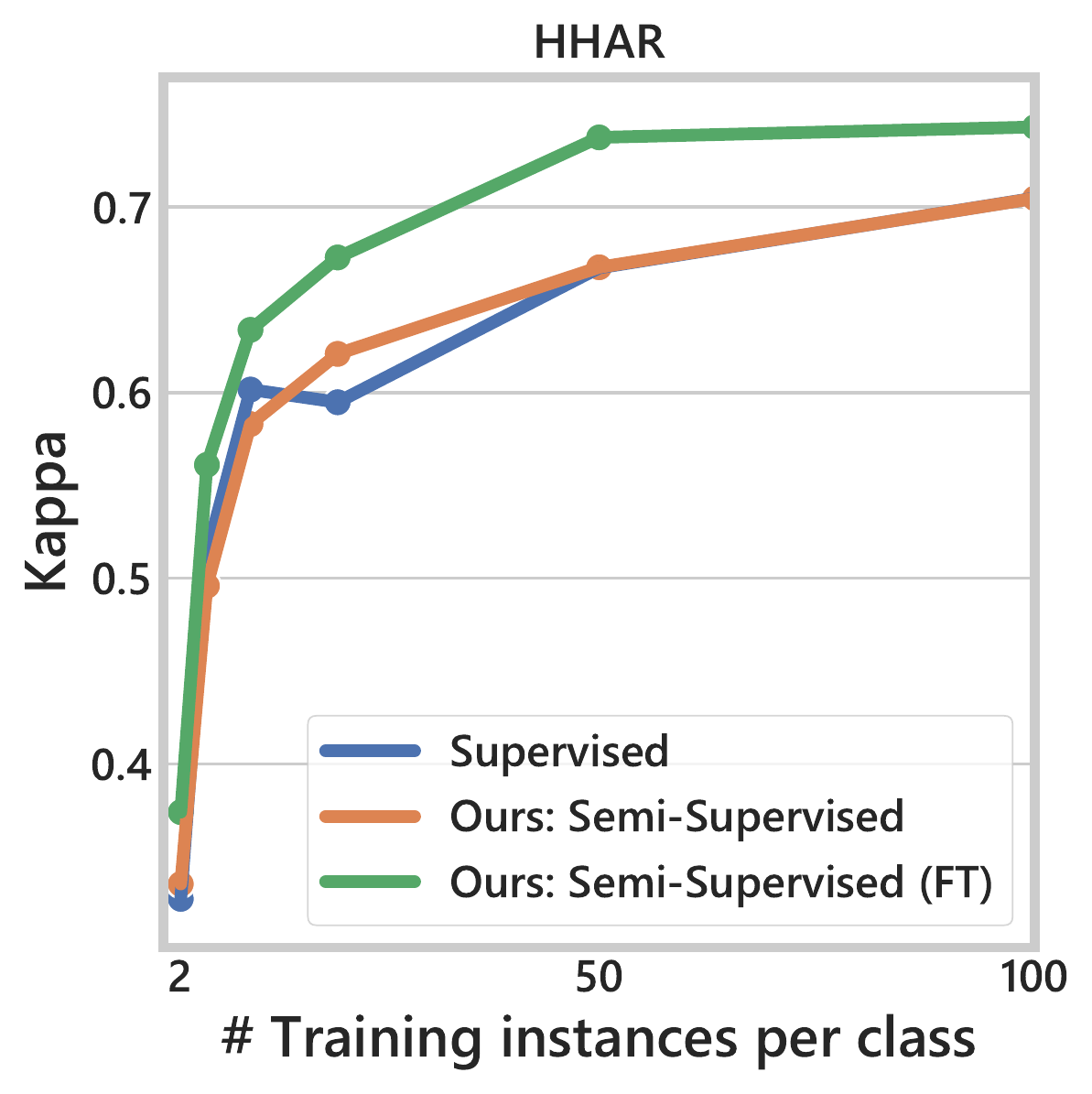}} \hspace{0.01cm}
\subfloat{\includegraphics[width=.3\textwidth]{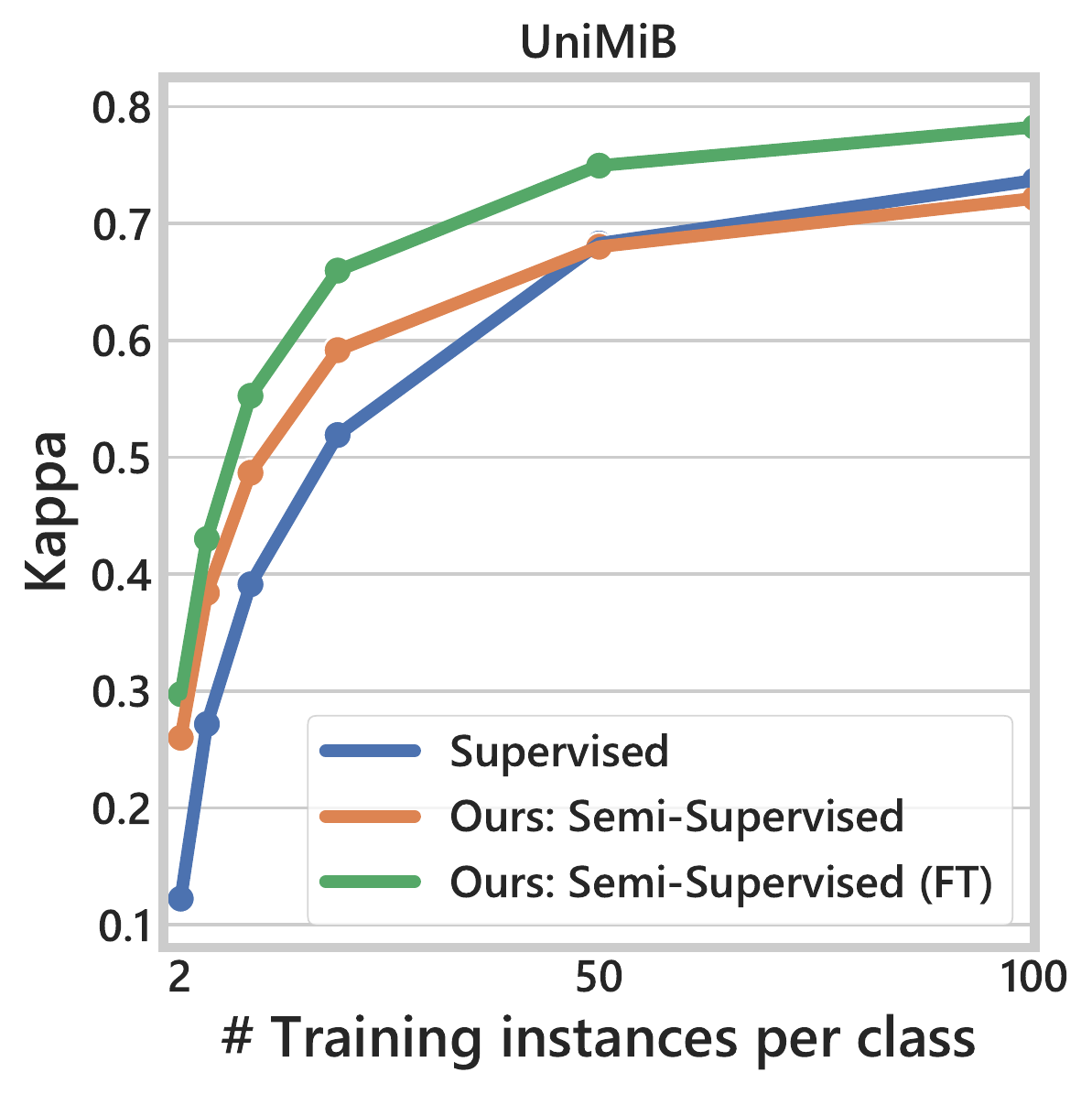}} \hspace{0.01cm}
\subfloat{\includegraphics[width=.3\textwidth]{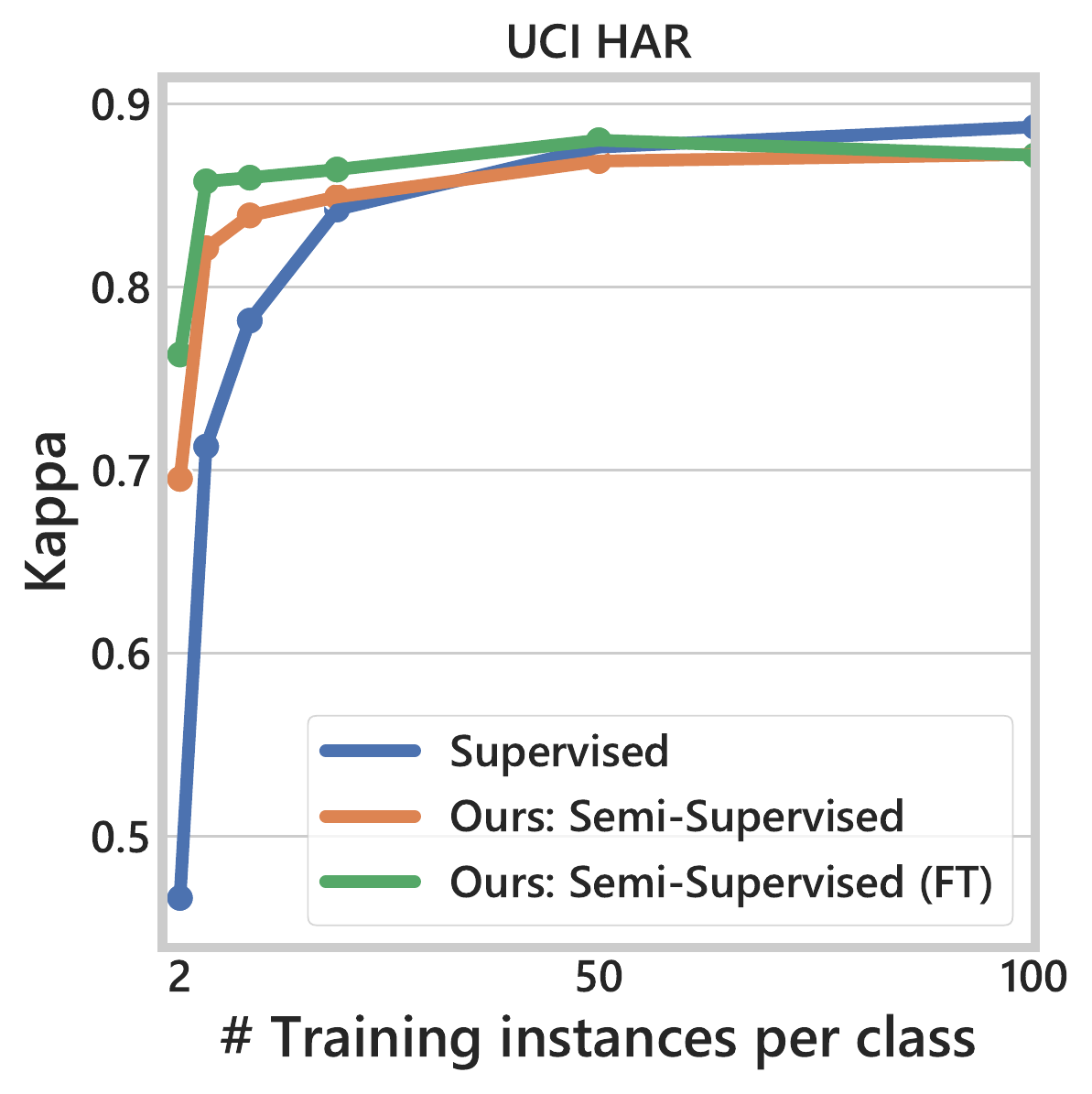}} \hspace{0.01cm}\\
\subfloat{\includegraphics[width=.3\textwidth]{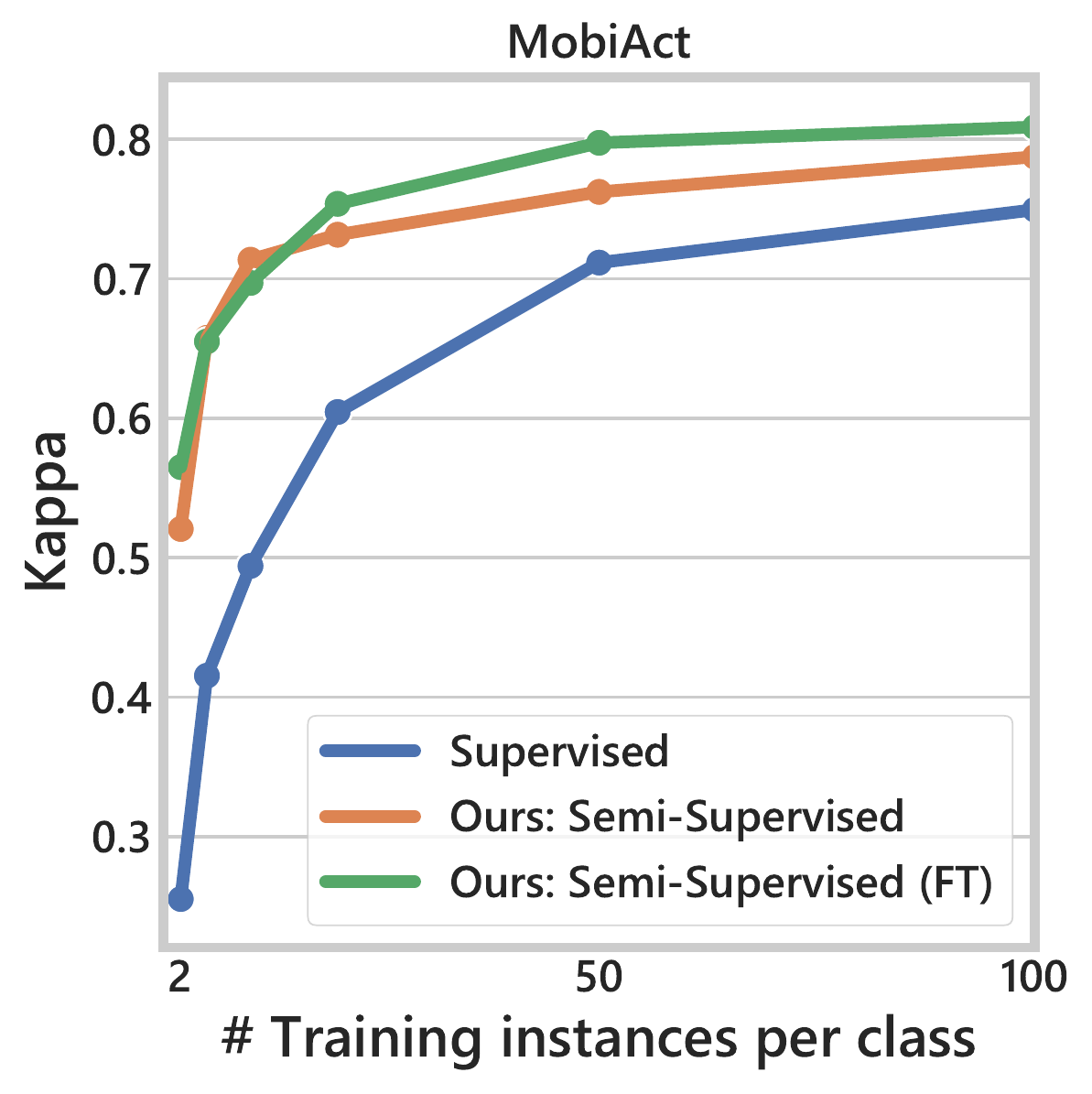}} \hspace{0.01cm}
\subfloat{\includegraphics[width=.3\textwidth]{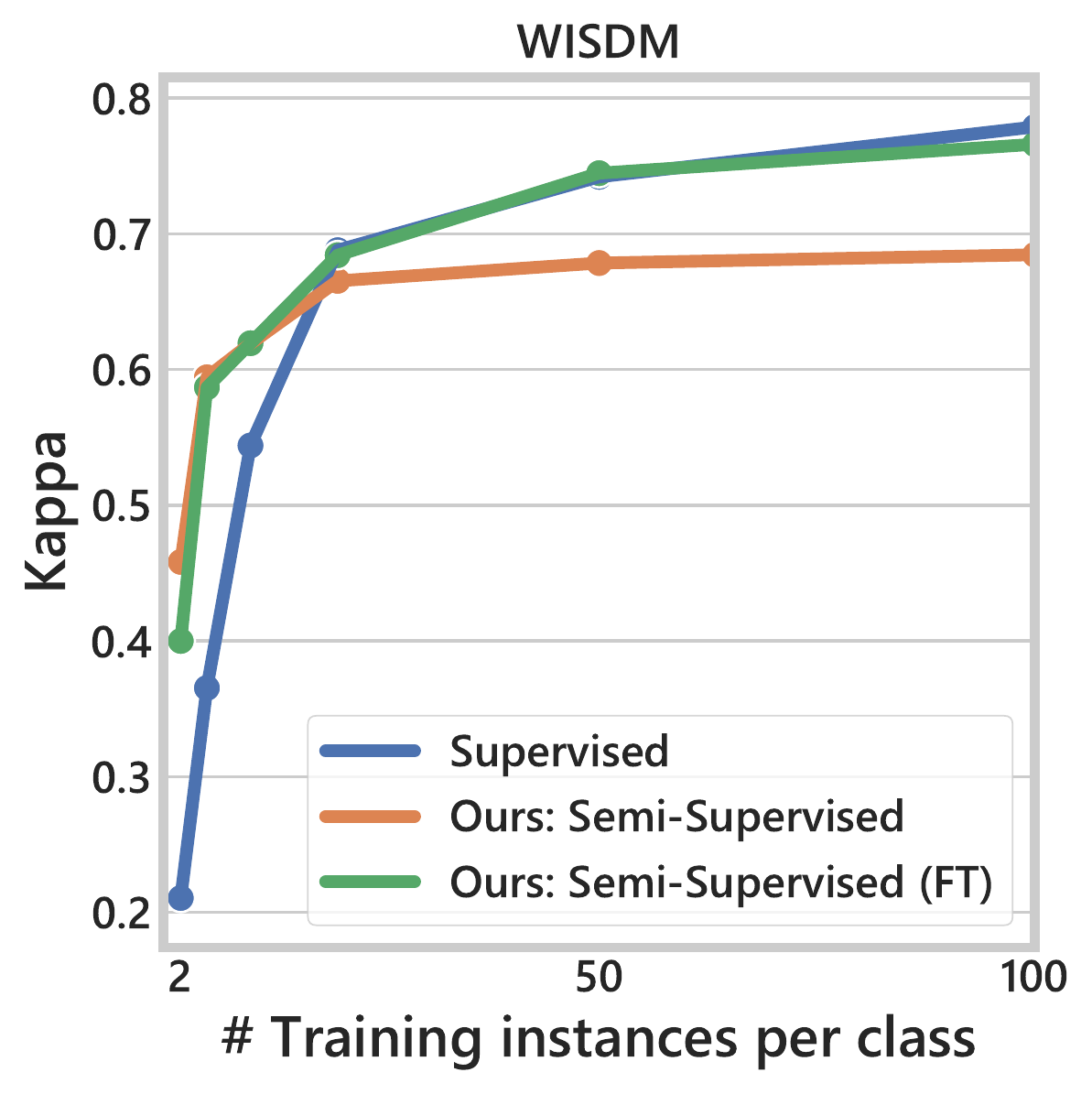}} \hspace{0.01cm}
\subfloat{\includegraphics[width=.3\textwidth]{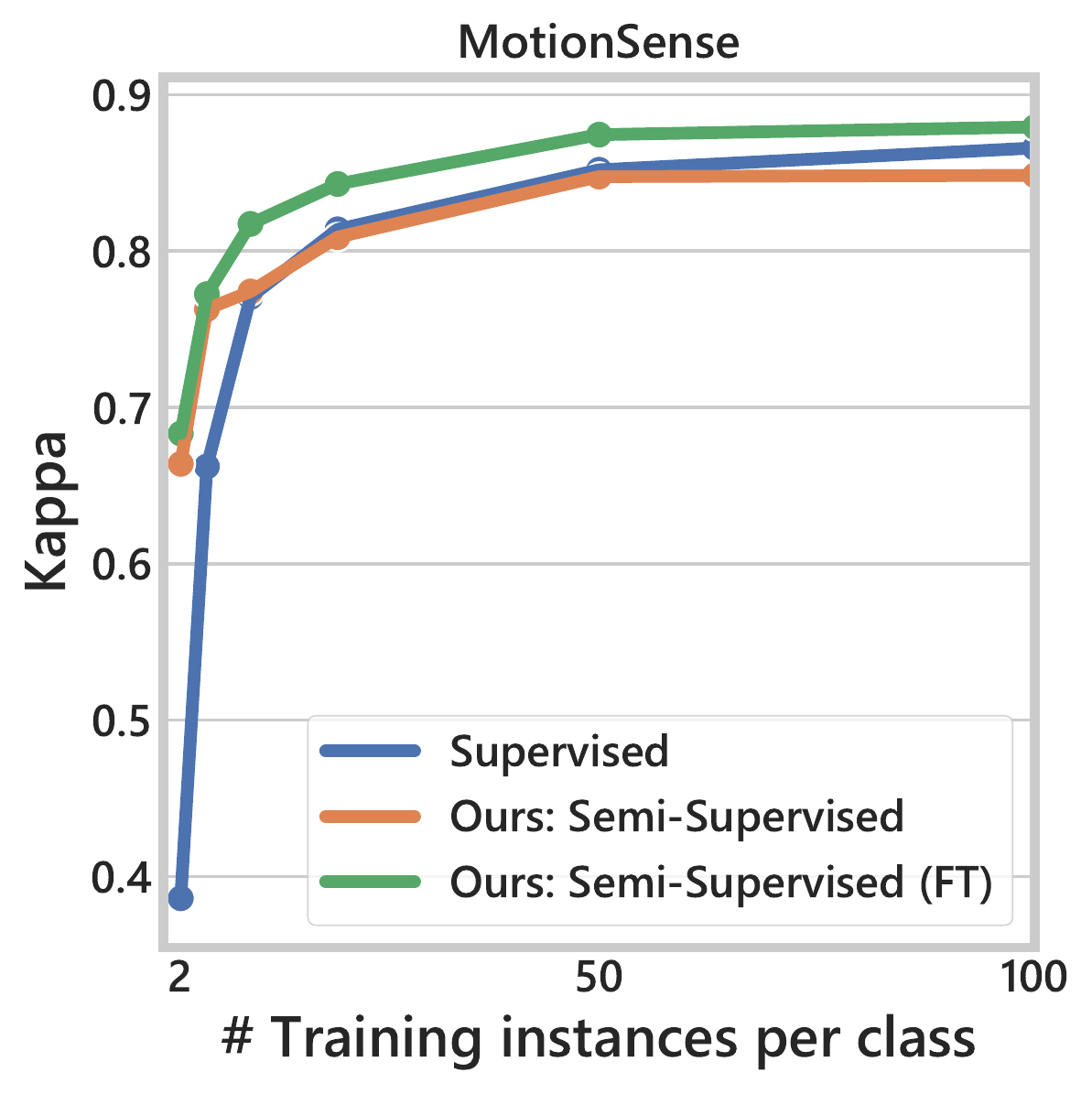}} 
\caption{Generalization of the self-supervised learned features under semi-supervised setting. \small{The TPN is pre-trained on an entire set of unlabeled data in a self-supervised manner and the activity classifier is trained from scratch on $2$, $5$, $10$, $20$, $50$, and $100$ labeled instances per class. The blue curve (baseline) depicts the performance when an entire network is trained in a standard supervised way while the orange curve shows performance when we keep the transferred layers frozen. The green curve illustrates the kappa score when the last layer is fine-tuned along with the training of a classifier on the available set of labeled instances. The reported results are averaged over $10$ independent runs for each of the evaluated approaches. The results with weighted f-score are provided in Figure~\ref{fig_appendix:semi_eval_fscore} of the Appendix.}}
\label{fig:semi_eval}
\end{figure}

In Figure~\ref{fig:semi_eval}, we plot the average kappa of 10-independent runs as a function of the number of available training examples. For each run, we randomly sample desired training instances and train a model from scratch. Note that, we utilize the same instances for evaluating both supervised baseline and our proposed method. The fully-supervised baseline (blue curve) shows network performance when a model is trained only with the labeled data. The proposed self-supervised pre-training technique, in particular, the version with fine-tuning of the last $ConvC$ layer, tremendously improved the performance. The difference in the performance between supervised and self-supervised feature learning is significant on \textit{MotionSense}, \textit{UCI HAR}, \textit{MobiAct}, and \textit{HHAR} datasets in low-data regime (i.e. with $2$-$10$ labeled instances per class). More notably, we observe that pre-training helps more in a semi-supervised setting when the data are collected from a wide variety of devices; simulating a real-life setting. Finally, we highlight that a simple convolutional network is used in our experiments to show the feasibility of self-supervision from unlabeled data. We believe a deeper network trained on a bigger unlabeled dataset will further improve the quality of learned representations for the semi-supervised setting.

\subsubsection{Evaluating Knowledge Transferability}
\label{sec:dkt}
We have shown that representations learned by the self-supervised TPN consistently achieve the best performance as compared to other unsupervised/supervised techniques and also in a semi-supervised setting. As we have utilized the unlabeled data from the same data source for self-supervised pre-training, a next logical question that arises is \textit{can we utilize a different (yet similar) data source for self-supervised representation extraction and gain a performance improvement on a task of interest (also in a low-data regime)?} In Table~\ref{tab:tf_ss_eval}, we assess the performance of our unsupervised learned features across datasets and tasks by fine-tuning them on \textit{HAHR}, \textit{UniMiB}, \textit{UCI HAR}, \textit{WISDM}, and \textit{MotionSense} datasets. For self-supervised feature learning, we utilized the unlabeled \textit{MobiAct} dataset as it is collected from a diverse group of users that performed twelve activities; highest among other considered datasets both in terms of the number of users and activities. This makes \textit{MobiAct} a suitable candidate to perform transfer learning as it encompasses all the activity classes in other datasets. Of course, we do not utilize activity labels in \textit{MobiAct} for self-supervised representation learning. We begin by pre-training a network on \textit{MobiAct} dataset and utilize the learned weights for initialization of an activity recognition model. Moreover, the latter model is trained in a fully-supervised manner on an entire training set of a particular dataset (e.g., \textit{UniMiB}). In comparison with supervised training of the network (from scratch), the weights learned through our technique from a different and completely unlabeled data source improved the performance in all the cases. On \textit{WISDM} and \textit{HHAR} our results are $3$ percentage points better in terms of kappa score. Similarly, on \textit{UniMiB} we obtained $4$ percentage points improvement over supervised model, i.e. kappa score increase from $0.781$ to $0.821$. 

\begin{table}[htbp]
\caption{Task and Dataset Generalization: Quantifying the quality of transferred self-supervised network. \small{We pre-train a TPN on \textit{MobiAct} dataset with the proposed self-supervised approach. The classifier is added on the transferred model and trained in an end-to-end fashion on a particular activity recognition dataset. We chose \textit{MobiAct} for transfer learning evaluation because of the large number of users and activity classes it covers. The reported results are averaged over $10$ independent runs, where $P$, $R$, $F$, and $K$ refer to Precision, Recall, F-score, and Kappa, respectively.}}
\scriptsize{
\begin{tabular}{lcccccccc} 
& \multicolumn{4}{c}{\textbf{Supervised (From Scratch)}} & \multicolumn{4}{c}{\textbf{Transfer (Self-Supervised)}}                                                         \\ \cline{2-9} 
\multirow{-2}{*}{\textbf{Dataset}} & \textbf{P} & \textbf{R} & \textbf{F} & \textbf{K} & \multicolumn{1}{c}{\textbf{P}} & \multicolumn{1}{c}{\textbf{R}} & \multicolumn{1}{c}{\textbf{F}} & \multicolumn{1}{c}{\textbf{K}} \\ \hline
\textbf{HHAR}                      & 0.7624$\pm$0.0312&0.7353$\pm$0.0308&0.7276$\pm$0.0297&0.6816$\pm$0.0371       & 0.7816$\pm$0.0405&0.7617$\pm$0.0469&0.7549$\pm$0.0452&\textbf{0.713}$\pm$\textbf{0.056} \\
\textbf{UniMiB}                    & 0.8276$\pm$0.0148&0.8096$\pm$0.0266&0.8097$\pm$0.0248&0.7815$\pm$0.0299       & 00.8557$\pm$0.0123&0.8444$\pm$0.0191&0.8445$\pm$0.0185&\textbf{0.8214}$\pm$\textbf{0.0217}   \\ 
\textbf{UCI HAR}                   & 0.9059$\pm$0.0133&0.8998$\pm$0.0139&0.8981$\pm$0.0148&0.8789$\pm$0.0168      & 0.9097$\pm$0.0129&0.9073$\pm$0.0145&0.9065$\pm$0.0152&\textbf{0.8879}$\pm$\textbf{0.0175}   \\
\textbf{WISDM}                     & 0.9024$\pm$0.0076&0.8657$\pm$0.0206&0.8764$\pm$0.0168&0.8211$\pm$0.0258      & 0.9058$\pm$0.0102&0.8907$\pm$0.0113&0.8946$\pm$0.0108&\textbf{0.8517}$\pm$\textbf{0.0153}   \\
\textbf{MotionSense}               & 0.9164$\pm$0.0053&0.8993$\pm$0.0091&0.9027$\pm$0.0085&0.8763$\pm$0.011        & 0.9223$\pm$0.0081&0.9059$\pm$0.0132&0.9096$\pm$0.0126&\textbf{0.8843}$\pm$\textbf{0.016}   \\ \hline
\end{tabular}
}
\label{tab:tf_ss_eval}
\end{table}

\begin{figure}[htbp]
\centering
\subfloat{\includegraphics[width=.3\textwidth]{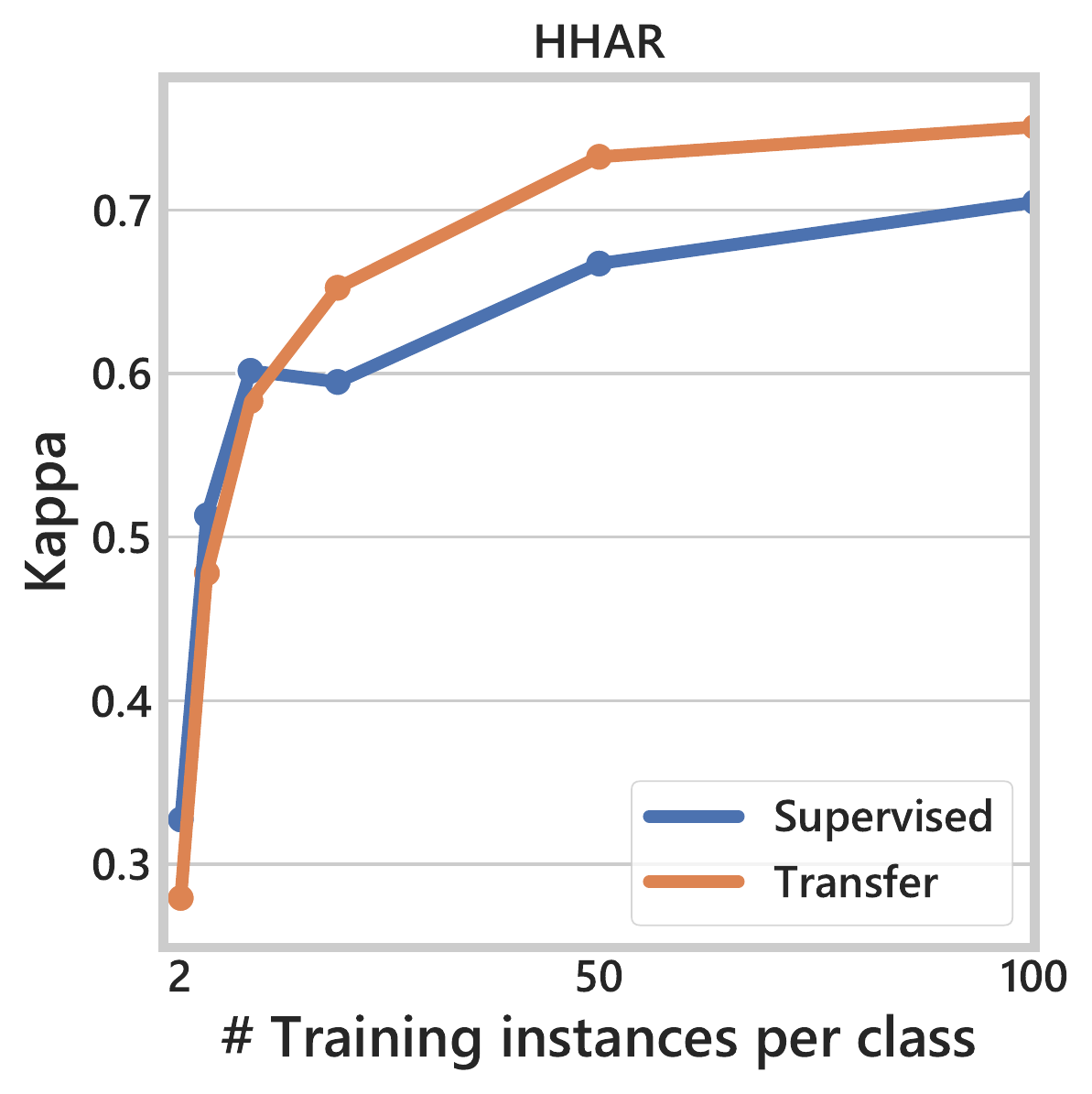}} \hspace{0.01cm}
\subfloat{\includegraphics[width=.3\textwidth]{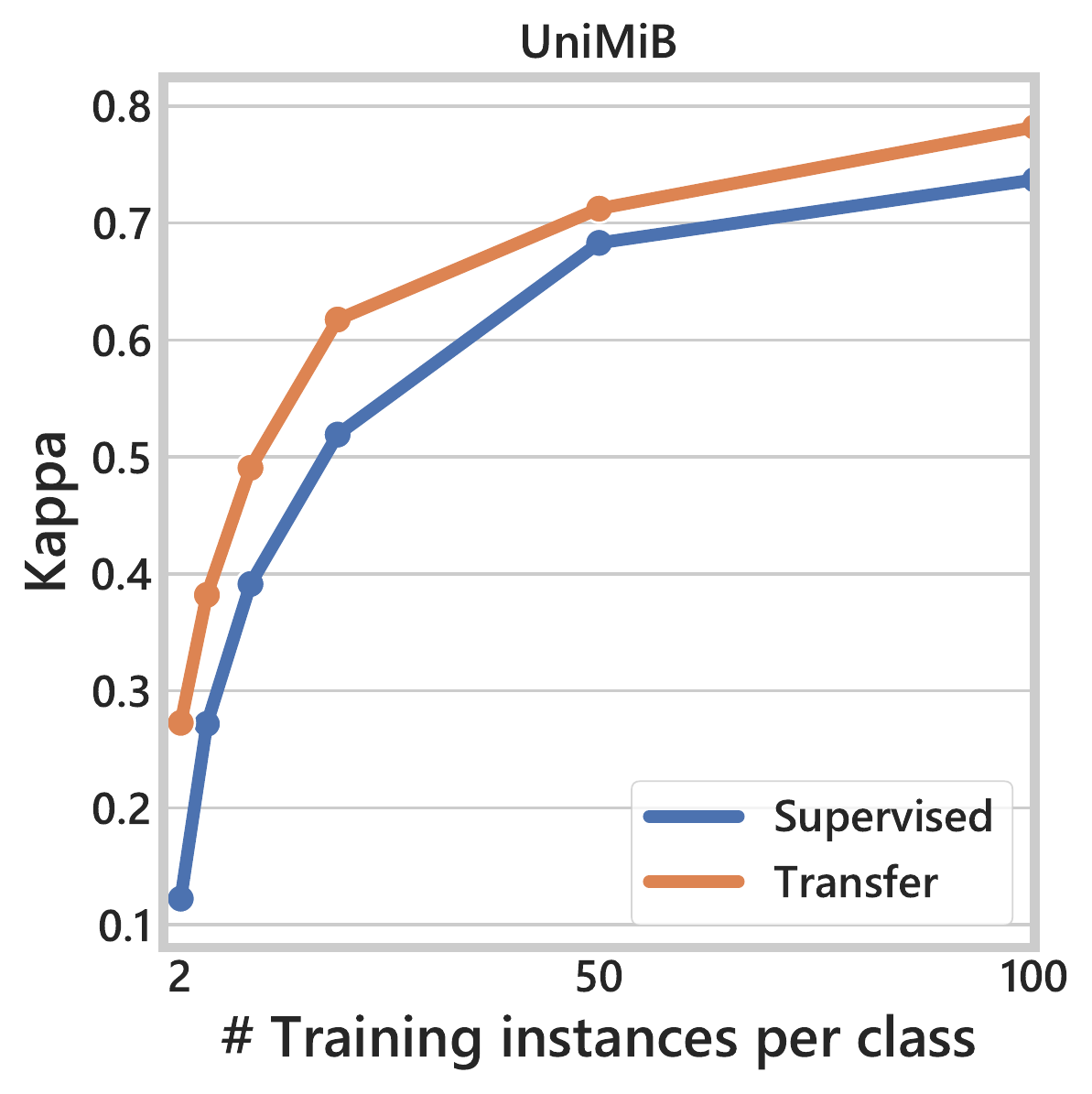}} \hspace{0.01cm}
\subfloat{\includegraphics[width=.3\textwidth]{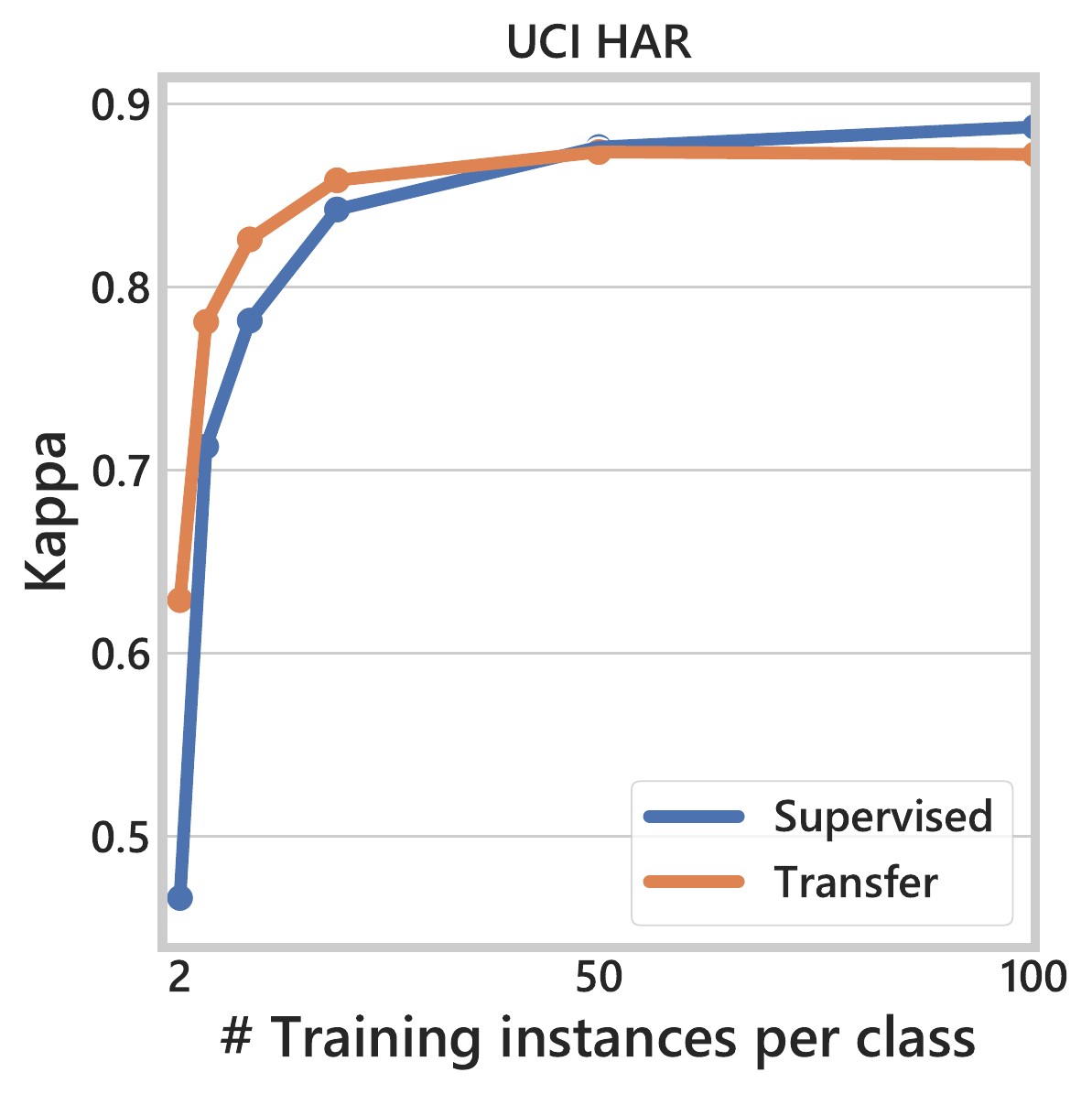}} \hspace{0.01cm}
\subfloat{\includegraphics[width=.3\textwidth]{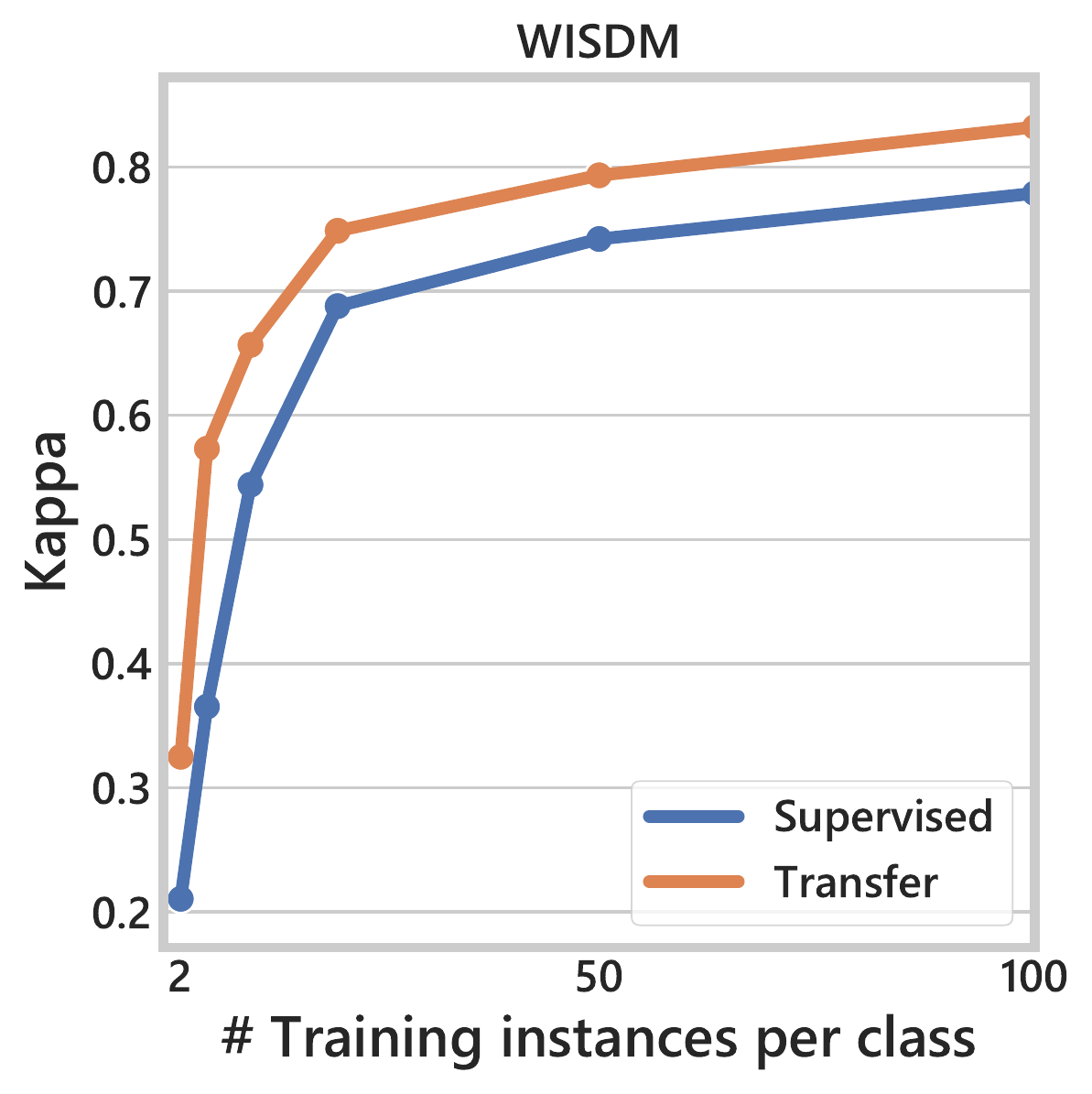}} \hspace{0.01cm}
\subfloat{\includegraphics[width=.3\textwidth]{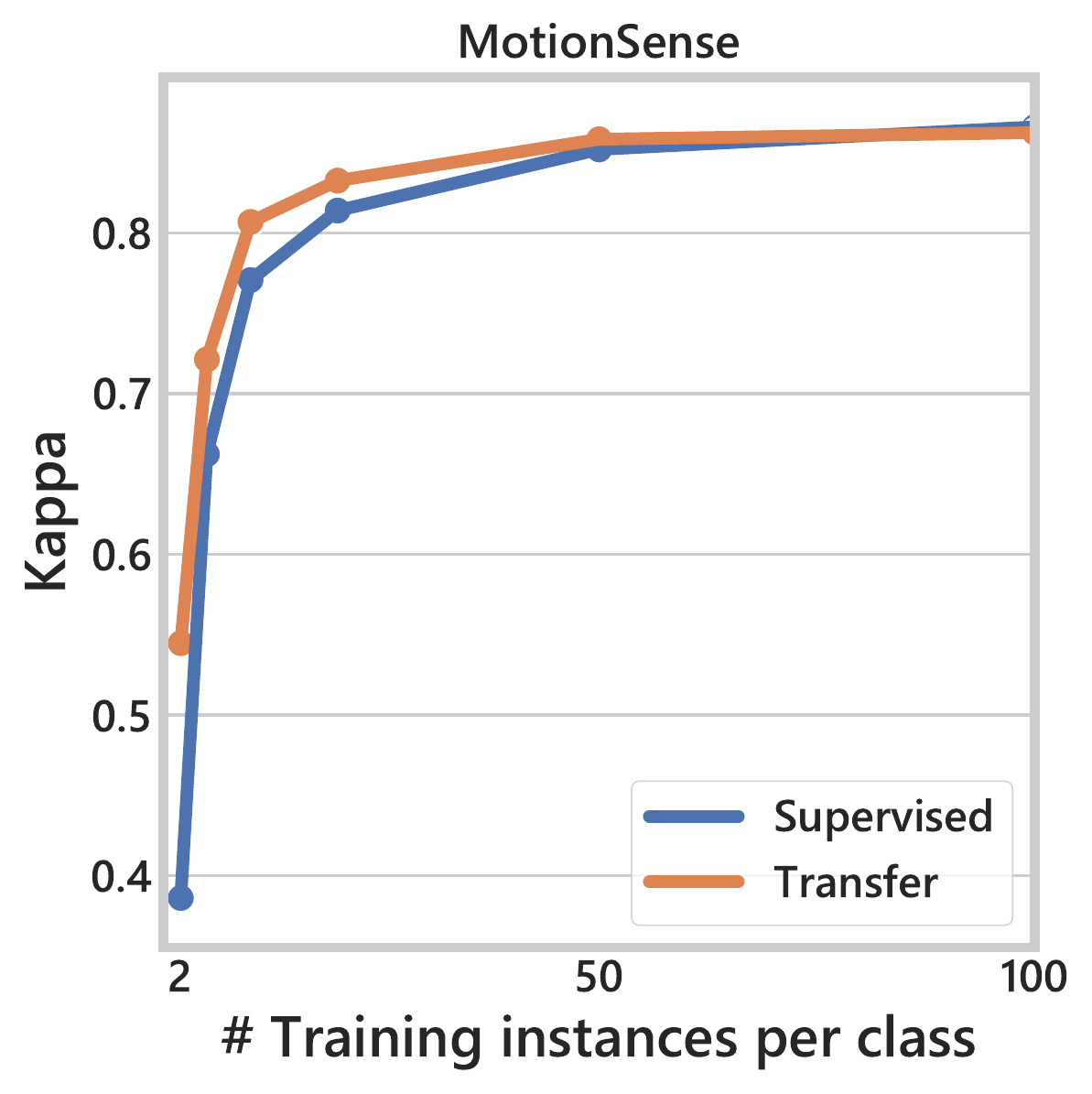}}
\caption{Assessment of the transferred self-supervised learned features from a different but related dataset (MobiAct) under semi-supervised setting. \small{We evaluate the performance of the self-supervised approach when different unlabeled data are accessible for representation learning but very few labeled instances are available for training a network on the task of interest. The TPN is pre-trained initially on \textit{MobiAct} data and the activity classifier is added on-top; later an entire network is trained in an end-to-end fashion on few labeled instances. The reported results are averaged over $10$ independent runs for each of the evaluated approaches when we randomly sample $2$, $5$, $10$, $20$, $50$, and $100$ for learning an activity classifier. The results with weighted f-score are provided in Figure~\ref{fig_appendix:tf_eval_fscore} of the Appendix.}}
\label{fig:tf_eval}
\end{figure}

Further, we determine the generalization ability in a low-data regime setting, i.e., when very few labeled data are attainable from an end-task of interest. We transfer self-supervised learned representations on the \textit{MobiAct} dataset as initialization for an activity recognizer. The network is trained in a supervised manner on the available labeled instances of a particular dataset. Figure~\ref{fig:tf_eval} shows average kappa score of $10$-independent runs of a fully-supervised (learned from scratch) and transferred models for $2$, $5$, $10$, $20$, $50$, and $100$ labeled instances. For each training run, the desired instances are randomly sampled, and for both techniques, the same instances are used for learning the activity classifier. In the majority of the cases, transfer learning improves the recognition performance especially when the number of labeled instances per class are very few, i.e. between $2$ to $10$. In particular, on \textit{HHAR} the performance of a model trained with weights transfer is slightly lower in low-data setting but improves significantly as the number of labeled data points increases. We think it may be because of the complex characteristics of the \textit{HHAR} dataset as it is particularly collected to show heterogeneity of devices (and sensors) having varying sampling rates and its impact on the activity recognition performance.

\subsection{Determining Representational Similarity}
The previous experiments establish the effectiveness of self-supervised sensor representations for activity classification that are significantly better than unsupervised and on-par with fully-supervised approaches. The critical question that arises is \textit{whether the self-supervised representations are similar to those learned via direct supervision, i.e., with activity labels.} The interpretability of the neural networks and deciphering of the learned representations have recently gained significant attention, especially, for images (see \cite{olah2018the} for an excellent review). Here, to better understand the similarity of the extracted representation from TPN and the supervised network, we utilize singular vector canonical correlation analysis (SVCCA)~\cite{raghu2017svcca}, saliency maps~\cite{simonyan2013deep} and t-distributed stochastic neighbor embedding (t-SNE)~\cite{maaten2008visualizing}. 

\subsubsection*{Insights on Representational Similarity with Canonical Correlation}
\begin{figure}[!htbp]
\centering
\includegraphics[width=5in]{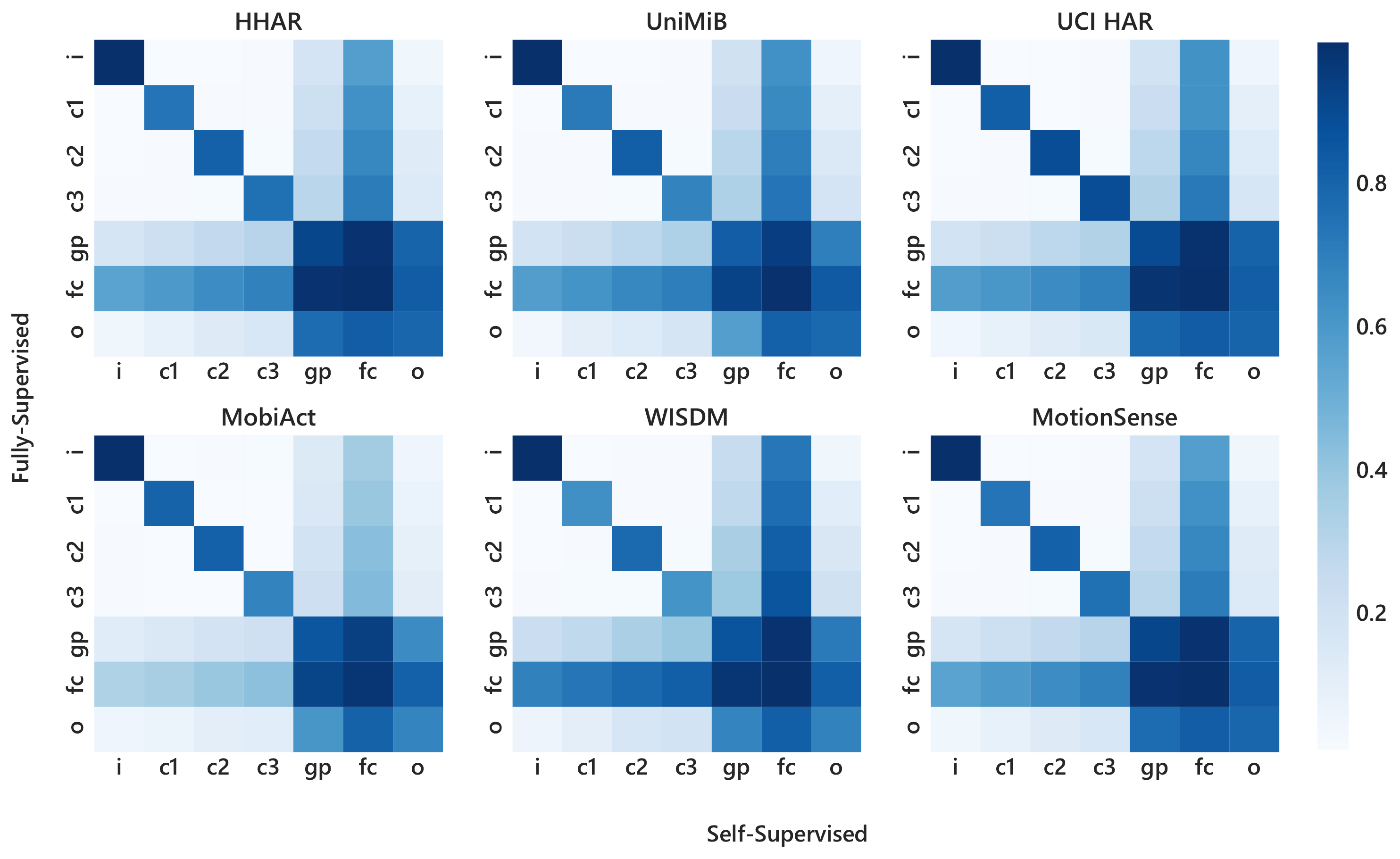}
\caption{CCA similarity between fully-supervised and self-supervised networks. \small{We employ the SVCAA technique~\cite{raghu2017svcca} to determine the representational similarity between model layers trained with our proposed approach and standard supervised setting. Each pane is a matrix of size layers $\times$ layers with each entry showing mean similarity (i.e., an average of top-$20$ correlation coefficients) between the two layers. Note that there is a strong relation between convolutional layers even though the self-supervised network is pre-trained with unlabeled data; showing that features learned by our approach are very similar to those learned directly via supervised learning, with activity classes. Likewise, a grid-like structure appears between the last layers of the networks depicting high similarity as those layers are always (randomly initialized and) trained with activity labels.}}
\label{fig:svcca}
\end{figure}

The SVCCA allows for a comparison of the learned distributed representations across different networks and layers. It does so through identifying optimal linear relationships between two sets of multidimensional variates (i.e., neuron activation vectors) arising from an underlying process (i.e., a neural network being trained on a specific task)~\cite{raghu2017svcca}. Figure~\ref{fig:svcca} provides a mean similarity of top $20$ SVCCA correlation coefficients for all pairs of layers for a self-supervised (trained to predict transformations) and a fully-supervised network. We averaged $20$ coefficients as SVCCA implicitly assumes that all CCA vectors are equally crucial for the representations at a specific layer. However, there is plenty of evidence that high-performing deep networks do not utilize the entire dimensionality of a layer~\cite{morcos2018importance, lecun1990optimal, li2018measuring}. Due to this, averaging over all the coefficients underestimates the degree of representational similarity. To apply SVCCA, we train both the networks as explained earlier and produce activations of each layer. For a layer, where the number of neurons is larger than the layer in comparison, we randomly sample neuron activation vectors to have comparable dimensionality. In Figure~\ref{fig:svcca} each grid entry represents a mean SVCCA similarity between two layers of different networks. We observe a high correlation among temporal convolution layers trained with two different methods across all the evaluated datasets. In particular, a strong grid-like structure emerges between the last layers of the networks, which is because those layers are learned from scratch with activity labeled data and result in identical representations. 

\subsubsection*{Visualizing Salient Regions}
To further understand the predictions produced by both models, we visualize saliency maps~\cite{simonyan2013deep} for the highest-scoring class on randomly selected instances from the \textit{MotionSense} dataset. Saliency maps highlight which time steps largely affect the output through computing gradient of the loss function with respect to each input time step. More formally, let $x = [x_1, \ldots, x_N]$ be an accelerometer sample of length $N$ and $C_\theta(x)$ be the class probability produced by a network $C_\theta$(.). The saliency score of each input element $x_k$ indicating its influence on the prediction is calculated as: 
\begin{equation*}
S_k = \ \mid \frac{\partial \mathcal{L}}{\partial x_k} \mid
\end{equation*}
\noindent where $\mathcal{L}$ is the negative log-likelihood loss of an activity classification network for an input example $x$.

\begin{figure}[htbp]
\centering
\subfloat{\includegraphics[width=.45\textwidth]{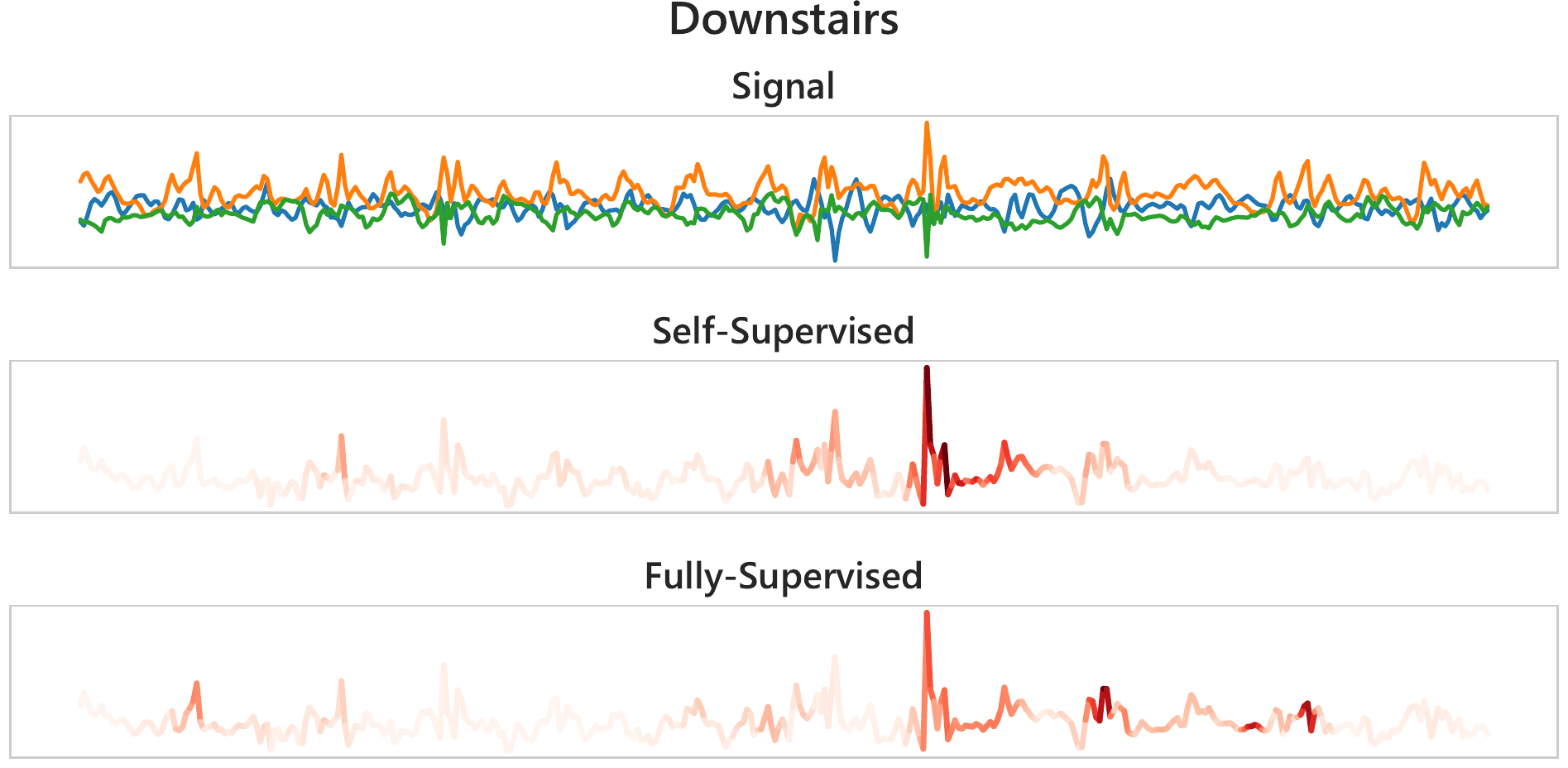}} \hspace{0.01cm}
\subfloat{\includegraphics[width=.45\textwidth]{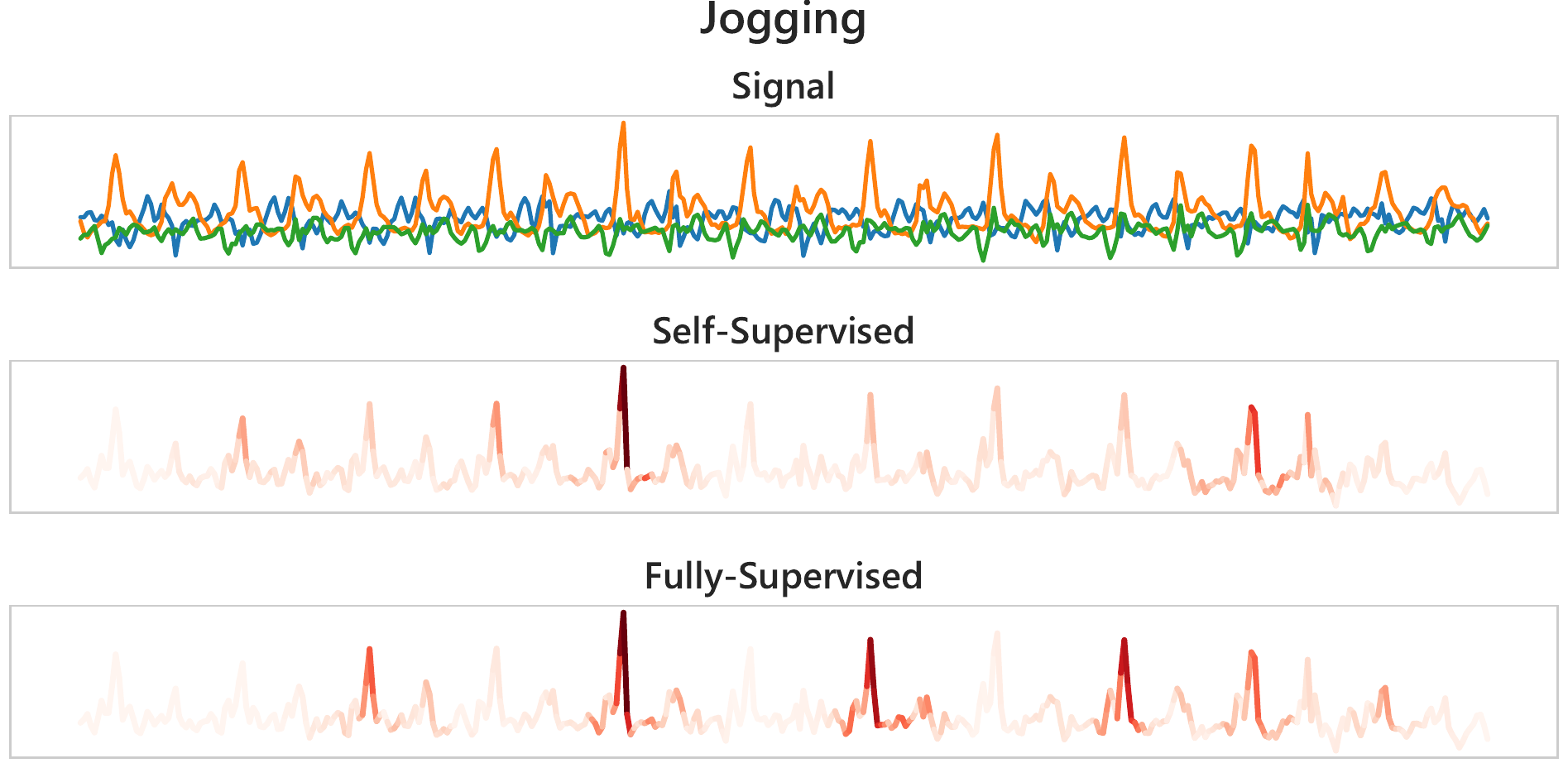}} \\
\subfloat{\includegraphics[width=.45\textwidth]{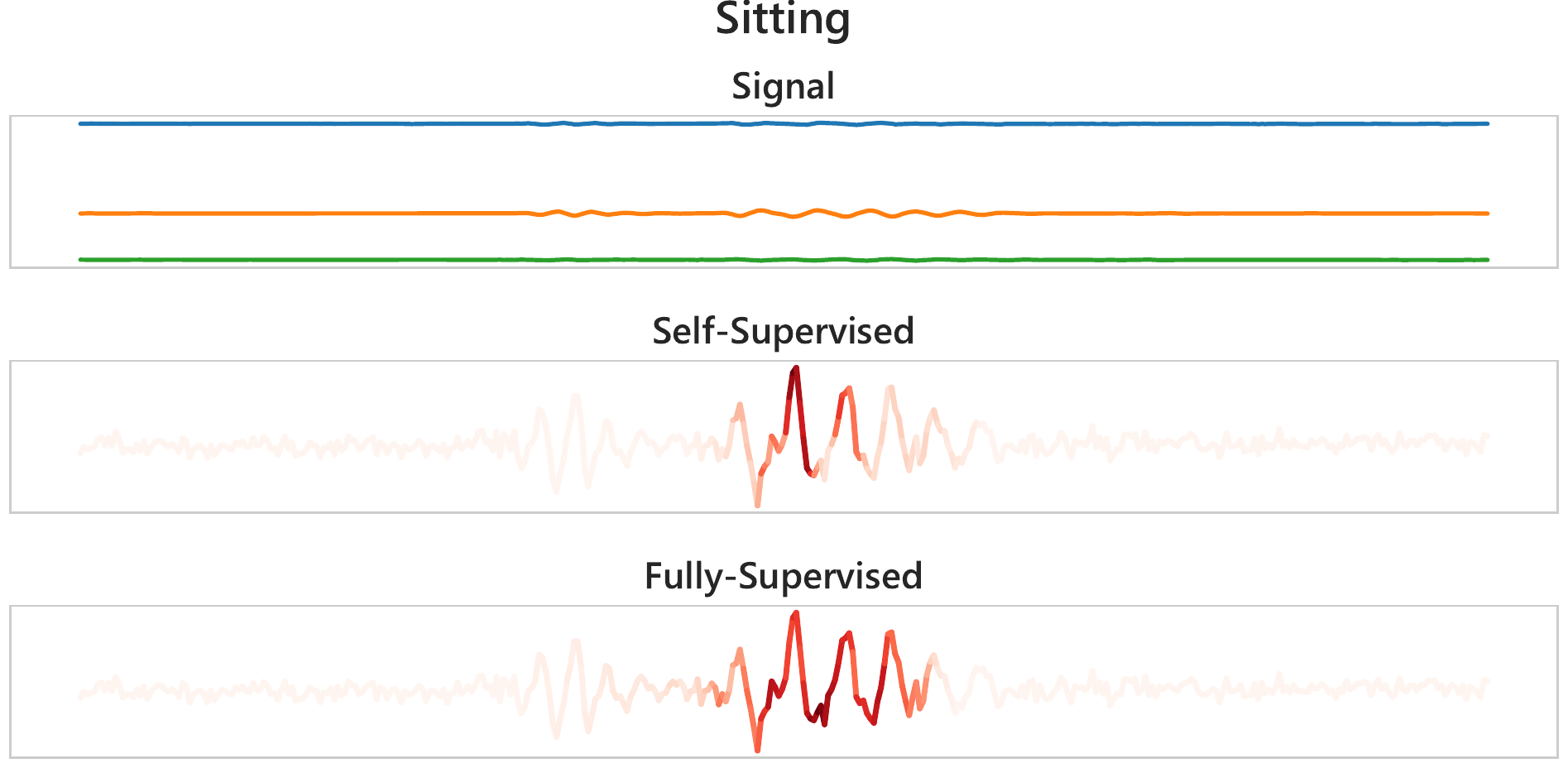}} \hspace{0.01cm}
\subfloat{\includegraphics[width=.45\textwidth]{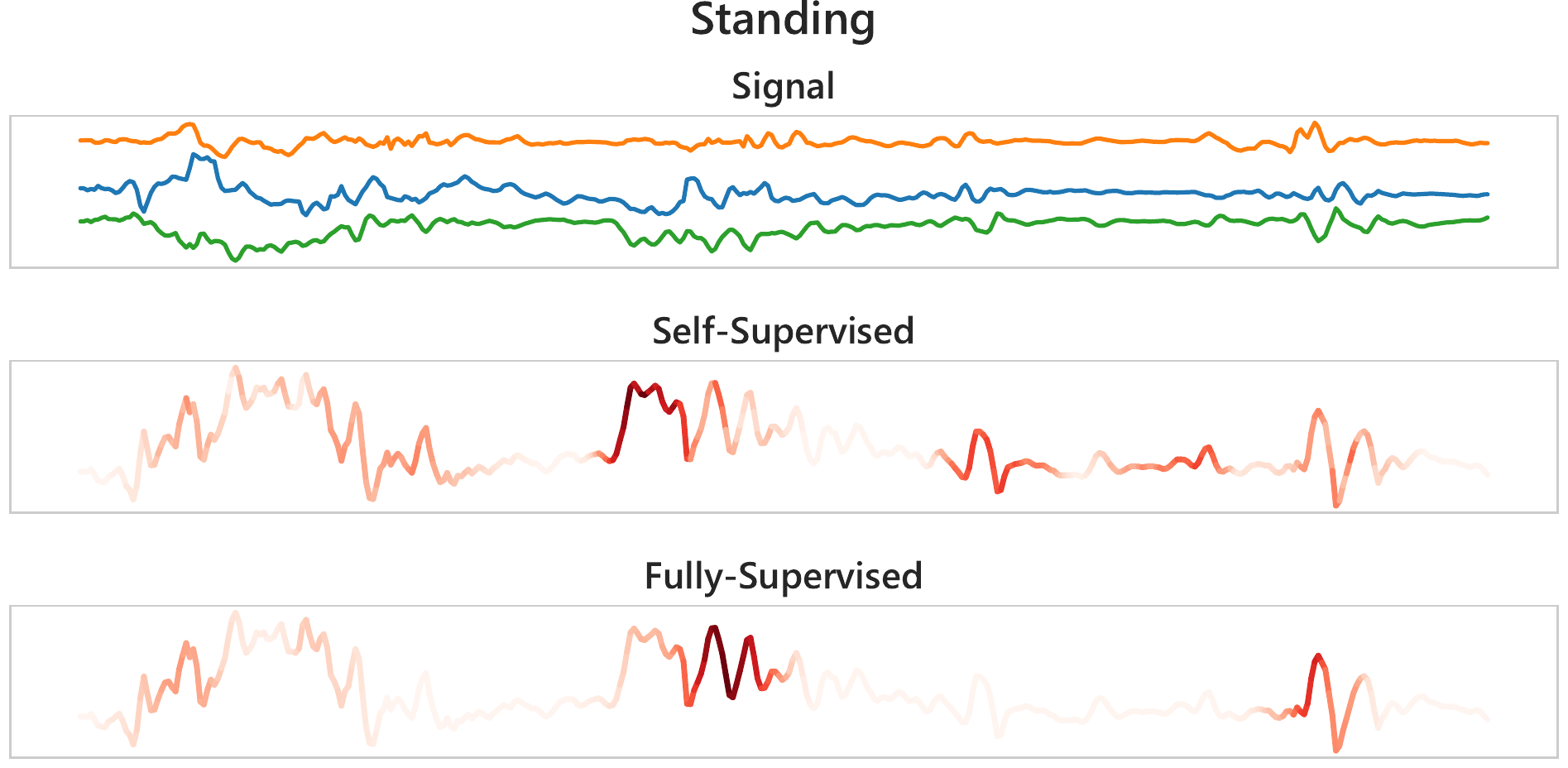}} \\
\subfloat{\includegraphics[width=.45\textwidth]{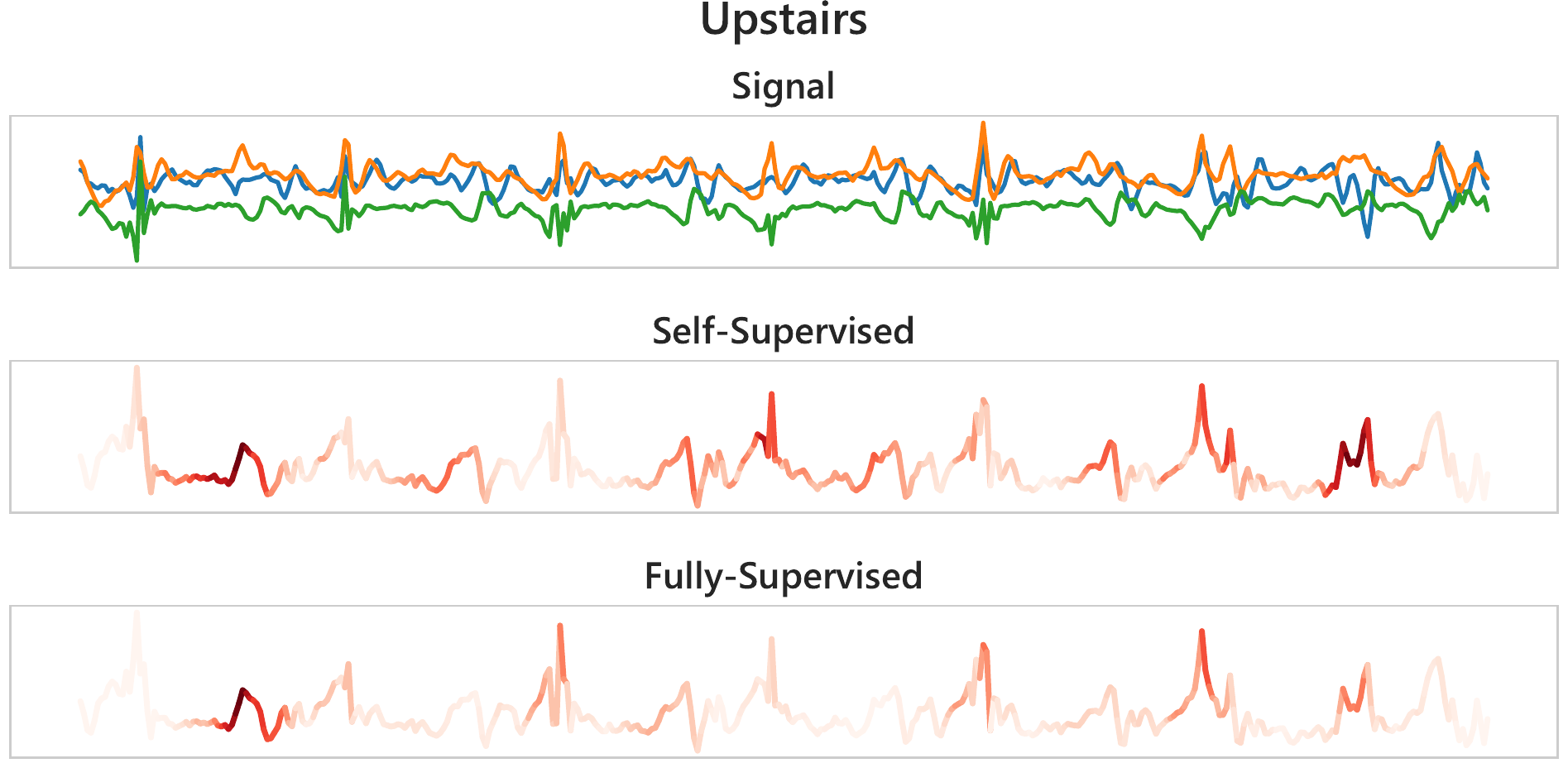}} \hspace{0.01cm}
\subfloat{\includegraphics[width=.45\textwidth]{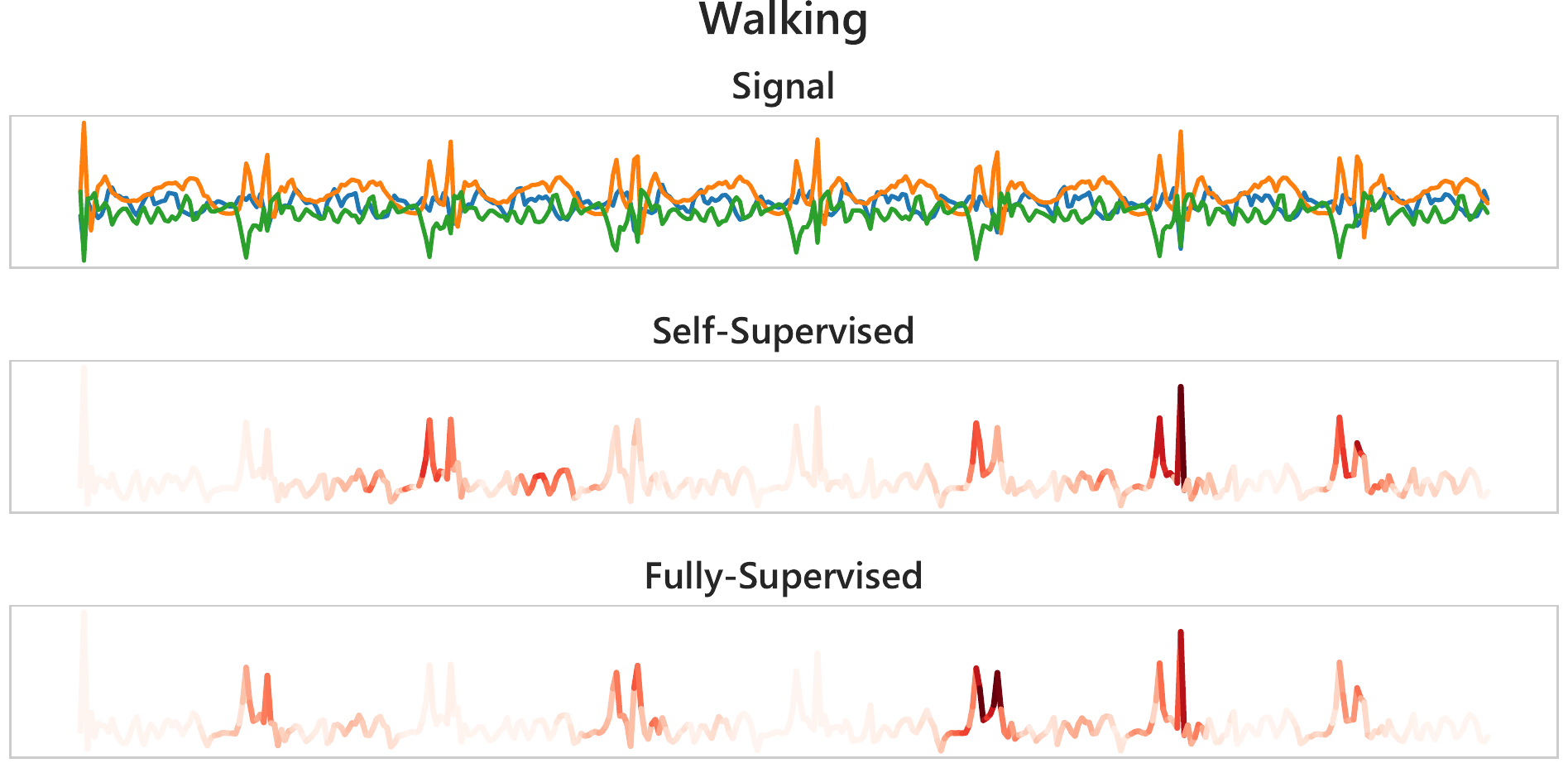}} 
\caption{Saliency maps~\cite{simonyan2013deep} of randomly selected instances from \textit{MotionSense} dataset. \small{The input signal is illustrated in the top pane with the \textit{magnitude} computed from the sample shown in the bottom panes for better interpretability. The strong colored intensities exhibit the regions that substantially affect the model predictions. The saliency mapping of both networks focus on similar input areas which shows that the self-supervised representations are useful for the end-task.}}
\label{fig:saliency}
\end{figure}

Figure~\ref{fig:saliency} provides a saliency mapping of the same input produced by the two networks for a class with the highest score. To aid interpretability of the saliency score, we calculate a magnitude of each tri-axial accelerometer sample, effectively combining all three channels. The actual input is given in the top-most pane, the magnitudes with varying color intensity are shown in the bottom panes. The dark color illustrates the regions that contribute most to the network's prediction. We observe that the saliency maps of both self-supervised and fully-supervised networks hint towards similar regions that are crucial for deciding on the class label.

Interestingly, for the \textit{Sitting} class instance both network mainly focus on a smaller region of the input with slightly more variation in the values. We think it could be because one thing that a network learns is to find periodic variations in the signal (such as peaks and slopes). Hence, it pays more attention even to slightest fluctuation, but it decides on the \textit{Sitting} label as the signal remains constant (before and after minor changes) which is an entirely different pattern as compared to the instances of other classes. This analysis further validates the point that our self-supervised network learns generalizable features for activity classification.

\subsubsection*{Visualization of High-Level Feature Space through t-SNE}
\begin{figure}[htbp]
\centering
\subfloat{\includegraphics[width=.4\textwidth]{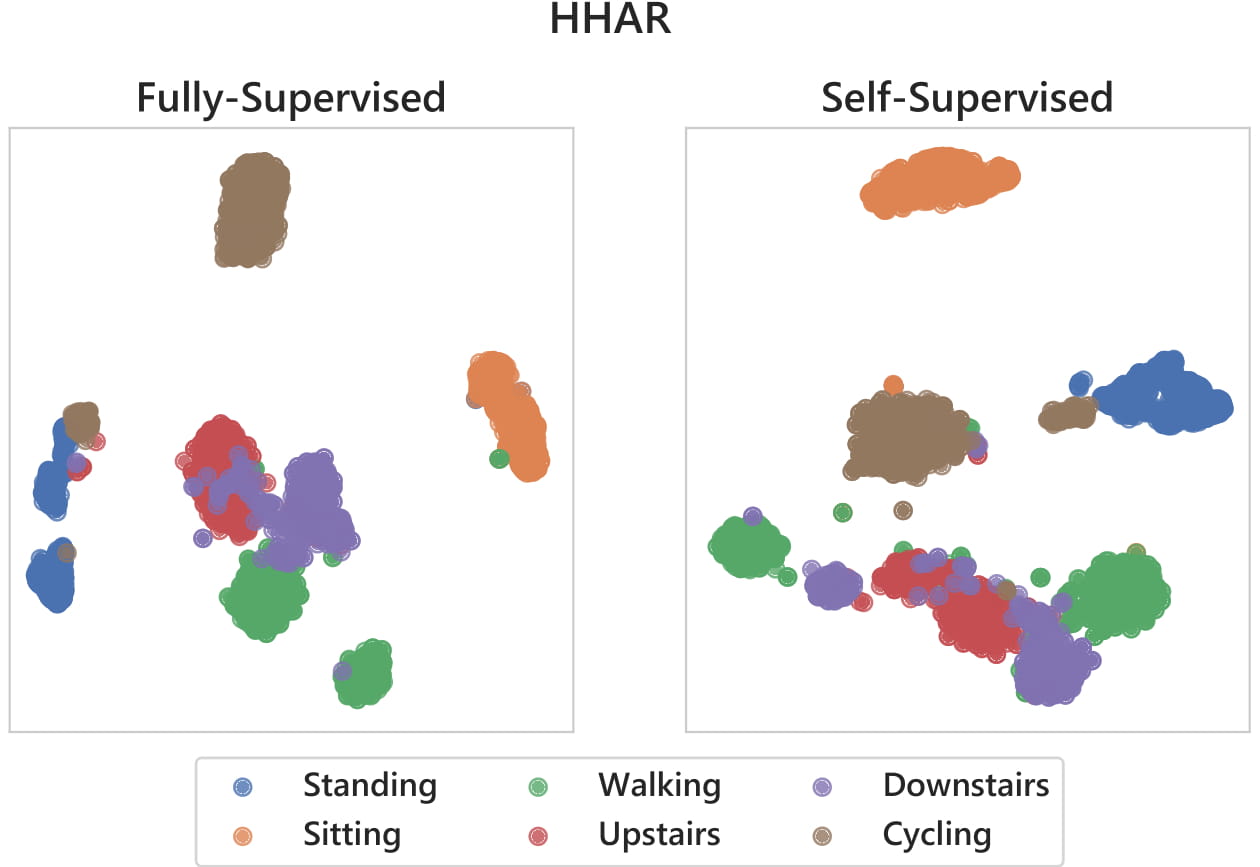}} \hspace{0.01cm}
\subfloat{\includegraphics[width=.4\textwidth]{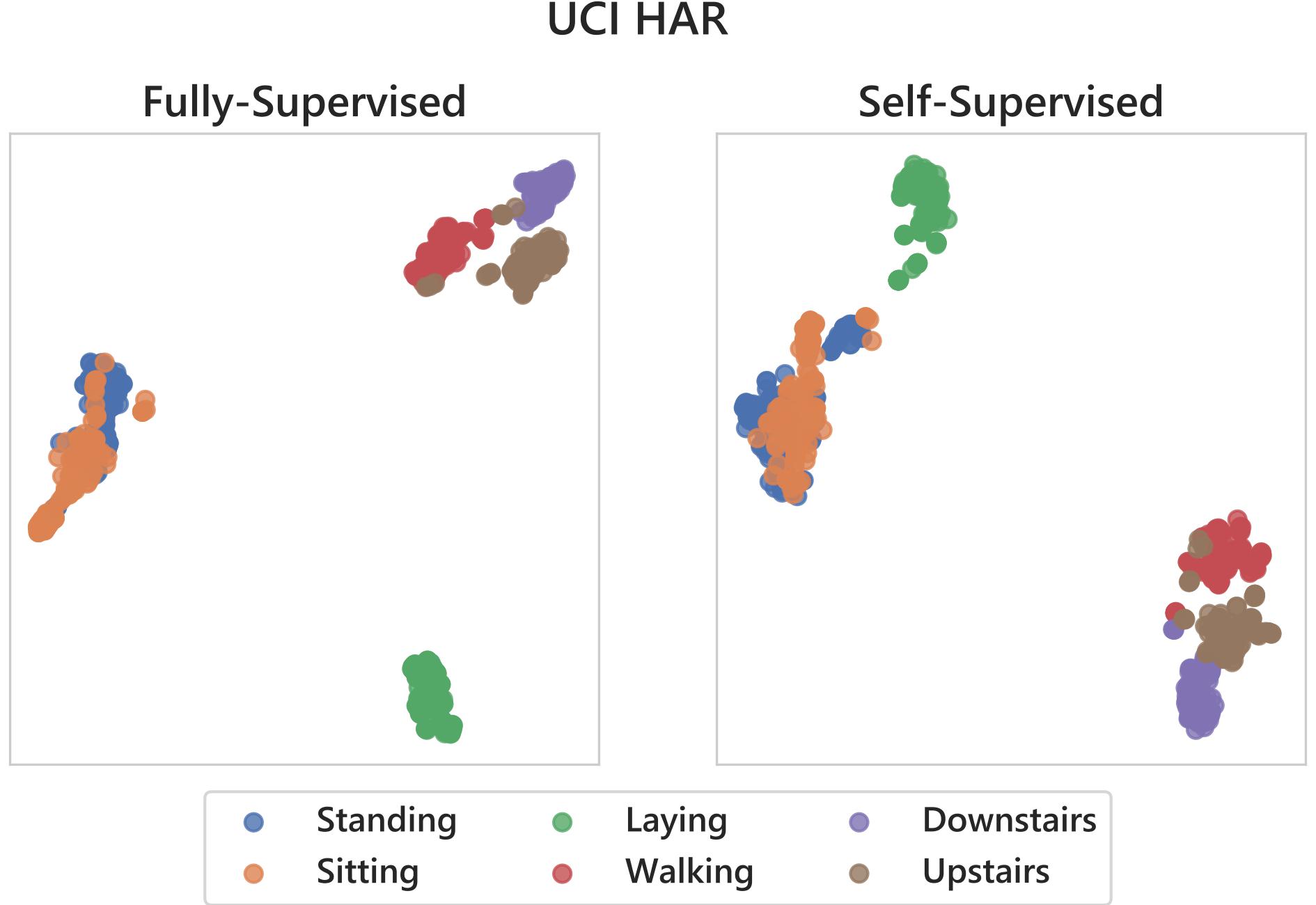}} \\
\subfloat{\includegraphics[width=.4\textwidth]{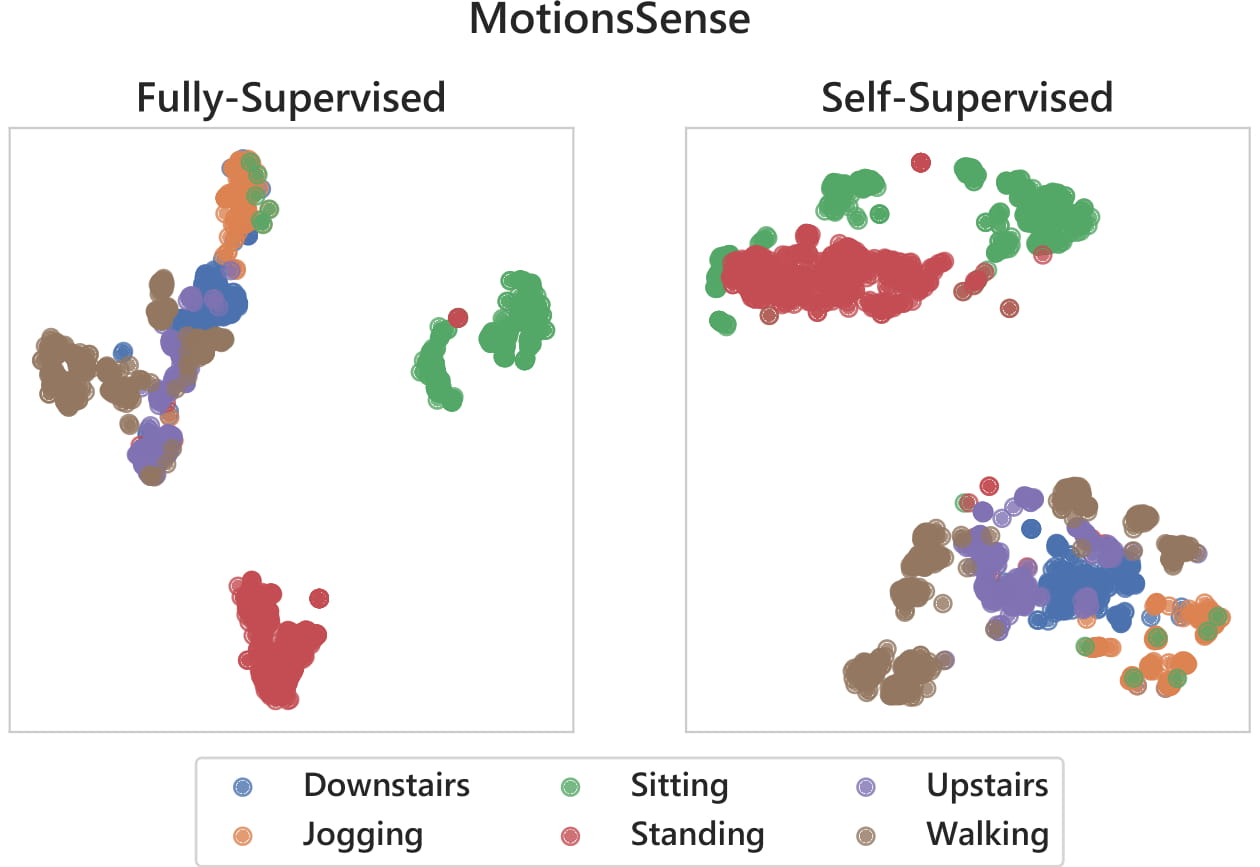}} \hspace{0.01cm}
\subfloat{\includegraphics[width=.4\textwidth]{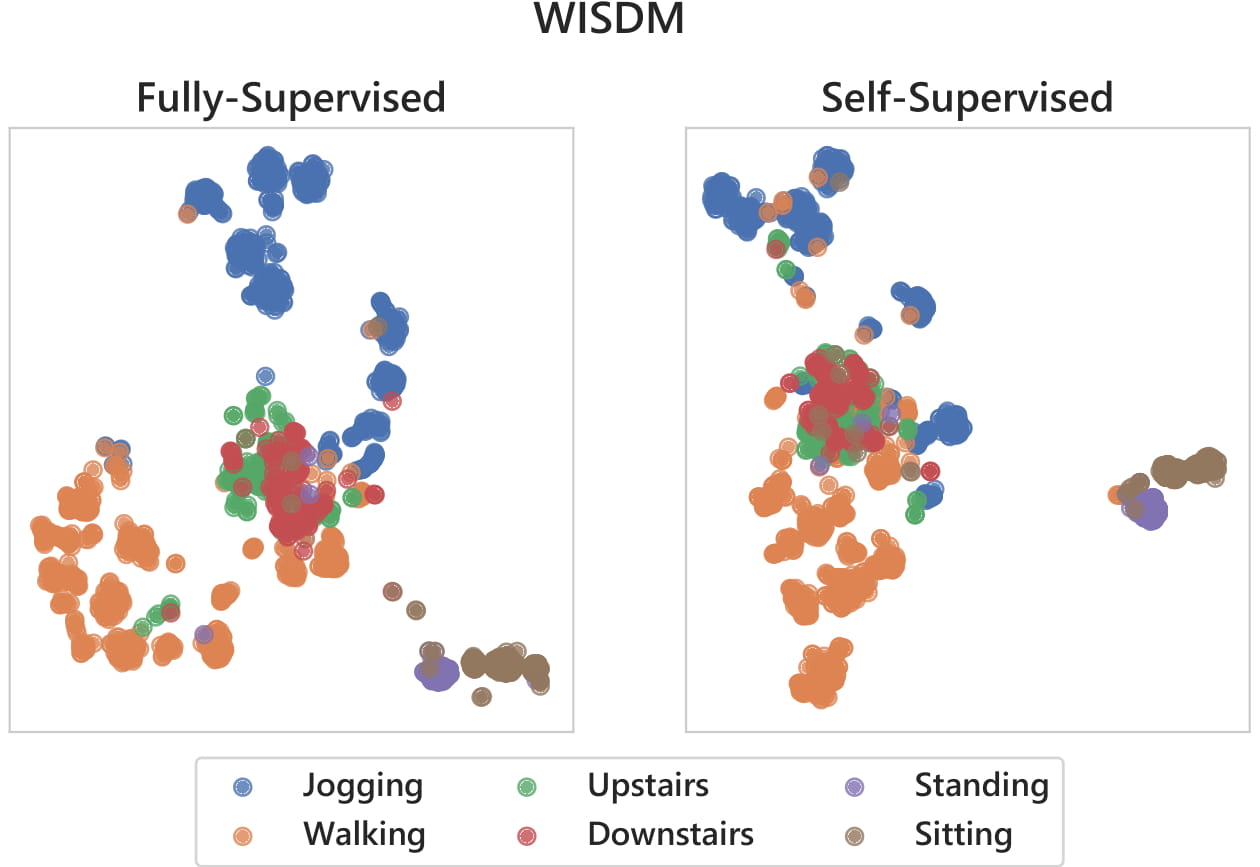}} \\
\subfloat{\includegraphics[width=.4\textwidth]{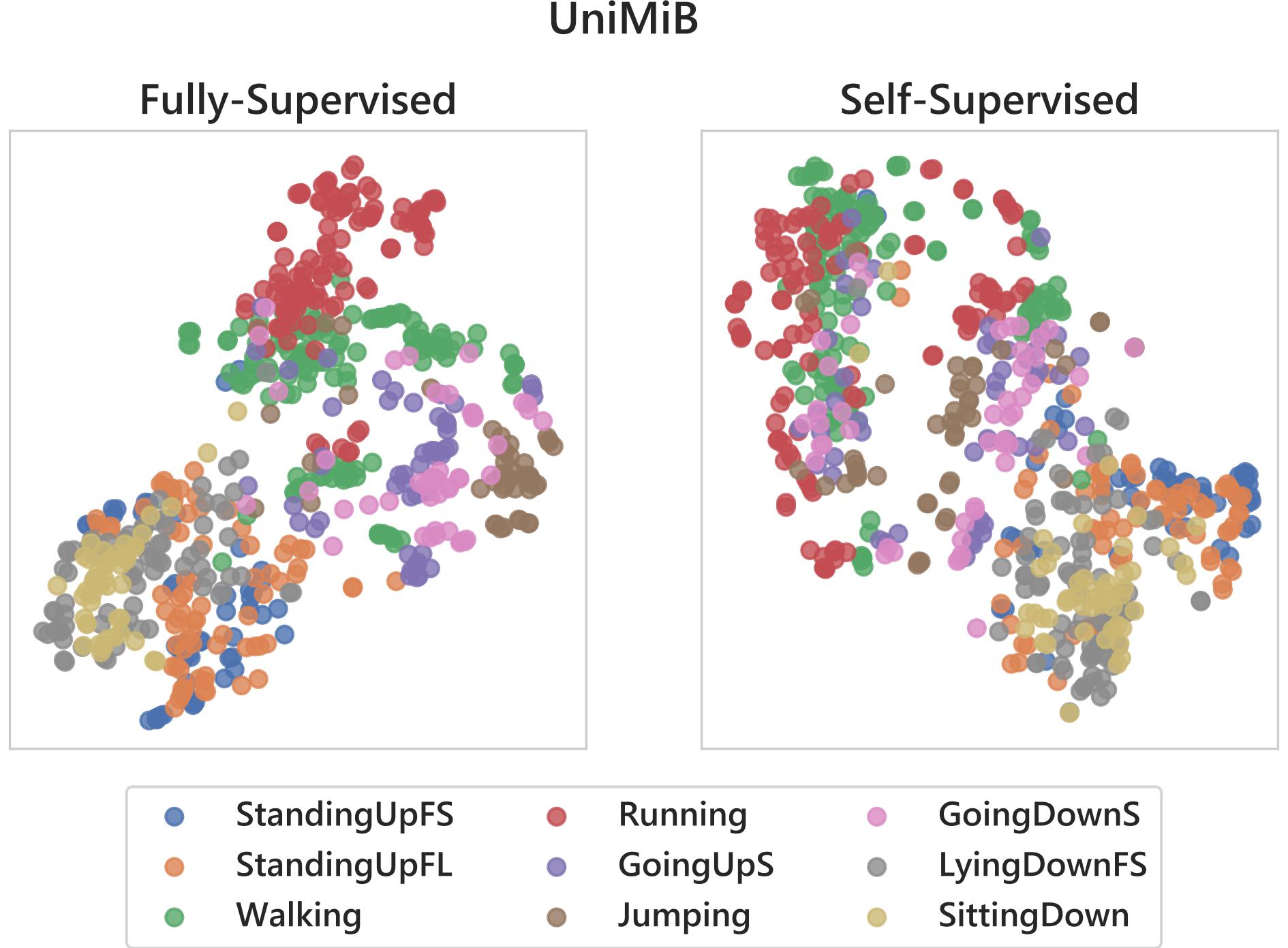}} \hspace{0.01cm}
\subfloat{\includegraphics[width=.4\textwidth]{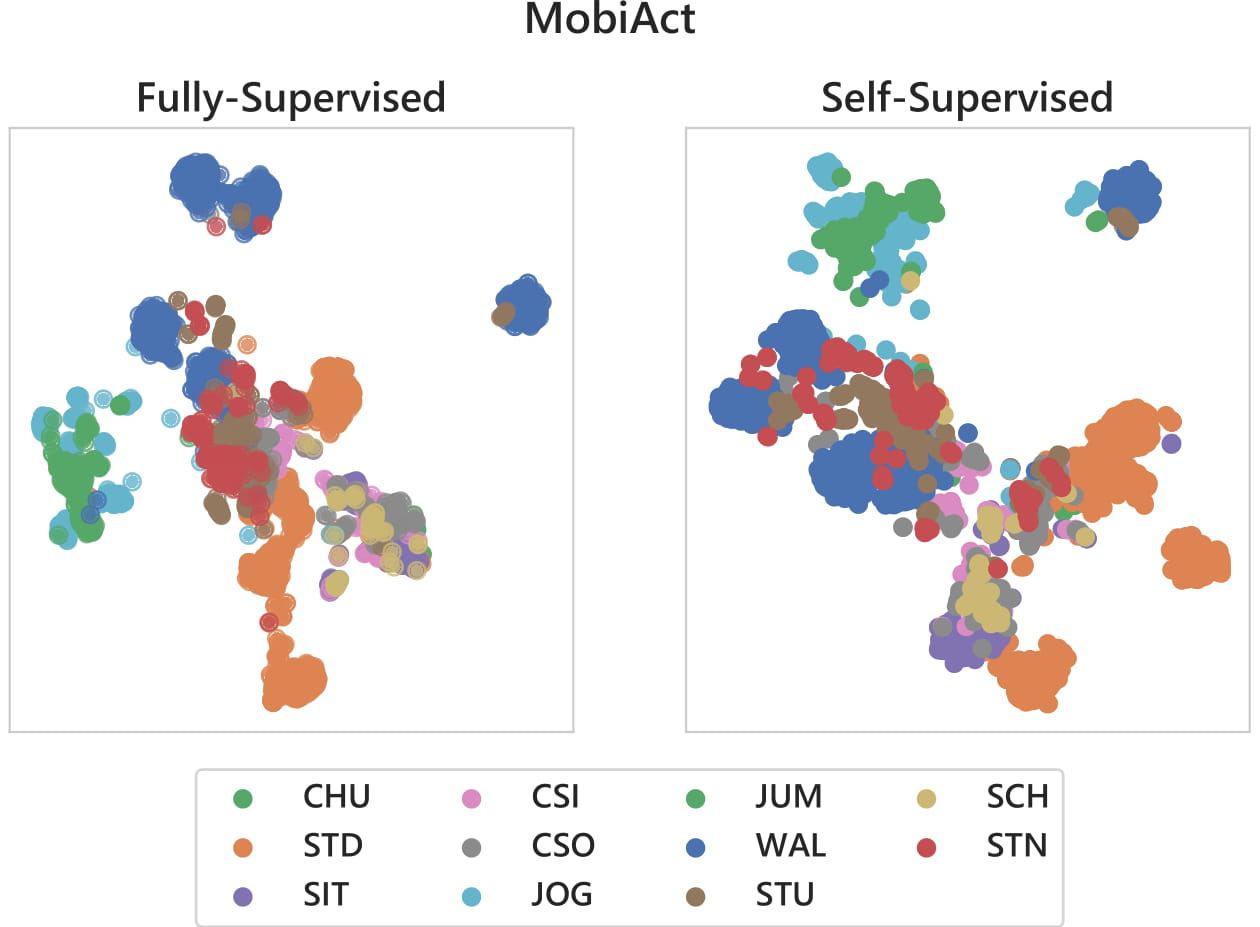}} 
\caption{t-SNE visualization of the learned representations. \small{We visualize the features from \textit{Global Max Pooling} layers of fully-supervised and self-supervised networks by projecting them on $2$D space. The clusters show high correspondence among the representations across datasets. For instance, in \textit{UniMiB} embeddings the samples belonging to the same class are close-by as opposed to those from a different class, such as \textit{Running} and \textit{Walking} are alongside each other while data point from \textit{SittingDown} class are very far. Note that t-SNE embeddings do not use activity labels, they are only used for final visualizations.}}
\label{fig:tsne}
\end{figure}

t-SNE is a non-linear technique for exploring and visualizing multi-dimensional data~\cite{maaten2008visualizing}. It approximates a low-dimensional manifold of a high-dimensional counterpart through minimizing Kullback-Leibler divergence between them with a gradient-based optimization method. More specifically, it maps multi-dimensional data onto a lower dimensional space and discovers patterns in the input through identifying clusters based on the similarity of the data points. Here, the activations from global max-pooling layers (of both self-supervised and fully-supervised networks) with $96$ hidden units are projected on to a $2$D space. Figure~\ref{fig:tsne} provides the t-SNE embeddings showing high semantic relevance of the learned features for various activity classes. We notice that the self-supervised features largely correspond to those learned with the labeled activity data. Importantly, the clusters of data points across two feature learning strategies are similar, e.g. in \textit{UCI HAR}, the activity classes like \textit{Upstairs}, \textit{Downstairs} and \textit{Walking} are grouped. Likewise, in \textit{HHAR}, the data points for \textit{Walking}, \textit{Upstairs}, and \textit{Downstairs} are close-by as opposed to others in the embeddings of both networks. Finally, it is important to note that t-SNE is an unsupervised technique which does not use class labels; the activity labels are just used for final visualization. 

\section{Related Work}
\label{sec:related_work}
Deep learning methods have been successfully used in several applications of ubiquitous computing,  pervasive intelligence, health, and well-being~\cite{radu2018multimodal, georgiev2017low, saeed2017personalized, Hannun2019, liu2016deepfood, yao2018sensegan} and eliminate the need of hand-crafted feature engineering. Convolutional and recurrent neural networks have shown dominant performance in solving numerous high-level recognition tasks from temporal data such as activity detection and stress recognition~\cite{wang2018deep, hammerla2016deep, saeed2017personalized}. In particular, CNNs are becoming increasingly popular in sequence (or time-series) modeling due to their ability of weight sharing, translation invariance, scale separation and localization of filters in space and time~\cite{bai2018empirical, LeCun2015}. In fact, ($1$D) temporal CNNs are now widely used in the area of HAR (see~\cite{wang2018deep} for a detailed review), but the prior works are mostly concerned with supervised learning approaches. The training of deep networks requires a huge (carefully) curated dataset of labeled instances, which in several domains is infeasible due to required manual labeling effort or can only be possible on a small-scale in a controlled lab environment. This inherent limitation of the fully-supervised learning paradigm emphasizes the importance of unsupervised learning to leverage a large amount of unlabeled data for representation learning~\cite{bengio2013representation} that can be easily acquired in a real-world setting. 

Unsupervised learning has been well-studied in the literature over the past years. Before the era of end-to-end learning, manual feature design strategies~\cite{figo2010preprocessing} such as those that employ statistical measures have been used with clustering algorithms to discover a latent group of activities~\cite{7321473}. Although deep learning techniques have almost entirely replaced hand-crafted feature extraction with directly learning rich features from data, representation learning still stands as a fundamental problem in machine learning (see \cite{bengio2013representation} for an in-depth review). The classical approaches for unsupervised learning include autoencoders~\cite{baldi2012autoencoders}, restricted Boltzmann machines~\cite{nair2010rectified}, and convolutional deep belief networks~\cite{lee2009convolutional}. Another emerging line of research for unsupervised feature learning (also studied in this work), which has shown promising results and does not require manual annotations, is self-supervised learning~\cite{raina2007self, doersch2015unsupervised,agrawal2015learning}. These methods exploit the inherent structure of the data to acquire a supervisory signal for solving a pretext task with reliable and widely used supervised learning schemes. 

Self-supervision has been actively studied recently in the vision domain, and several surrogate tasks have been proposed for learning representations from static images, videos, sound, and in robotics~\cite{noroozi2016unsupervised, owens2016ambient, gomez2017self, zhang2017split, larsson2017colorization, jenni2018self, gidaris2018unsupervised, lee2017unsupervised, doersch2017multi, fernando2017self, misra2016shuffle, arandjelovic2017objects, owens2018audio, pathak2017curiosity}. For example, in images and videos,  spatial and temporal contexts, respectively, provide forms of rich supervision to learn features. Similarly, colorization of gray-scale images~\cite{larsson2017colorization,zhang2017split}, rotation classification~\cite{gidaris2018unsupervised}, odd sequence detection~\cite{fernando2017self}, frame order prediction~\cite{misra2016shuffle}, learning the arrow of time~\cite{wei2018learning}, audio-visual correspondence~\cite{owens2016ambient, arandjelovic2017objects} and synchronization~\cite{korbar2018cooperative, owens2018audio} are some of the recently explored directions of self-supervised techniques. Furthermore, multiple such tasks are utilized together in a multi-task learning setting for solving diverse visual recognition problems~\cite{doersch2017multi}. These self-supervised learning paradigms have shown to extract high-level representations that are on par with those acquired through fully-supervised pre-training techniques (e.g., with ImageNet labels) and they tremendously help with transfer and semi-supervised learning scenarios. Inspired from this research direction, we explore multi-task self-supervision for learning representations from sensory data through utilizing transformations of the signals. 

Some earlier works on time-series analysis have explored transformations to exploit invariances either through architectural modifications (to automatically learn task-relevant variations) or less commonly with augmentation and synthesis. In~\cite{um2017data} task-specific transformations (such as added noise and rotation) are applied to wearable sensor data to augment and improve the performance of Parkinson's disease monitoring systems. Saeed et al.~\cite{saeed2018synthesizing} utilized an adversarial autoencoder for class-conditional (multimodal) synthetic data generation for the behavioral context in a real-life setting. Moreover, Oh et al.~\cite{oh2018learning} focused on learning invariances directly from clinical time-series data with specialized neural network architecture. Razavian et al.~\cite{razavian2016multi} used convolution layers of varying size filters to capture different resolutions of temporal patterns. Similarly, through additional pre-processing of the original data Cui et al.~\cite{cui2016multi} used transformed signals as extra channels to the model for learning multiscale features. To summarize, these works are geared towards learning supervised networks for specific tasks through exploiting invariances, but they do not address the topics of semi-supervised and unsupervised learning.

To the best of our knowledge, the work presented here is the first attempt of self-supervision for sensor representation learning, in particular for HAR. Our work differs from the aforementioned works in several ways as we learn representations with self-supervision from completely unlabeled data and without using any specialized architecture. We show that when training a CNN to predict generally known (time-series) transformations~\cite{batista2011complexity, um2017data} as a surrogate task, the model can learn features that are on a par with a fully-supervised network and far better than unsupervised pre-training with an autoencoder. We also demonstrate that the learned representations from a different (but related) unlabeled data source can be successfully transferred to improve the performance of diverse tasks even in the case of semi-supervised learning. In terms of transfer learning, our approach also differs significantly from some earlier attempts~\cite{morales2016deep, wang2018deep(b)} that were concerned with features transferability from a fully-supervised model learned from inertial measurement units data, as our approach utilizes widely available smartphones and does not require labeled data. Finally, the proposed technique is also different from previously studied unsupervised pre-training methods such as autoencoders~\cite{li2014unsupervised}, restricted Boltzmann machines~\cite{plotz2011feature} and sparse coding~\cite{bhattacharya2014using} as we employ an end-to-end (self) supervised learning paradigm on multiple surrogate tasks to extract features. 

\section{Conclusions and Future Work}
We present a novel approach for self-supervised sensor representation learning from unlabeled data with a focus on smartphone-based human activity recognition (HAR). We train a multi-task temporal convolutional network to recognize potential transformations that may have been applied to the raw input signal. Despite the simplicity of the proposed self-supervised technique (and the network architecture), we show that it enables the convolutional model to learn high-level features that are useful for the end-task of HAR. We exhaustively evaluate our approach under unsupervised learning, semi-supervised learning and transfer learning settings on several publicly available datasets. The performance we achieve is consistently superior to or comparable with fully-supervised methods, and it is significantly better than traditional unsupervised learning methods such as an autoencoder. Specifically, our self-supervised framework drastically improved the detection rate under semi-supervised learning setting, i.e., when very few labeled instances are available for learning. Likewise, the transferred features learned from a different but related unlabeled dataset (\textit{MobiAct} in our case), further improves the performance in comparison with merely training a model from scratch. Notably, these transferred representations even boost the performance of an activity recognizer in semi-supervised learning from a dataset (or task) of interest. Finally, canonical correlation analysis, saliency mapping, and t-SNE visualizations show that the representations of the self-supervised network are very similar to those learned by a fully-supervised model that is trained in an end-to-end fashion with activity labels. We believe that, through utilizing more sophisticated layers and deep architectures, the presented approach can further reduce the gap between unsupervised and supervised feature learning. 

In this work, we provided the basis for self-supervision of HAR with smartphones through a few labeled data. In the Internet of Things era, there are many exciting opportunities for future works in related areas, such as in industrial manufacturing, electrical grid, smart wearable technologies, and home automation. In particular, we believe that self-supervision is of immense value for automatically extracting generalizable representations in domains, where labeled data are challenging to acquire, but unlabeled data are available in vast quantities. We hope that the presented perspective of self-supervision inspires the development of additional approaches, specifically for the selection of appropriate auxiliary tasks (based on domain expertise) that enables the network to learn useful features to solve a particular problem. Likewise, combining self-supervision with network architecture search is another crucial area of improvement that will automate the process of optimal model discovery. Another exciting avenue for future research is evaluating self-supervised representations on an imbalanced activity dataset, where, the number of classes are high and collecting a few labeled data points for each activity class is not feasible. Finally, evaluation in a real-world setting (application deployed on real devices) is of prime importance to further understand the aspects that need improvement concerning computational, energy and, labeled data requirements.

\begin{acks}
This work is funded by SCOTT (www.scott-project.eu) project. It has received funding from the Electronic Component Systems for European Leadership Joint Undertaking under grant agreement No 737422. This Joint Undertaking receives support from the European Union's Horizon 2020 research and innovation programme and Austria, Spain, Finland, Ireland, Sweden, Germany, Poland, Portugal, Netherlands, Belgium, Norway. \\
\noindent We thank Prof. Peter Baltus for a helpful discussion and anonymous reviewers for their insightful comments and suggestions. Various icons used in the figures are created by Anuar Zhumaev, Korokoro, Gregor Cresnar, Becris, Hea Poh Lin, AdbA Icons, Universal Icons, and Baboon designs from the Noun Project. 
\end{acks}

\bibliographystyle{plainnat}
\bibliography{main}

\newpage
\section*{Appendix}

\begin{table}[!ht]
   \caption{Evaluating self-supervised representation with (user-split based) 5-folds cross-validation for activity recognition. \small{We perform this assessment based on user-split of the data with no overlap between training and test sets i.e. distinct users' data are used for training and testing of the models. The reported results are averaged over $5$-folds.}}
   \centering
   \subfloat[HHAR]{
     \scriptsize
     \centering
     \begin{tabular}{l|cccc}
& \textbf{P}    & \textbf{R}    & \textbf{F}    & \textbf{K}    \\ \hline
Random Init.         & 0.3429$\pm$0.1395&0.302$\pm$0.0465&0.2023$\pm$0.0333&0.1611$\pm$0.0536 \\
Supervised           & 0.8403$\pm$0.0349&0.816$\pm$0.0518&0.8076$\pm$0.0612&0.7788$\pm$0.0624 \\
Autoencoder          & 0.6068$\pm$0.2149&0.594$\pm$0.1858&0.5474$\pm$0.2182&0.5159$\pm$0.2208 \\
Self-Supervised      & 0.8454$\pm$0.0378&0.8239$\pm$0.0462&0.8153$\pm$0.053&\textbf{0.7881}$\pm$\textbf{0.0556}  \\
Self-Supervised (FT) & 0.8439$\pm$0.0753&0.8101$\pm$0.1004&0.8038$\pm$0.1072&0.7719$\pm$0.1204
\end{tabular}
   } \\
   \subfloat[UniMiB]{
     \scriptsize
     \centering
    \begin{tabular}{l|cccc}
 & \textbf{P}    & \textbf{R}    & \textbf{F}    & \textbf{K}    \\ \hline
Random Init.         & 0.416$\pm$0.0307&0.3615$\pm$0.0503&0.2875$\pm$0.0722&0.2281$\pm$0.0622 \\
Supervised           & 0.805$\pm$0.0095&0.7899$\pm$0.0153&0.7866$\pm$0.0165&0.7576$\pm$0.0181 \\
Autoencoder          & 0.5989$\pm$0.0313&0.5743$\pm$0.0192&0.5494$\pm$0.0227&0.5009$\pm$0.0242 \\
Self-Supervised      & 0.77$\pm$0.0211&0.7618$\pm$0.0191&0.7577$\pm$0.0208&0.724$\pm$0.0218 \\
Self-Supervised (FT) & 0.8396$\pm$0.0226&0.8311$\pm$0.0269&0.8285$\pm$0.0283&\textbf{0.8046}$\pm$\textbf{0.0309}
\end{tabular}
 } \\
   \subfloat[UCI HAR]{
     \scriptsize
     \centering
     \begin{tabular}{l|cccc}
& \textbf{P}    & \textbf{R}    & \textbf{F}    & \textbf{K}    \\ \hline
Random Init.         & 0.6147$\pm$0.1845&0.5019$\pm$0.0999&0.4297$\pm$0.1141&0.3872$\pm$0.1277 \\
Supervised           & 0.9068$\pm$0.0332&0.9035$\pm$0.0366&0.903$\pm$0.0377&0.8827$\pm$0.0446 \\
Autoencoder          & 0.8745$\pm$0.0367&0.8485$\pm$0.0604&0.8461$\pm$0.0642&0.8161$\pm$0.0734 \\
Self-Supervised      & 0.9054$\pm$0.0273&0.8919$\pm$0.0388&0.889$\pm$0.0444&0.8688$\pm$0.0472 \\
Self-Supervised (FT) & 0.9125$\pm$0.0403&0.906$\pm$0.0473&0.9046$\pm$0.049&\textbf{0.8859}$\pm$\textbf{0.0573}
\end{tabular}
   } \\
   \subfloat[MobiAct]{
     \scriptsize
     \centering
\begin{tabular}{l|cccc}
 & \textbf{P}    & \textbf{R}    & \textbf{F}    & \textbf{K}    \\ \hline
Random Init.         & 0.4814$\pm$0.1405&0.3828$\pm$0.0417&0.3018$\pm$0.0422&0.191$\pm$0.0435\\
Supervised           & 0.9121$\pm$0.0118&0.9029$\pm$0.016&0.9043$\pm$0.0153&0.876$\pm$0.0198\\
Autoencoder          & 0.7488$\pm$0.0402&0.749$\pm$0.0398&0.7323$\pm$0.0408&0.6692$\pm$0.0542\\
Self-Supervised      & 0.8977$\pm$0.0128&0.8838$\pm$0.0133&0.8868$\pm$0.013&0.8521$\pm$0.0165\\
Self-Supervised (FT) & 0.9185$\pm$0.0056&0.9129$\pm$0.0077&0.9127$\pm$0.0067&\textbf{0.8883}$\pm$\textbf{0.0101}
\end{tabular}
   } \\
  \subfloat[WISDM]{
     \scriptsize
     \centering
\begin{tabular}{l|cccc}
& \textbf{P}    & \textbf{R}    & \textbf{F}    & \textbf{K}    \\ \hline
Random Init.         & 0.6074$\pm$0.0555&0.3231$\pm$0.0958&0.3318$\pm$0.0885&0.1828$\pm$0.0623\\
Supervised           & 0.8983$\pm$0.0185&0.8799$\pm$0.0331&0.8846$\pm$0.0297&0.8359$\pm$0.0452\\
Autoencoder          & 0.6878$\pm$0.093&0.6868$\pm$0.0715&0.6671$\pm$0.0797&0.5571$\pm$0.1125\\
Self-Supervised      & 0.8711$\pm$0.0389&0.8369$\pm$0.0636&0.8462$\pm$0.0556&0.7802$\pm$0.0826\\
Self-Supervised (FT) & 0.8842$\pm$0.0285&0.8518$\pm$0.048&0.8597$\pm$0.0409&0.7995$\pm$0.0645
\end{tabular}
  } \\
  \subfloat[MotionSense]{
     \scriptsize
     \centering
\begin{tabular}{l|cccc}
& \textbf{P}    & \textbf{R}    & \textbf{F}    & \textbf{K}    \\ \hline
Random Init.         & 0.5152$\pm$0.0997&0.4237$\pm$0.1006&0.3451$\pm$0.1069&0.2713$\pm$0.1246  \\
Supervised           & 0.9332$\pm$0.0144&0.9219$\pm$0.0186&0.9242$\pm$0.0172&0.9034$\pm$0.0231  \\
Autoencoder          & 0.8297$\pm$0.0658&0.8194$\pm$0.0734&0.818$\pm$0.075&0.7767$\pm$0.0892 \\
Self-Supervised      & 0.9261$\pm$0.0189&0.9176$\pm$0.0195&0.9189$\pm$0.019&0.8979$\pm$0.0244 \\
Self-Supervised (FT) & 0.9476$\pm$0.0174&0.9373$\pm$0.028&0.9391$\pm$0.0254&\textbf{0.9225}$\pm$\textbf{0.0345}
\end{tabular}
  }
\label{tab_appendix:fs_ss_eval}
\end{table}

\begin{figure}[htbp]
\centering
\subfloat{\includegraphics[width=.3\textwidth]{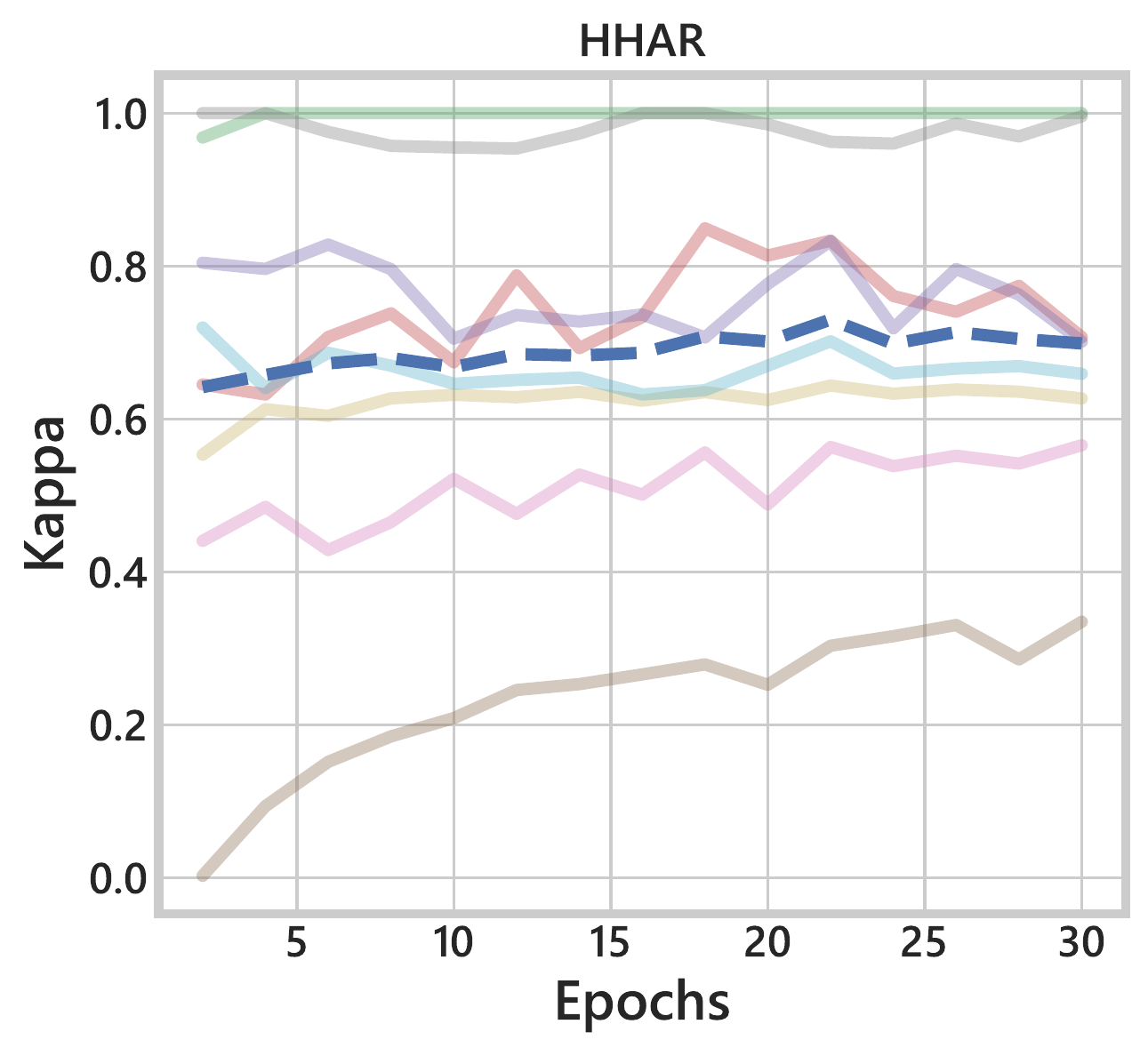}} \hspace{0.01cm}
\subfloat{\includegraphics[width=.3\textwidth]{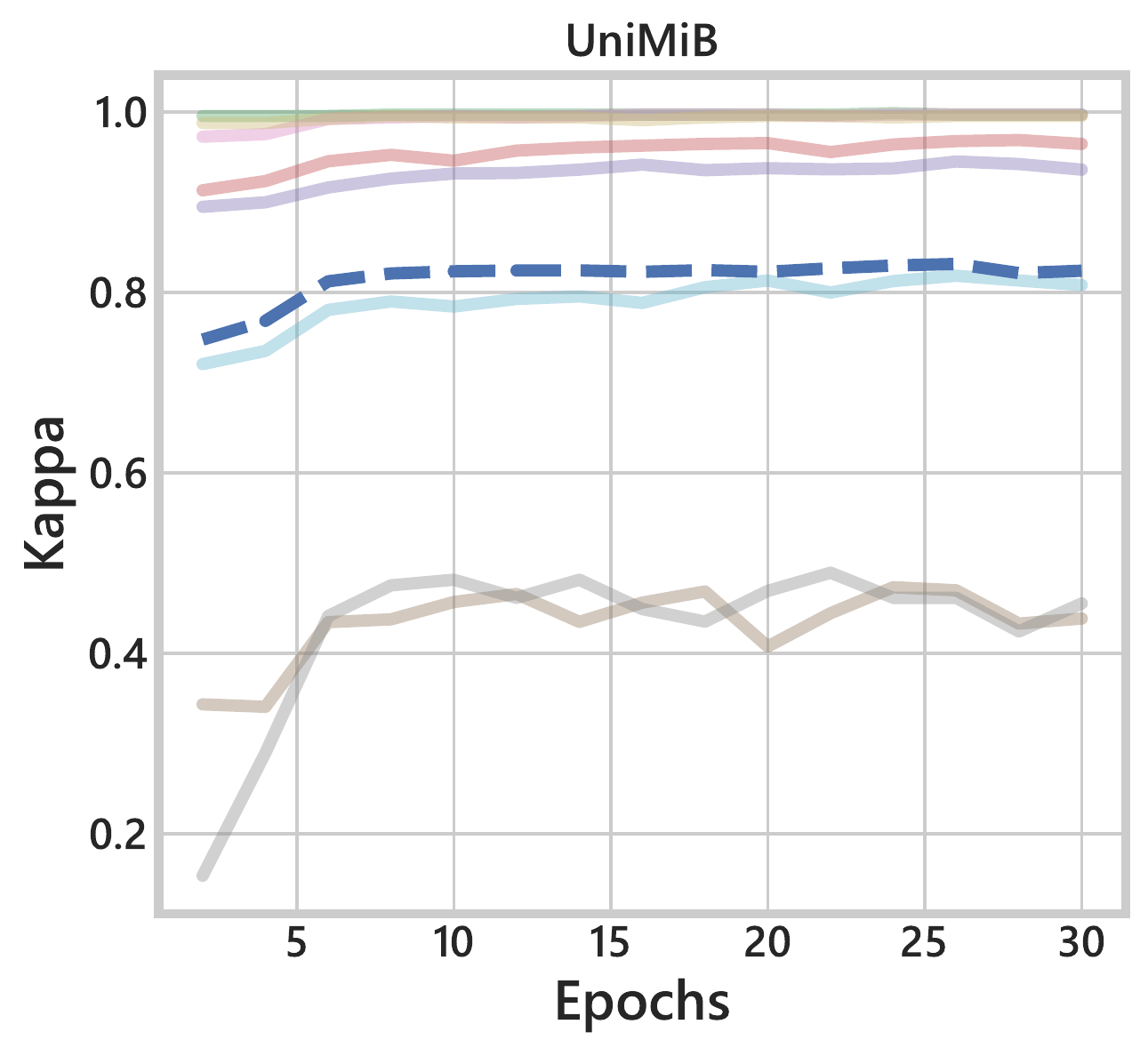}} \hspace{0.01cm}
\subfloat{\includegraphics[width=.3\textwidth]{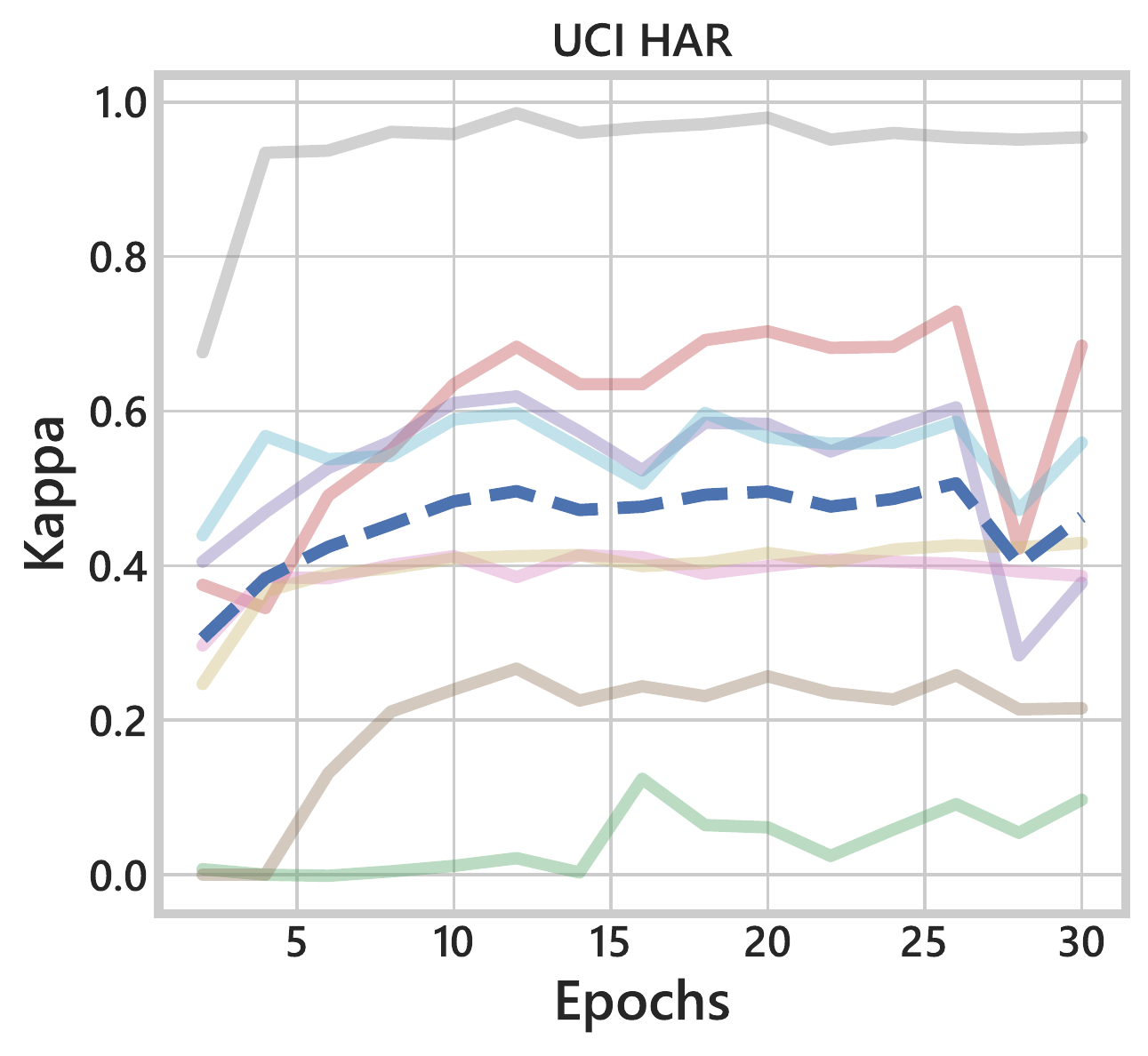}} \hspace{0.01cm} \\
\subfloat{\includegraphics[width=.3\textwidth]{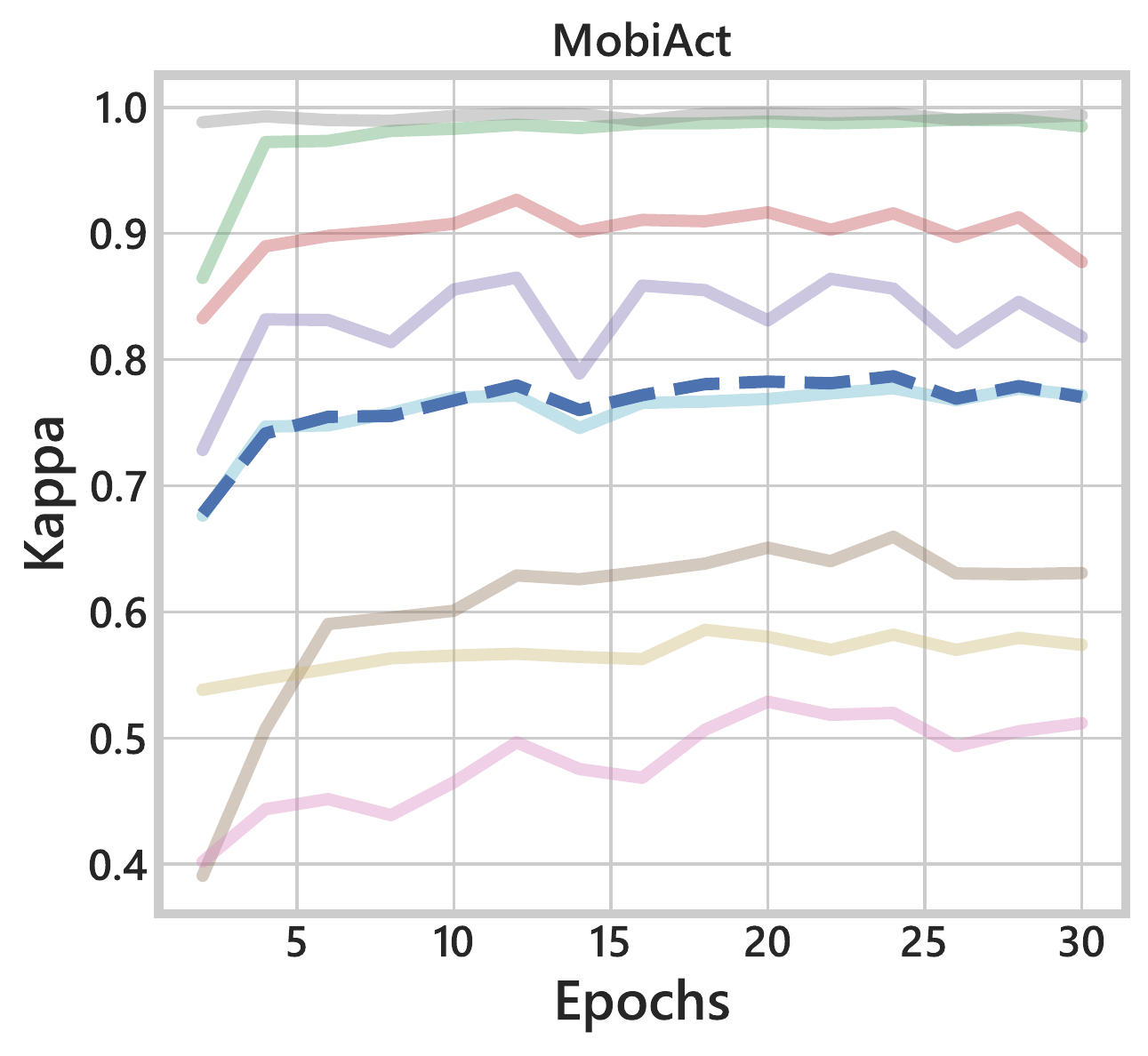}} \hspace{0.01cm}
\subfloat{\includegraphics[width=.3\textwidth]{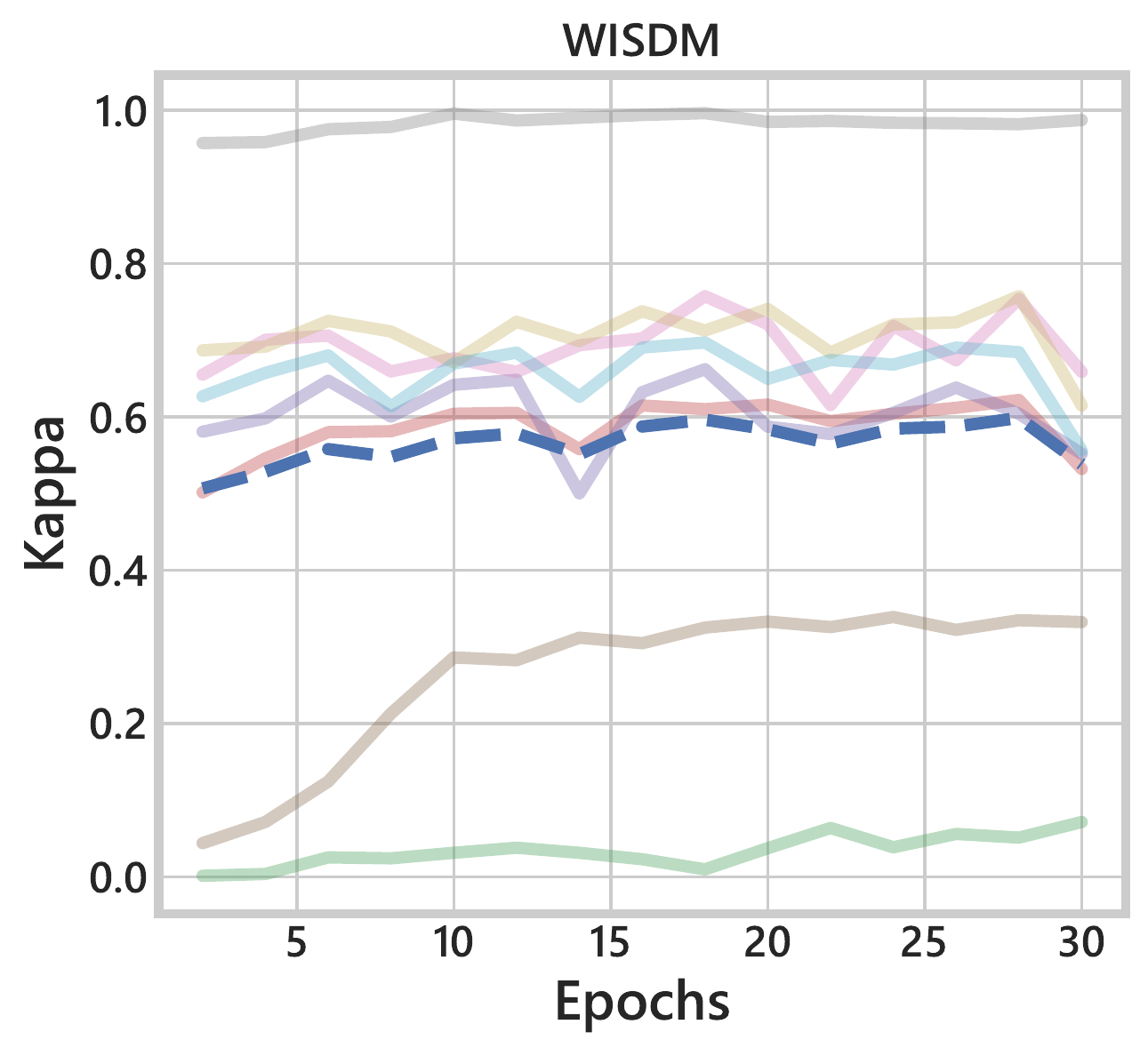}} \hspace{0.01cm}
\subfloat{\includegraphics[width=.3\textwidth]{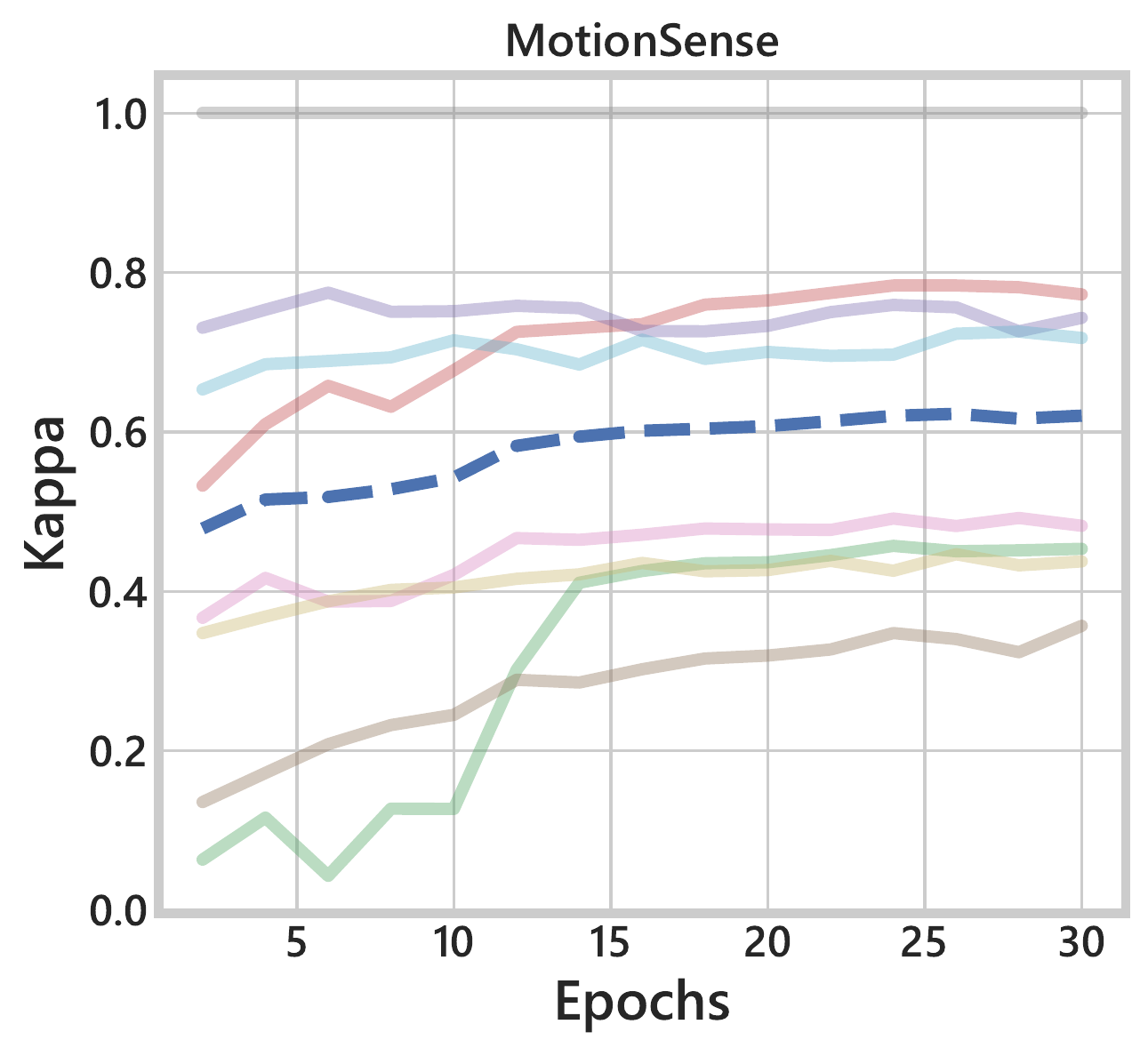}} \\
\subfloat{\includegraphics[width=.4\textwidth]{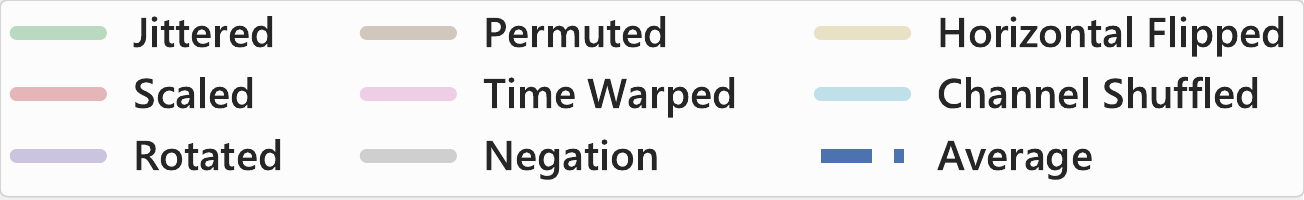}}
\caption{Convergence analysis of transformation recognition tasks. \small{We plot the kappa score of self-supervised tasks (i.e. transformation prediction) as a function of the training epochs. In order to produce the kappa curves, the TPN model's snapshot is saved every second epoch until the defined number of training epochs. For each saved network, we evaluated its performance on the self-supervised data obtained through processing the corresponding test sets. Note that the TPN never sees a test set data in any way during its learning phase.}}
\label{fig_appendix:trf}
\end{figure}

\begin{figure}[htbp]
\centering
\subfloat{\includegraphics[width=.3\textwidth]{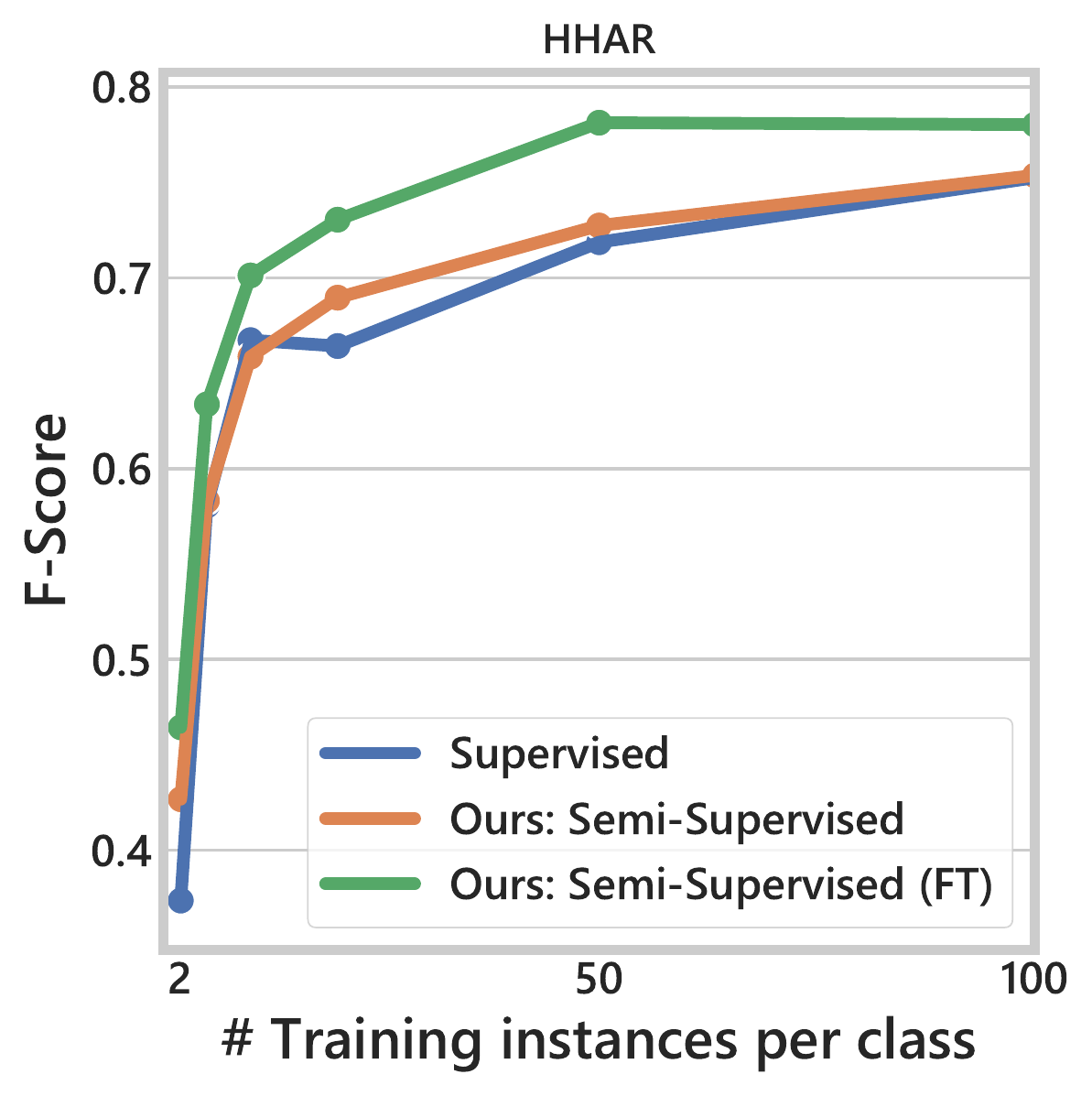}} \hspace{0.01cm}
\subfloat{\includegraphics[width=.3\textwidth]{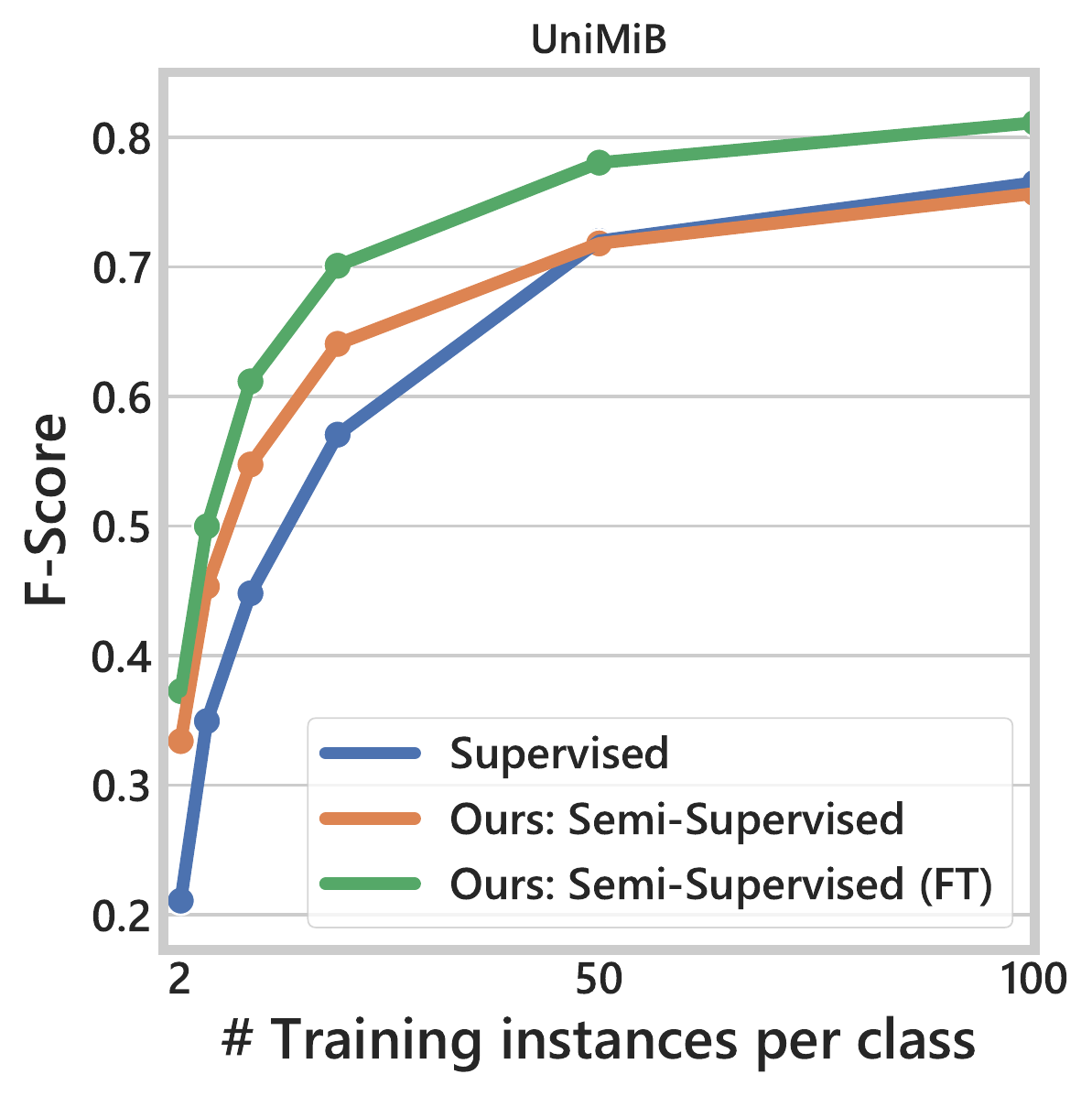}} \hspace{0.01cm}
\subfloat{\includegraphics[width=.3\textwidth]{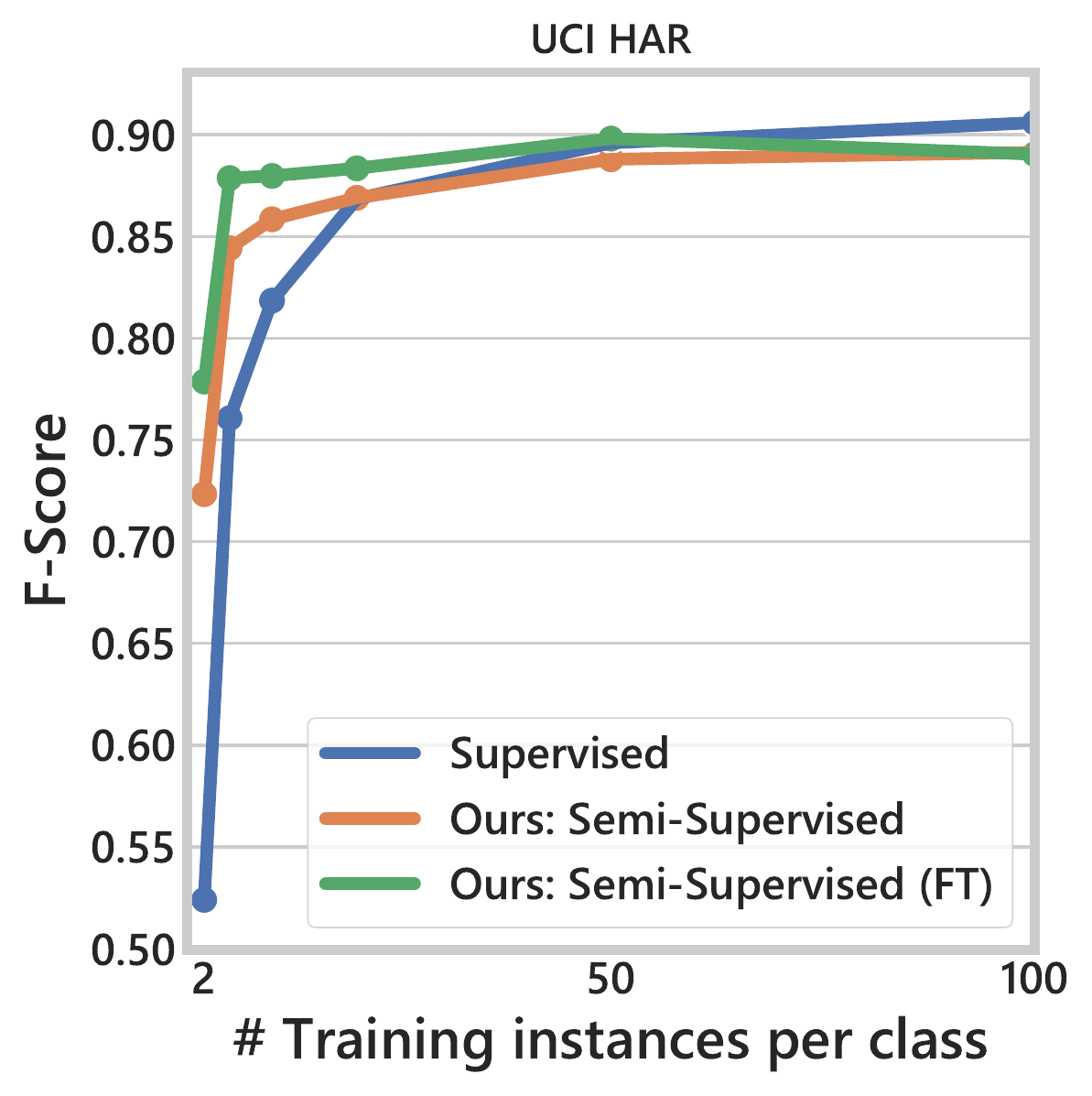}} \hspace{0.01cm}\\
\subfloat{\includegraphics[width=.3\textwidth]{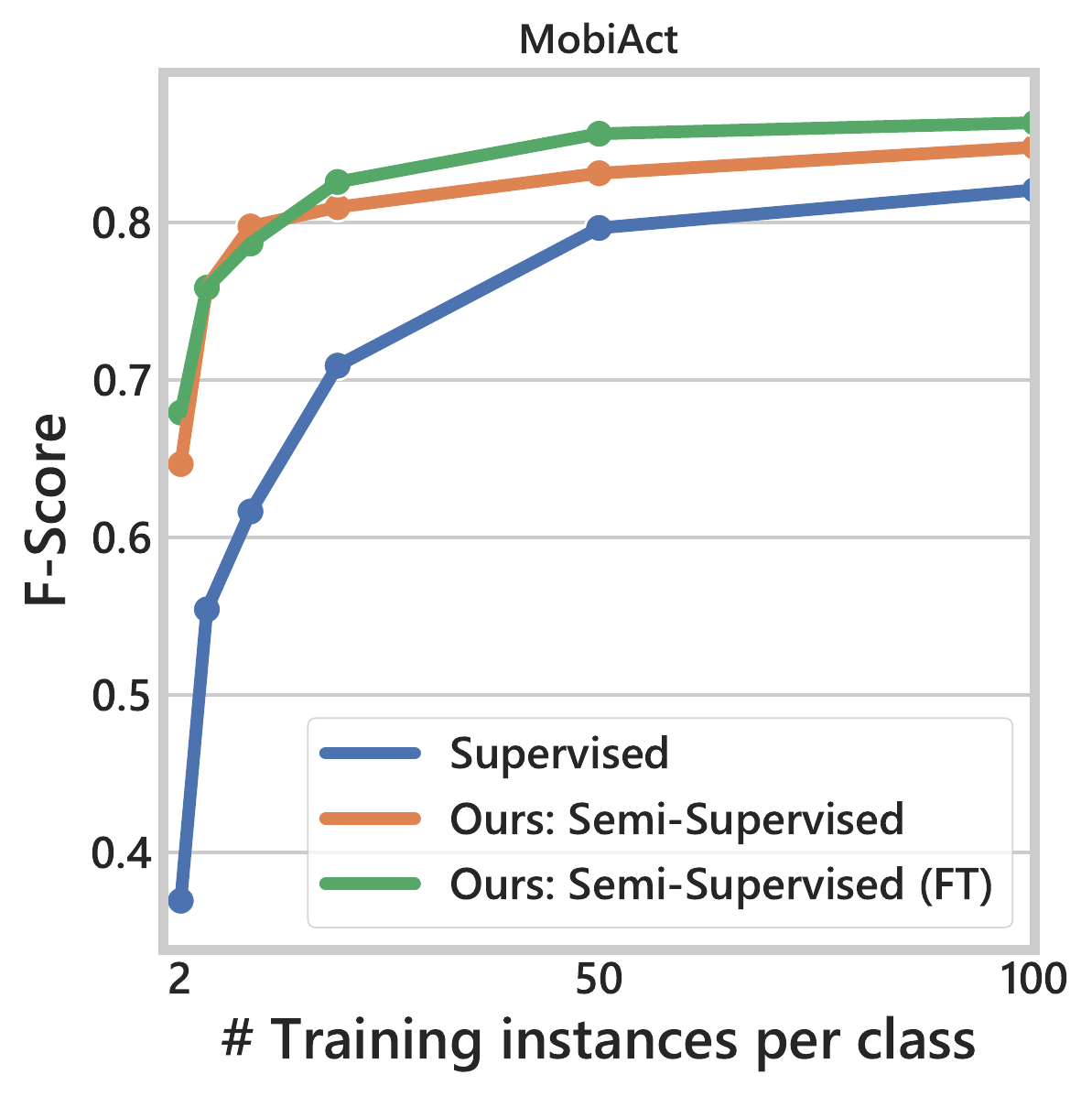}} \hspace{0.01cm}
\subfloat{\includegraphics[width=.3\textwidth]{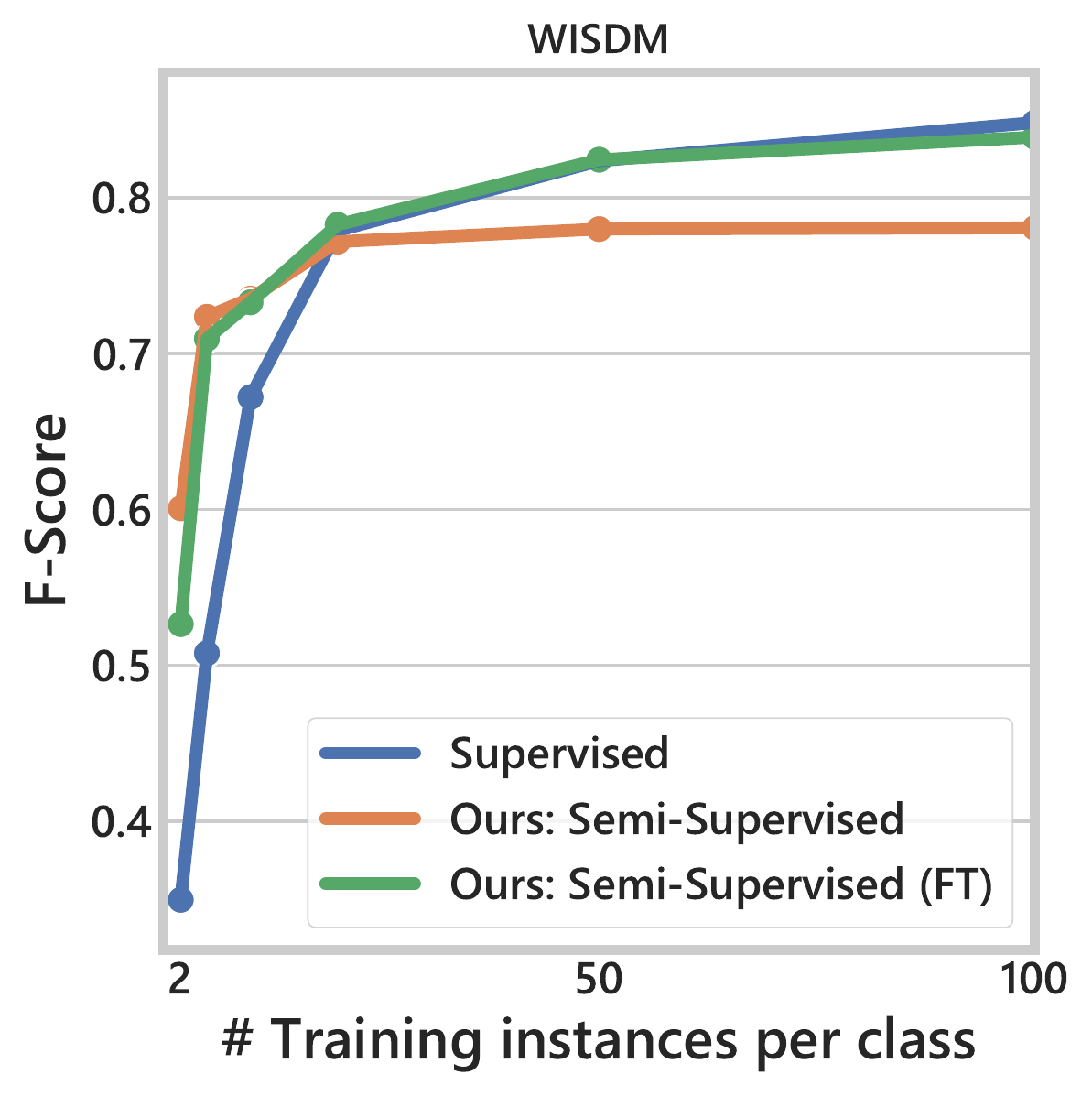}} \hspace{0.01cm}
\subfloat{\includegraphics[width=.3\textwidth]{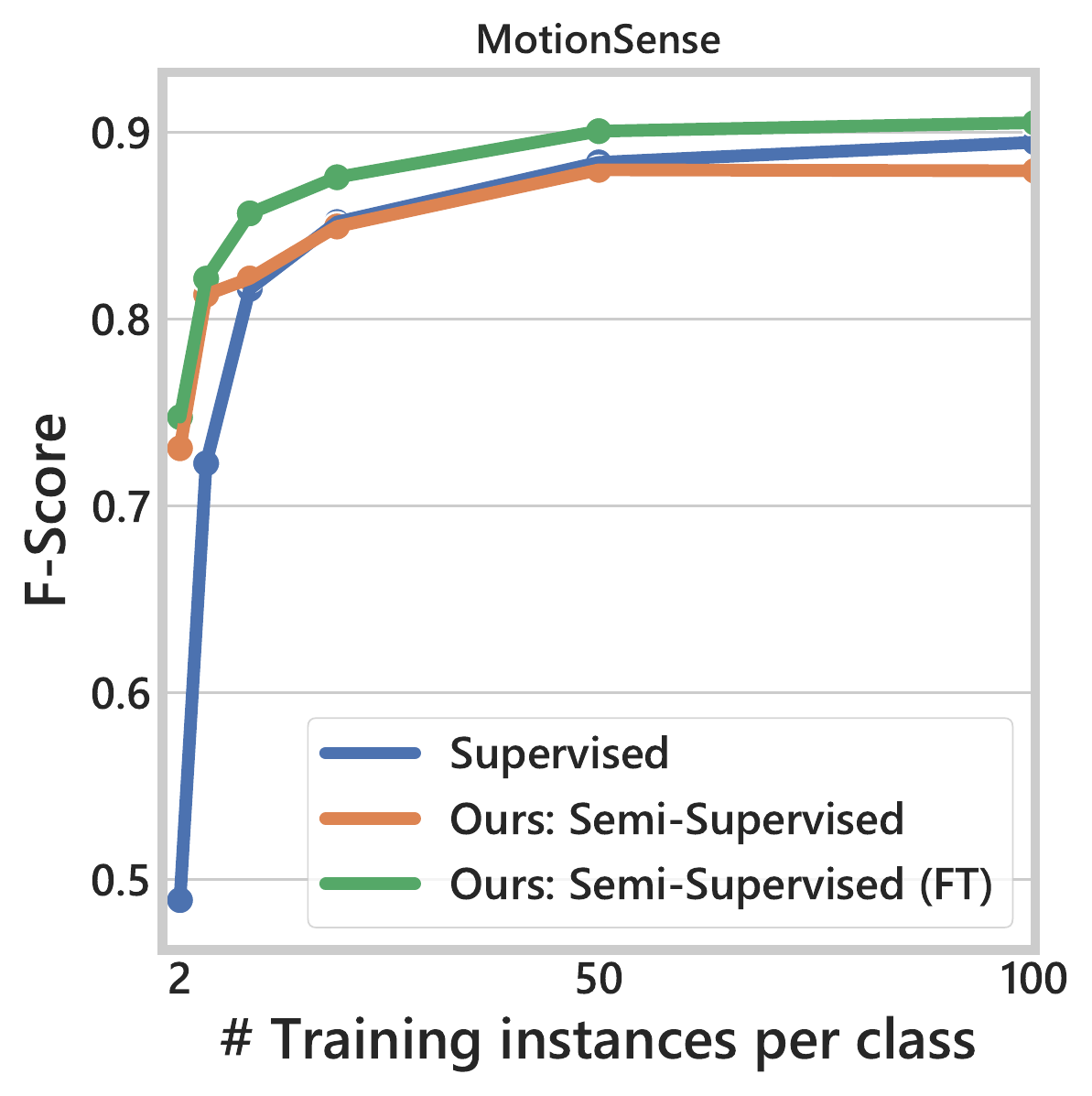}} 
\caption{Weighted F-score: Generalization of the self-supervised learned features under semi-supervised setting. \small{The reported results are averaged over $10$ independent runs for each of the evaluated approaches, for more details see Section~\ref{sec:esss}.}}
\label{fig_appendix:semi_eval_fscore}
\end{figure}

\begin{figure}[htbp]
\centering
\subfloat{\includegraphics[width=.3\textwidth]{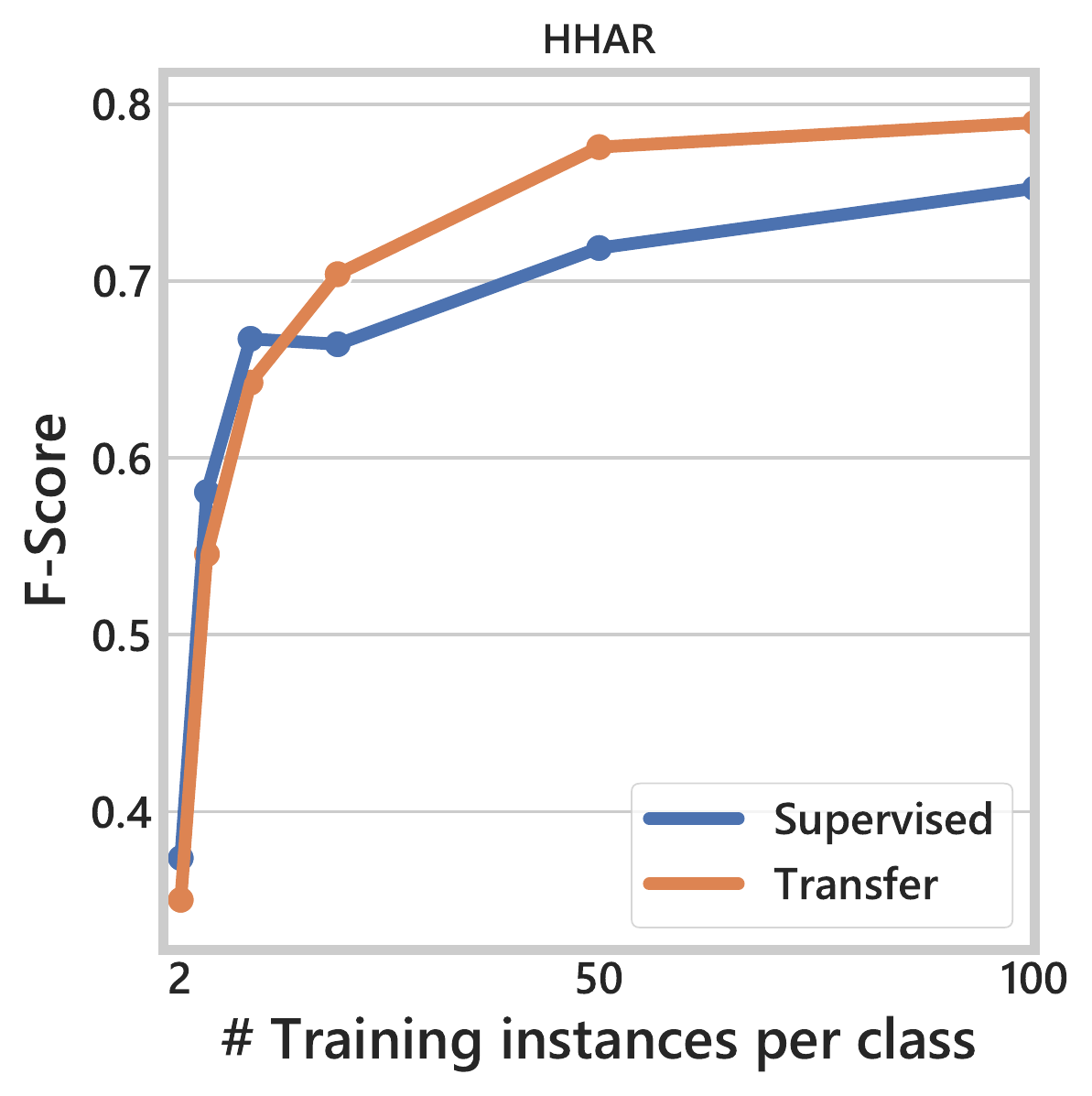}} \hspace{0.01cm}
\subfloat{\includegraphics[width=.3\textwidth]{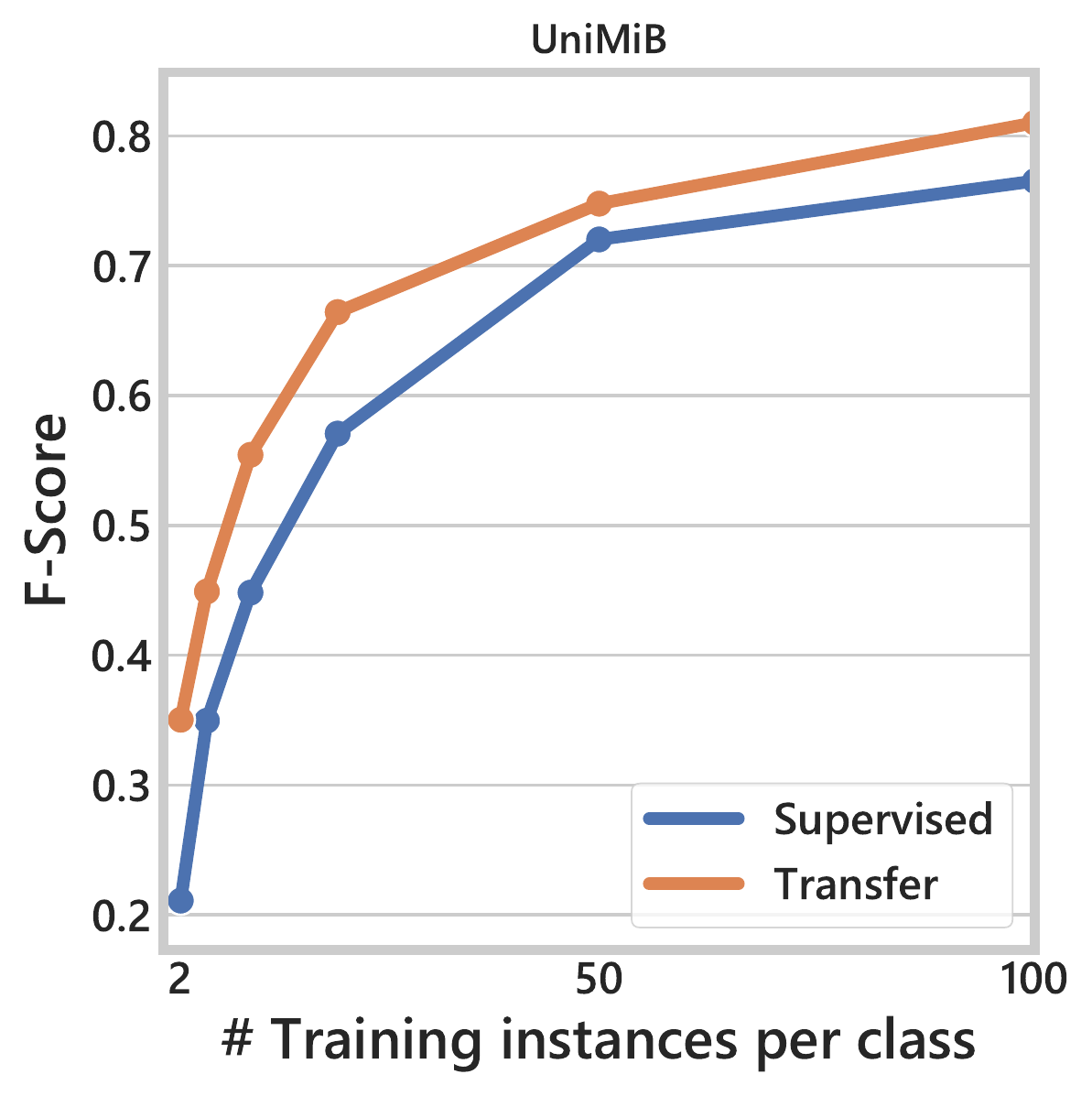}} \hspace{0.01cm}
\subfloat{\includegraphics[width=.3\textwidth]{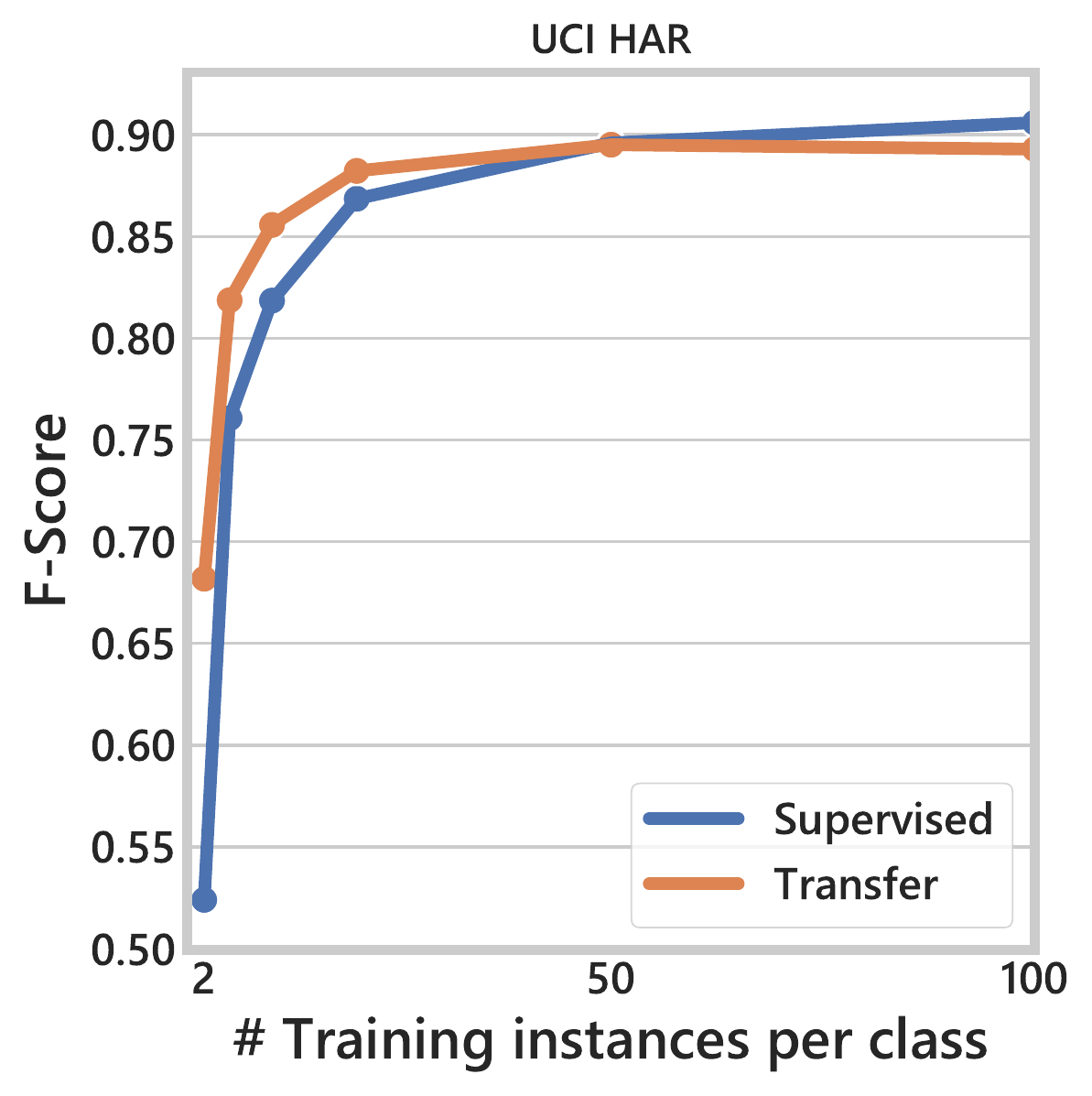}} \hspace{0.01cm}
\subfloat{\includegraphics[width=.3\textwidth]{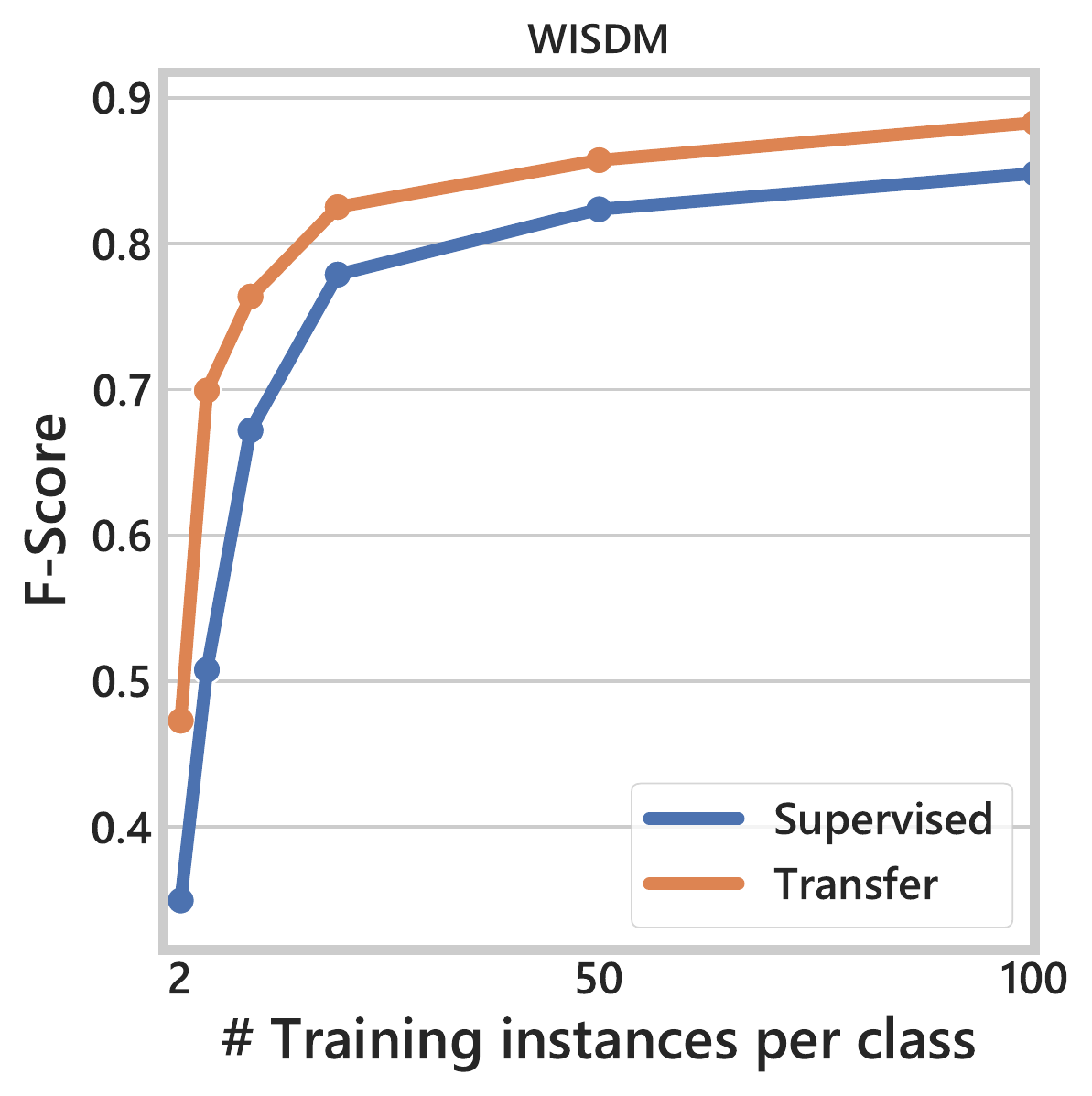}} \hspace{0.01cm}
\subfloat{\includegraphics[width=.3\textwidth]{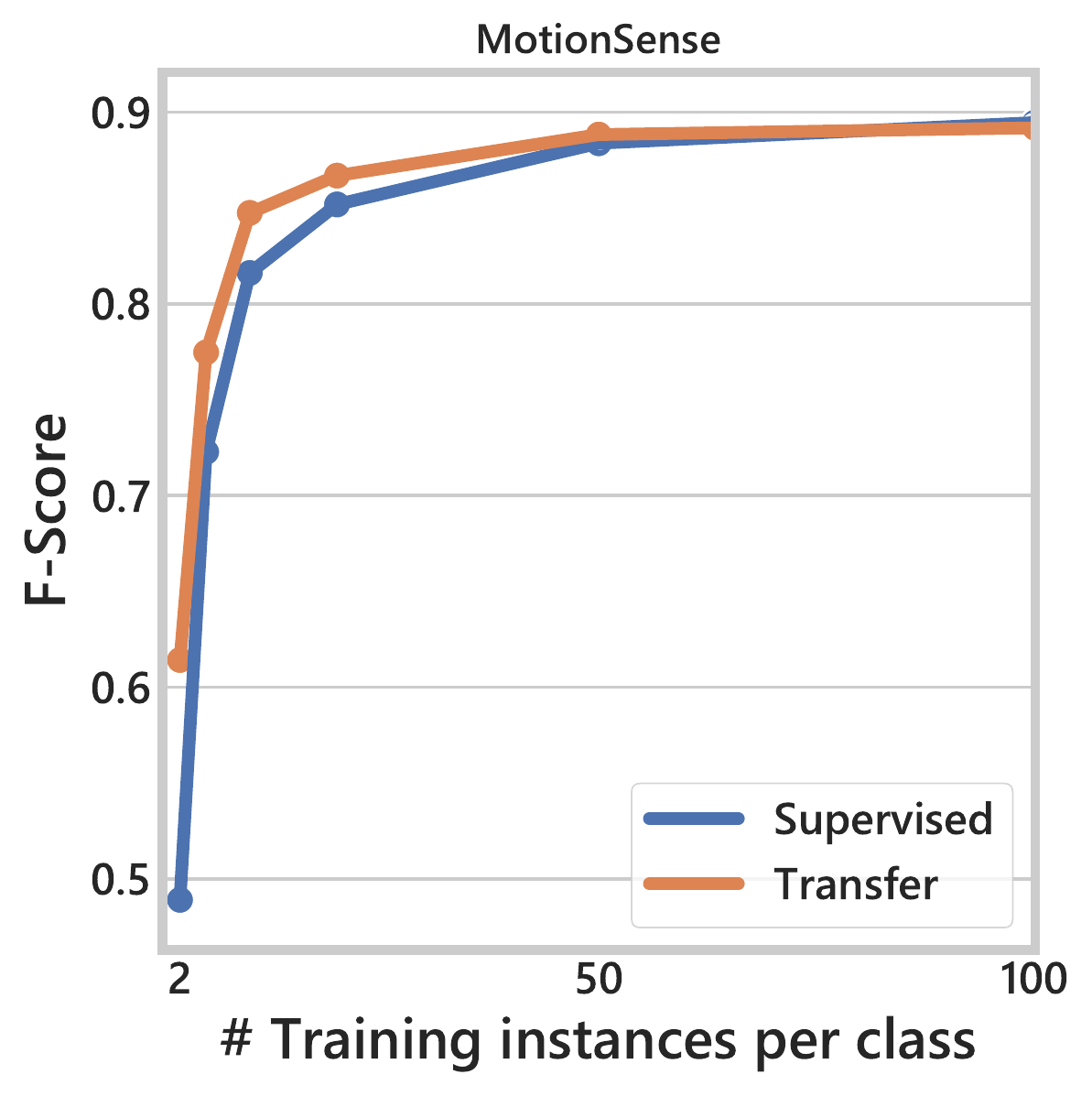}}
\caption{Weighted F-score: Assessment of the transferred self-supervised learned features from a different but related dataset (MobiAct) under semi-supervised setting. \small{The reported results are averaged over $10$ independent runs for each of the evaluated approaches, for more details see Section~\ref{sec:dkt}.}}
\label{fig_appendix:tf_eval_fscore}
\end{figure}

\end{document}